\documentclass[a4,11pt,draftcls,onecolumn]{IEEEtran}

\usepackage{graphicx}
\usepackage{epstopdf}
\usepackage{float}
\usepackage{subfig}

\usepackage[utf8]{inputenc} 
\usepackage[T1]{fontenc}    
\usepackage{hyperref}       
\usepackage{url}            
\usepackage{booktabs}       
\usepackage{amsfonts}       
\usepackage{nicefrac}       
\usepackage{microtype}      
\usepackage{amsmath, amssymb, mathbbol,color}

 \newtheorem{corollary}{\bf Corollary}
\newtheorem{theorem}{\bf Theorem}
\newtheorem{lemma}{\bf Lemma}
\newtheorem{proposition}{\bf Proposition}
\newtheorem{definition}{\bf Definition}
\newtheorem{remark}{\bf Remark}

\begin{document}
\title{Studying the Interplay between Information Loss and Operation Loss in Representations for Classification}

\author{Jorge F. Silva$^*$, Felipe Tobar, Mario Vicuña and Felipe Cordova
\thanks{J. F. Silva, M. Vicuña and F. Cordova  are with the Information and Decision Systems (IDS) Group,  University of Chile, Av. Tupper 2007 Santiago, 412-3, Room 508, Chile, (email: josilva@ing.uchile.cl).}
\thanks{F. Tobar is with the Initiative for Data$\&$Artificial Intelligence,  University of Chile, Beucheff 851, Santiago, Chile (email: ftobar@dim.uchile.cl).}
\thanks{We would like to acknowledge support for this project from ANID-Chile (Fondecyt 1210315) and the Advanced Center for Electrical and Electronic Engineering (Basal Project FB0008).}
}

\maketitle

\begin{abstract}
Information-theoretic measures have been widely adopted in the design of features for learning and decision problems. Inspired by this, we look at the relationship between  {\bf i)} a weak  form of  information loss in the {\em Shannon} sense and {\bf ii)} the operation loss in the {\em minimum probability of error} (MPE) sense when considering a family of lossy continuous representations (features) of a continuous observation. We present several results that shed light on this interplay. Our first result offers a lower bound on a weak form of information loss as a function of its respective operation loss when adopting a discrete lossy representation (quantization) instead of the original raw observation. From this, our main result shows that a specific form of vanishing information loss (a weak notion of asymptotic informational sufficiency) implies a vanishing MPE loss (or asymptotic operational sufficiency) when considering a general family of lossy continuous representations. Our theoretical findings support the observation that the selection of feature representations that attempt to capture informational sufficiency is appropriate for learning, but this selection is a rather conservative design principle if the intended goal is achieving MPE in classification. Supporting this last point, and under some structural conditions, we show that it is possible to adopt an alternative notion of informational sufficiency (strictly weaker than pure sufficiency in the mutual information sense) to achieve operational sufficiency in learning.
\end{abstract}

\newpage

\section{Introduction}
\label{sec_intro}
Given a continuous random object $X$, the problem of representation learning formalizes the task of finding lossy descriptions (or features) of $X$, denoted by $U$, that are sufficient (in some sense) to discriminate a target discrete variable of interest $Y$ (e.g., a class or concept).
In numerous contexts, the raw observation $X$ lives in a finite dimensional continuous space $\mathbb{R}^d$.  In this mixed {\em continuous-discrete setting}, a reasonable assumption is that the raw $X$ is redundant, i.e., there are many explanatory factors that interact in the expression of $X$ beyond $Y$ and, consequently, a lossy description (aka coding) $U$ has the potential to fully capture almost all, or ideally all, the information that $X$ offers to discriminate $Y$ \cite{benyio_2013}. Supporting this idea, it has been shown that under some structural conditions \cite{Bloem_2019,Dubois_2021}, there is a lossy description $U=g(X)$ that is information sufficient in the sense that $I(X;Y) = I(U;Y)$, where $I(X;Y)$ denotes the mutual information (MI) between $X$ and $Y$ \cite{cover_2006}.  
From the data-processing inequality \cite{cover_2006,gray_1990_b},  informational sufficiency implies that $I(X;Y|U)=0$, meaning 
that $X$ and $Y$ are conditionally independent given $U$. A relevant context where this strong separation structure arises is in problems with probabilisitic symmetries or invariances with respect to a group of transformations \cite{Bloem_2019,Dubois_2021}. 

In practice,  lossy descriptions have been instrumental in learning problems because they
regularize the hypothesis space by reducing the complexity/dimensionality of the features, thus providing better generalization
from training to unseen testing conditions, which is arguably the cornerstone of the learning problem \cite{xu_2012,bousquet_2002,shalev-2010,devroye1996,bousquet_2004}. There is a large body of work that addresses the design of lossy representations from data. Many of these approaches rely on the use of information-theoretic measures to quantify the predictive relationship between $X$ and $Y$, using for instance MI $I(X;Y)$, or conditional entropy $H(Y|X)$ or other approaches \cite{achille_2018,amjad_2019,alemi_2017,achille_2018b}. 
%
Along the same lines of learning a minimal (or compressed) sufficient representation from $X$,  
the  Information Bottleneck (IB) method has been adopted in learning and decision \cite{amjad_2019,alemi_2017,achille_2018b,tishby_1999,vera_2018}  to optimize a tradeoff between relevance $I(U;Y)$ and compression $I(U;X)$ over a collection of probabilistic mappings (or channels) from $X$ to  a (latent) variable $U$ \cite{Zaidi_2020}. There is also a deterministic version of the IB problem where the objective is to find the optimal tradeoff between $I(Y,U)$ and $H(U)$ where $U$ is generated through a family of finite (alphabet) deterministic mappings (or vector quantizations) of $X$ \cite{tishby_1999,strouse_2017,tegmark_2020}.

In the context of learning representation as outlined above, the concept of (asymptotic) \emph{sufficiency} can be introduced: an infinite collection of lossy descriptions $U_1,U_2,....$ of $X$ is said to be 
\emph{information sufficient} (IS) 
if $\lim_{i \rightarrow \infty} I(U_i;Y)=I(X;Y)$. In contrast, a collection $U_1,U_2,....$ is said to be \emph{operationally sufficient} (OS) if the performance of classifying $Y$ from $U_i$, in the minimum probability of error (MPE) sense, achieves---as $i$ tends to infinity---the performance of the optimal MPE classifier that uses $X$ losslessly to predict $Y$. Then, a natural question is the following: If a method designs a collection of IS descriptions, is this collection also OS? More generally, is there a strictly weaker notion of IS that implies OS? 

To address this question of whether there is a strictly weak notion of IS that implies OS, in this paper, we focus on studying the interplay between a weak form of information loss and the operation loss over a family of problems (models) induced by lossy continuous representations of $X$.
In particular, we consider a model $(X,Y)$ with joint distribution $\mu_{X,Y}$ and a family of lossy  representations (encoders) $\left\{ U_i\right\}_{i\geq 1}$ of $X$,  where $U_i=g_i(X)$ is a continuous mapping and $\mu_{U_i,Y}$ denotes the joint distribution of $(U_i,Y)$. 
In this context, we introduce a weak form of information loss\footnote{This information loss is formally introduced in Definition \ref{def_wis} - Section \ref{sec_formalization}.} $I((r^*(X),U_i);Y) - I(U_i;Y)\geq 0$ 
where $r^*(X)$ denotes the {\em MPE decision rule}\footnote{$r^*(\cdot)$ is formally introduced in Eq.(\ref{eq_sec_II_6}) - Section \ref{sec_formalization}.} (a finite-size representation of $X$).

\subsection{Contributions}
To justify our weak information loss selection, for the case of discrete representation (i.e., $U_i$ is induced by a quantizer), Theorem \ref{th_2} presents a lower bound for $I((r^*(X),U_i);Y) - I(U_i;Y) \geq 0$ as a function of its respective operation loss (OL) $\ell(\mu_{U_i,Y})- \ell(\mu_{X,Y})\geq 0$ where $\ell(\mu_{U_i,Y}) = \int_{\mathcal{U}_i} (1-\max_{y\in \mathcal{Y}} \mu_{Y|U_i}(y|u))\partial \mu_{U_i}(u)$  and  $\ell(\mu_{X,Y}) = \int_{\mathcal{X}} (1-\max_{y\in \mathcal{Y}} \mu_{Y|X}(y|x))\partial \mu_X(x) \geq 0 $ are the MPE associated to 
the model $\mu_{U_i,Y}$ and $\mu_{X,Y}$, respectively.  OL is the loss attributed to the use of $U_i$ instead of $X$ in classifying $Y$.
Using this bound, our main expressivity result (Theorem \ref{th_1}) shows that if $\left\{ U_i \right\}_{ i\geq 1}$ is  weakly information sufficient (WIS), in the sense that $\lim_{i \rightarrow \infty} \left[  I((r^*(X),U_i);Y) - I(U_i;Y)  \right] = \lim_{i \rightarrow \infty} I(r^*(X) ;Y| U_i) = 0$, then $\left\{ U_i \right\}_{i\geq 1}$ is operationally sufficient (OS) to discriminate $Y$ in the sense that $\lim_{i \rightarrow \infty} \ell(\mu_{U_i,Y}) = \ell(\mu_{X,Y})$ (i.e., $U_i$ achieves the MPE of $X$ in the limit). In other words, a  form of informational sufficiency (strictly weaker than IS mentioned above) implies a vanishing operation loss when $\left\{ U_i\right\}_{i\geq 1}$ is a family of general continuous representations  of $X$. 

On the technical side, we obtain Theorem \ref{th_1} using the bound in Theorem \ref{th_2}, i.e.,  the argument goes from discrete (vector quantizers  (VQ)) to continuous representations. In particular, we build the argument from the scenario of discrete (finite alphabet) representations (stated in Theorem \ref{th_3}, Section \ref{sub_sec_finite_asymtotic}) to prove Theorem \ref{th_1} in the general continuous case in Section \ref{sec_continuous}.
The proofs of Theorems \ref{th_1} and \ref{th_2} rely on two important information theoretic results: The first by Ho {\em et al.} \cite{ho_2010} that characterizes, using a specific rate-distortion function, a tight upper bound for the conditional entropy (equivocation entropy) given an error probability and the second from Liese {\em et al.} \cite{liese_2006} on asymptotic sufficient partitions for mutual information.  

Regarding the optimality of the WIS condition for $\left\{U_i \right\}_{i\geq 1}$ in Theorem \ref{th_1}, we show in Theorem \ref{th_OS_imply_WIS} that when the (optimal) MPE rule 
is unique almost surely w.r.t. to the model $\mu_{X,Y}$, then ``{\em WIS is equivalent to OS}'' 
for a general class of continuous representations of $X$. 
This result offers a context where WIS is tight and optimal as a condition, in the sense that no weaker representation condition on $\left\{U_i \right\}_{i\geq 1}$ could be found to guarantee OS.
 
In the second part of this paper, we work on applications and extensions of our main result (Theorem \ref{th_1}) in learning settings. First, on the dimension of applying Theorem \ref{th_1}, we show that discrete representations (in the form of data-driven VQ) used for the problem of mutual information estimation \cite{silva_2010,silva_2010b,silva_2012} can be adopted in classification to obtain non-supervised representations of $X$ that are operationally sufficient (OS) with  high probability (almost surely). This operational expressivity is demonstrated for two specific data-driven schemes: statistically equivalent blocks \cite{devroye1996,gessaman_1970} and  tree-structured partitions \cite{devroye1996,silva_2010}. These results,  obtained from Theorem \ref{th_1}, reinforce two interesting aspects in representation learning: the expressive power that can be achieved using partitions (digitalization) of the observation space for classification and the capacity of non-supervised methods to achieve informational and operational sufficiency. 

Addressing the more important dimension of extending Theorem \ref{th_1} in learning, we present a learning setting where the true (unknown) model belongs to class of models $\Lambda$. In particular, we study the operational structure of $\Lambda$ needed to extend Theorem \ref{th_1} in this context.  
We found that ifthere is a lossy mapping $\eta_{\Lambda}(\cdot)$ that is operationally sufficient (OS) for  the class $\Lambda$, in the sense that  $\ell(\mu_{X,Y})= \ell(\mu_{\eta_\Lambda(X),Y})$ for any $\mu_{X,Y} \in \Lambda$, then an extension of Th.2 is the following result: if $I(\eta_{\Lambda}(X);Y|U_i) \rightarrow 0$ then $\left\{U_i \right\}_{i\geq 1}$ is OS.  Importantly, this new result (stated in Theorem \ref{th_wis_over_a_family})  replaces $\tilde{U}$ (oracle) by  $\eta_{\Lambda}(X)$ (prior knowledge).  At this point, the problem reduces to finding the ``{\em simplest lossy OS representation}'' ---denoted $\eta_\Lambda(\cdot)$--- for the class $\Lambda \subset \mathcal{P}(\mathcal{X} \times \mathcal{Y})$, if any. This is in general a very difficult problem. Indeed, we could say that a class of models that admits a lossy mapping that is OS for any member is ``{\em compressible}''. Importantly, we determined two concrete  classes of models where this compressible condition is met and, consequently, Theorem \ref{th_wis_over_a_family} can be applied. 
On this, we highlight  the class of invariant models to the action of a compact group\footnote{An excellent exposition for this family  is presented in [7,13].}. For this class, there is a well-stablished lossy mapping $\eta_{\Lambda}(\cdot)$, which is maximal invariant with respect to a group of transformations on $\mathcal{X}$, that precisely meets this operational requirement, i.e.,  $\ell(\mu_{X,Y})= \ell(\mu_{\eta_\Lambda(X),Y})$ for any $\mu_{X,Y} \in \Lambda$.  A simpler example is when $\Lambda$ is a finite collection of models. Here, it is straightforward to construct a lossy mapping $\eta_{\Lambda}(\cdot)$, which is a VQ, that is operationally sufficient for $\Lambda$. In these two contexts, we have that our extended condition $I(\eta_{\Lambda}(X);Y|U_i) \rightarrow 0$  is 
weaker than IS and it could be used in practice to guarantee OS. In summary, by extending Theorem \ref{th_1} to this prior knowledge setting ($\mu \in \Lambda$), we have a way of transferring prior information of the problem (in the form of knowing $\eta_\Lambda(\cdot)$ for $\Lambda$) into a weaker information sufficient condition that is non-oracle and practical from that perspective.
 
\subsection{Related Work}
Our analyses relate fundamentally to the interplay between (minimum) probability of error and conditional entropy (or equivocation entropy) that has been studied systematically in information theory \cite{feder_1994,ho_2010,prasad_2015}.  One of the most recognized results in this area is {\em Fano's inequality} \footnote{$H(Y|X) \leq h(\ell(\mu_{X,Y})) + \ell(\mu_{X,Y}) \log (  \left| \mathcal{Y} \right| -1)$ where $h(r)=-r\log(r) -(1-r) \log (1-r)$ is the binary entropy \cite{cover_2006}.} that offers a lower bound for the probability of error as a function of the entropy \cite{cover_2006}. A refined analysis between conditional entropy and minimum error probability was presented by Feder and Merhav \cite{feder_1994}. They explored the relationship between these quantities providing tight (achievable) lower and upper bounds for the conditional entropy given a minimum error probability restriction. Refining this analysis, Ho and Verd\'{u} \cite{ho_2010} studied a more specific problem that is relevant in the Bayesian treatment of classification: given the prior distribution $\mu_Y$ of $Y$, they were interested in the interplay between the error probability of predicting $Y$ from an observation $X$ and the conditional entropy of $Y$ given $X$ when $X$ is a discrete (finite-alphabet) observation. They provided a closed-form expression for the maximal conditional entropy that can be achieved  as a function of the prior distribution $\mu_Y$ and the minimum probability error $\epsilon$ in the non-trivial regime when $\epsilon \leq  (1-\max_{y\in \mathcal{Y}} \mu_y(y))$. These results offer tight (achievable) bounds between conditional entropy and error probability, thus providing refined and specialized versions of {\em Fano's type of bounds} \cite{ho_2010}. 

Indeed, these bounds were extended to countably-infinite alphabets, a regime for which Fano's original inequality has not been defined.  A relevant corollary of these bounds is the fact that a vanishing  probability of error implies a vanishing conditional entropy.  The converse result is also true under some conditions \cite{feder_1994}. Then,  in cases when the classification task is almost perfect or degenerate (zero probability of error), the relationship between error probability and conditional entropy is rather evident (zero error $\Leftrightarrow$ zero conditional entropy). This connection, however, is less evident for the majority of cases that deviate from this highly discriminative context as it is clearly presented in \cite{feder_1994,ho_2010}. 

The focus of our work in this paper is different from the results mentioned in this subsection as we are interested in the interplay between a form of information loss and its respective operation loss over a family of problems induced by lossy representations of $X$. 

\subsection{Organization}
The rest of the paper is organized as follows. Sections \ref{sec_II} and \ref{sec_formalization} formalize our main question and introduce notation, required concepts and preliminary results. Section \ref{sec_main} presents the statement and interpretations of the main asymptotic result (Theorem \ref{th_1}).  Section \ref{sec_main}  also covers a result for a finite alphabet representation of $X$ (Theorem \ref{th_2}), which is essential to extend the analysis to continuous representations. Section \ref{sub_sec_OS_imply_WIS} discusses about the optimality 
of the WIS condition. 
The application of Theorem \ref{th_1}  in the context of information sufficient schemes for learning is presented in Section \ref{sec_learning_empirical}. Section \ref{sec_wis_applied_in_learning}  extends  the concept of weak informational sufficiency and Theorem \ref{th_1} into a general learning context and covers two relevant practical scenarios.  Concluding remarks and discussions are presented in Section \ref{sec_summary}. The proofs of the main results of this work are presented in Section \ref{sec_proof} and the presentation of supporting technical material are relegated to the Appendix section.

\section{Preliminaries}
\label{sec_II}
Let us consider a decision problem expressed in terms of the joint model (probability) $\mu_{X,Y}\in \mathcal{P}(\mathcal{X}\times \mathcal{Y})$ of a 
vector $(X,Y)$ where $Y$ takes values in a finite space $\mathcal{Y} = \left\{1,..,M \right\}$  (e.g., a \emph{class label}) and $X$ takes values in a finite dimensional space $\mathcal{X} = \mathbb{R}^d$. On the operational side, the {\em minimum probability of error} (MPE) of predicting $Y$ using $X$ as an observation, given the 
the model $\mu_{X,Y}$, is expressed by 
\begin{equation} \label{eq_sec_II_1}
	\ell(\mu_{X,Y})  \equiv  \int_{\mathcal{X}}  (1- \max_{y\in \mathcal{Y}} \mu_{Y|X}(y|x)) \partial \mu_X(x), 
\end{equation}
where 
$\mu_{Y|X}(\cdot|x)$ denotes the {\em probability mass function} (pmf) of $Y$ conditional to the event $ \left\{ X=x \right\}$ and $\mu_X$ denotes the marginal probability of $X$ in $\mathcal{X}$. 
On the information side, the conditional entropy of $Y$ given $X$ --- also known as the equivocation entropy (EE) \cite{feder_1994,ho_2010} ---  is 
\begin{equation} \label{eq_sec_II_2}
	H(Y|X)   \equiv  \int_{\mathcal{X}} \mathcal{H}(\mu_{Y|X}( \cdot |x)) \partial \mu_X(x), 
\end{equation}
where 
$$\mathcal{H}(\mu_{Y|X}(\cdot |x))  \equiv  -  \sum_{y\in \mathcal{Y}} \mu_{Y|X} (y|x) \log  \mu_{Y|X} (y|x)  \leq  \log M$$ 
is 
{\em the Shannon entropy} of  $\mu_{Y|X}(\cdot |x)  \in \mathcal{P}(\mathcal{Y})$ \cite{gray_1990_b,cover_2006}. 
The mutual information (MI) of $\mu_{X,Y}$ is \cite{gray_1990_b,cover_2006} 
\begin{equation} \label{eq_sec_II_3}
	\mathcal{I}(\mu_{X,Y}) = I(X;Y)  \equiv \mathcal{H}(\mu_{Y})   - H(Y|X). 
\end{equation}
The standard notation for MI is $I(X;Y)$, however we also use $\mathcal{I}(\mu_{X,Y})$ to emphasize in our analysis 
that MI is a functional of the joint distribution $\mu_{X,Y}$.

\subsection{Representations of $X$}
\label{sub_sec_representation}
A  representation of $X$ is a measurable function $\eta: (\mathbb{R}^d, \mathcal{B} (\mathbb{R}^d)) \rightarrow (\mathcal{U},\mathcal{B}(\mathcal{U}))$ where $\mathcal{U}$ is the representation space with its respected sigma field denoted by $\mathcal{B}(\mathcal{U})$. In general, we are interested in the case of a lossy mapping $\eta(\cdot)$. To begin our analysis, particular attention will be given to the relevant case where $\left| \mathcal{U} \right|=K<\infty$, meaning that $\eta(\cdot)$ is a {\em vector quantizer} (VQ) of $X$. This VQ\footnote{The main result of this work is for continuous representations. However, studying the case of finite VQs is instrumental as elaborated in Sections \ref{sub_sec_finite_non_asymtotic} and Appendix \ref{sec_proof_th1}.} induces the following finite partition of size $K$: 
\begin{equation} \label{eq_sec_II_4}
	\pi_\eta \equiv \left\{ \eta^{-1}(\left\{ u \right\}), u\in \mathcal{U} \right\}, 
\end{equation}
where it follows that  
$\eta(x)=\sum_{u\in \mathcal{U}} u \cdot  {\bf 1}_{\eta^{-1}(\left\{ u \right\})}(x)$. 

In general, we denote by $U \equiv \eta(X)$ the representation of $X$ induced by $\eta(\cdot)$, and we denote by $\mu_{U,Y}$ the joint distribution of $(U,Y)$ (induced by $\mu_{X,Y}$ and $\eta$)  in  $\mathcal{U} \times \mathcal{Y}$.  As the expressions in (\ref{eq_sec_II_1})  and (\ref{eq_sec_II_3}) are functions of the model $\mu_{X,Y}$, they can be extended naturally to $\mu_{U,Y}$, where  i) $\ell(\mu_{U,Y})$ is the MPE of predicting $Y$ from $U$, and ii) $\mathcal{I}(\mu_{U,Y})$ is the MI between $U$ and $Y$. 

\subsection{Information Loss and Operation Loss}
\label{sub_sec_IL_vs_OL}
We are interested in the \emph{information loss} (IL) of using $U$ (instead of $X$) to resolve $Y$ in the Shannon sense. This can be measured naturally by 
\begin{equation} \label{eq_sec_II_4}
	\mathcal{I}(\mu_{X,Y}) -  \mathcal{I}(\mu_{U,Y})  = I(X;Y | U) \geq 0,    
\end{equation}
where the identity in (\ref{eq_sec_II_4}) comes from the chain rule of MI and the definition of the conditional MI  \cite{gray_1990_b, cover_2006}.  
The main objective of this work is to understand how an information loss of the form in (\ref{eq_sec_II_4}) relates to its respective \emph{operation loss} (OL) of using $U$ (instead of $X$) to classify $Y$ in the MPE sense, i.e.,  
\begin{equation} \label{eq_sec_II_5}
	\ell(\mu_{U,Y}) -  \ell(\mu_{X,Y}) \geq 0. 
\end{equation}

In the following sections, we will study the interplay between a relaxed (weak) information loss expression (of the form in (\ref{eq_sec_II_4})) and the operation loss introduced in (\ref{eq_sec_II_5}) in different regimes and contexts. 

\section{Formalization and Basic Results}
\label{sec_formalization}
Let us consider a family of mappings (or encoders) $\eta_i:  \mathcal{X}  \rightarrow \mathcal{U}_i$, indexed by $i\in \mathbb{N}$, where $\mathcal{U}_i$ is a continuous space, for example a finite dimensional Euclidean space $\mathbb{R}^q$. Using $\eta_i(\cdot)$, we consider the representation variable $U_i = \eta_i(X)$ (e.g., a \emph{feature}) and the respective joint distribution of $(U_i,Y)$ characterized by $\mu_{U_i,Y}$ in  $\mathcal{U}_i \times \mathcal{Y}$. At this point,  we introduce the following definitions for informational and operational sufficiency, respectively.

\begin{definition}\label{def_os}
	A sequence of representations $\left\{ \eta_i (\cdot)\right\}_{i\geq 1}$ (and its respective representation variables $\left\{ U_i\right\}_{i\geq 1}$)  for $X$ is said to be operationally sufficient (OS) for the model $\mu_{X,Y}$ (in the MPE sense) if 
	\begin{equation} \label{eq_main_2}
		\lim_{i \longrightarrow \infty} \ell(\mu_{U_i,Y}) =  \ell(\mu_{X,Y}).
	\end{equation}
\end{definition}

\begin{definition}\label{def_is} 
	 A sequence of representations $\left\{ \eta_i (\cdot) \right\}_{i\geq 1}$ for $X$ (and $\left\{ U_i\right\}_{i\geq 1}$, respectively)  is said to be information sufficient (IS) for $\mu_{X,Y}$ if
	\begin{equation}\label{eq_main_1}
		\lim_{i \longrightarrow \infty}  \mathcal{I}(\mu_{U_i,Y}) =   \mathcal{I}( \mu_{X,Y}).
	\end{equation}
\end{definition}

Let us introduce a weak version of IS for $\mu_{X,Y}$. For this, let us recall that the MPE rule (a sufficient statistic) is a quantizer of $\mathcal{X}$ 
of size $M=\left| \mathcal{Y} \right|$ given by\footnote{The optimal rule in (\ref{eq_sec_II_6}) is not unique. If for some $x$ 
many $y$ achieves the minimum in (\ref{eq_sec_II_6}), we select the smallest one to define $\tilde{r}_{\mu_{X,Y}} (x)$.}
\begin{equation} \label{eq_sec_II_6}
	\tilde{r}_{\mu_{X,Y}} (x) \equiv  \arg \max_{y\in \mathcal{Y}} \mu_{Y|X}(y|x).
\end{equation}
This rule induces both a
(distribution dependent) partition of $\mathcal{X}$ given by\footnote{where $\tilde{r}_{\mu_{X,Y}} (x)= \sum_{y \in \mathcal{Y}} {\bf 1}_{A^*_y}(x) \cdot y$.}
\begin{equation} \label{eq_sec_II_7}
	\pi^* \equiv  \left\{ A^*_y \equiv \tilde{r}_{\mu_{X,Y}}^{-1}(\left\{ y \right\}), y \in \mathcal{Y} \right\}, 
\end{equation}
and 
a finite alphabet lossy representation of $X$ given by $\tilde{U} \equiv \tilde{r}_{\mu_{X,Y}}(X)\in \mathcal{Y}$.
\begin{definition}\label{def_wis} 
	 A sequence of representations $\left\{ \eta_i(\cdot) \right\}_{i\geq 1}$ for $X$ (and  $\left\{ U_i\right\}_{i\geq 1}$) is said to be weakly information sufficient (WIS) 
	 for   $\mu_{X,Y}$ if
	\begin{equation}\label{eq_sec_II_8d}
		\lim_{i \longrightarrow \infty}   \underbrace{\mathcal{I}( \mu_{(\tilde{U},U_i),Y}) - \mathcal{I}(\mu_{U_i,Y})}_{I(Y;\tilde{U}|U_i)\geq 0}= 0, 
	\end{equation}
	where  $I(Y;\tilde{U}|U_i)$ is the conditional MI between $Y$ and $\tilde{U}$ given $U_i$ \cite{cover_2006}.
\end{definition}

\subsection{Preliminary Analysis}
Let us first consider the discrete case where $\mathcal{U}_i = \left\{1,..,k_i \right\}$ for any $i\geq 1$. 
In this context, we can elaborate expressions for  
							$\mathcal{I}( \mu_{(\tilde{U},U_i),Y}) - \mathcal{I}(\mu_{U_i,Y})$
and $\ell(\mu_{U_i,Y}) -  \ell(\mu_{X,Y})$. 
For this, let us consider the finite partition induced by the mapping $\eta_i(\cdot)$  which is given by
\begin{equation} \label{eq_sec_II_8}
	\pi_{\eta_i} \equiv  \left\{ B_{i,j} \equiv  \eta_i^{-1}(\left\{ j \right\}), j \in \mathcal{U}_i = \left\{1,..,k_i \right\} \right\}, 
\end{equation}
where $\eta_i(x)= \sum_{j \in \mathcal{U}_i} {\bf 1}_{B_{i,j}}(x) \cdot j$.

The following results present useful  expressions for the losses in (\ref{eq_sec_II_5}) and (\ref{eq_sec_II_4}) in terms 
of the model $\mu_{X,Y}$ and the cells of $\pi^*$ and $\pi_{\eta_i}$, respectively.
\begin{proposition}\label{pro_opt_loss}
	$\ell(\mu_{U_i,Y}) -  \ell(\mu_{X,Y}) = \sum_{B_{i,j} \in \pi_{\eta_i}} \mu_X(B_{i,j}) \cdot g(\mu_{X,Y}, B_{i,j})$, where 
	\begin{equation}\label{eq_sec_II_8b}
		g(\mu_{X,Y}, B_{i,j}) \equiv  \left[   1- \max_{y\in \mathcal{Y}} \mu_{Y|X}(y | B_{i,j}) \right]  -  \sum_{ A^*_u \in \pi^*} \frac{\mu_X(A^*_u \cap B_{i,j} )}{\mu_X(B_{i,j})} \left[   1- \max_{y\in \mathcal{Y}} \mu_{Y|X}(y | A^*_u \cap B_{i,j}) \right].
	\end{equation}
\end{proposition}
The operation loss in Proposition \ref{pro_opt_loss} is expressed as the weighted sum of the terms $\left\{ g(\mu_{X,Y}, B_{i,j}) \right\}_{B_{i,j}\in \pi_{\eta_i}}$, each one of them associated with a non-negative contribution (in the loss) indexed by individual cells of $\pi_{\eta_i}$.
\begin{remark}\label{rmk_1}
The term $g(\mu_{X,Y}, B_{i,j}) \geq 0$ can be interpreted as the gain in MPE from a ``{\em prior scenario}" where the marginal distribution of $Y$ follows $(\mu_{Y|X}(y|B_{i,j}))_{y\in \mathcal{Y}} \in \mathcal{P}(\mathcal{Y})$  to a ``{\em posterior scenario}" where we observe $\tilde{U} = \tilde{r}_{\mu_{X,Y}}(X)$ to classify $Y$ under the joint conditional model  $(\mu_{\tilde{U},Y|X}(u, y | B_{i,j}) \equiv \frac{\mu_{X,Y}(A^*_u\cap B_{i,j} \times \left\{ y \right\} )}{\mu_X(B_{i,j})})_{(u,y) \in \mathcal{Y}^2}$ in $\mathcal{P}(\mathcal{Y} \times \mathcal{Y})$.\footnote{This Bayesian gain interpretation of the term $g(\mu_{X,Y}, B_{i,j})$ will be central for the results in Section \ref{sec_main}.} 
\end{remark}

On the information loss, instead of looking at $I(X;Y | U_i)$ in (\ref{eq_sec_II_4}), we decided to consider the MI loss of observing $U_i$ with respect to a re-defined reference case where we observe the vector $(\tilde{U} ,U_i)$ with $\tilde{U} = \tilde{r}_{\mu_{X,Y}}(X) $, which is a deterministic function of $X$. 
 Intuitively, this choice is based on the observation that $\tilde{U}$ is a sufficient statistic for $X$ in the operational MPE sense, see  Eq.(\ref{eq_sec_II_6}). Consequently, our re-defined information loss is
\begin{proposition}\label{pro_inf_loss}
	$\mathcal{I}( \mu_{(\tilde{U}, U_i),Y}) -  \mathcal{I}(\mu_{U_i,Y}) = \sum_{B_{i,j} \in \pi_i } \mu_X(B_{i,j}) \cdot I(\tilde{U}; Y| X\in B_{i,j})$, where 
	\begin{equation} \label{eq_sec_II_8c}
		I(\tilde{U}; Y| X\in B_{i,j}) \equiv \mathcal{H}(\mu_{Y|X}(|B_{i,j}))  -  \sum_{ A^*_u \in \pi^*} \frac{\mu_X(A^*_u \cap B_{i,j} )  }{\mu_X(B_{i,j})}  \mathcal{H}(\mu_{Y|X}( | A^*_u \cap B_{i,j}))
	\end{equation}
	is the MI between $Y$ and  $\tilde{U} = \sum_{u \in \mathcal{Y}} u \cdot  {\bf 1}_{A^*_u}(X)$  conditioning 
	on the event $\left\{ X\in B_{i,j} \right\}$. 
\end{proposition}

Alternatively, we have that $\mathcal{I}( \mu_{(\tilde{U}, U_i),Y}) -  \mathcal{I}(\mu_{U_i,Y}) = I(\tilde{U}, U_i;Y) - I(U_i;Y) = I(\tilde{U}; Y| U_i)$.

\begin{remark}\label{rmk_2}
Using the observation made in Remark \ref{rmk_1}, it is worth noting the conceptual connection between $g(\mu_{X,Y}, B_{i,j})$ in (\ref{eq_sec_II_8b}), prior minus posterior in the MPE sense condition on $\left\{ X\in B_{i,j} \right\}$,  and $I(\tilde{U}; Y| X\in B_{i,j})$ in (\ref{eq_sec_II_8c}), which is the prior minus the posterior Shannon entropy condition on the same event $\left\{ X\in B_{i,j} \right\}$. 
\end{remark}

\section{Interplay between Information Loss and Operation Loss}
\label{sec_main}
Before presenting the main asymptotic result of this section (Theorem \ref{th_1}), it is relevant to find a lower bound on the  information loss expressed in Proposition \ref{pro_inf_loss} as a function of the operation loss expressed in Proposition \ref{pro_opt_loss}. To that end, we introduce the following instrumental lemma:
\begin{lemma}\label{th_interplay_cond_entropy_error_prob} (Ho and Verd\'{u} \cite[Th.4]{ho_2010})
	Let us consider $Y$ a random variable in $\mathcal{Y}= \left\{1,..,M \right\}$ and  a finite observation space $\mathcal{X}$
	such that $ \left| \mathcal{X}  \right| \geq M$. If we denote by $\mathcal{P}(\mathcal{X}|\mathcal{Y})$ the collection 
	of conditional probabilities from $\mathcal{Y}$ to $\mathcal{X}$ (or channels),  then for any non-negative $\epsilon \leq \underbrace{(1- \max_{y \in \mathcal{Y}}\mu_Y(y))}_{\text{the prior error of $\mu_Y$}}$, it follows that  
	\begin{equation}\label{eq_main_1}
		f(\mu_Y, \epsilon) \equiv \min_{\rho_{X|Y} \in \mathcal{P}(\mathcal{X}|\mathcal{Y}) st. \ell(\rho_{X|Y} \mu_Y)=\epsilon} \mathcal{I}(\rho_{X|Y} \mu_Y) =  \mathcal{H} (\mu_Y) - \mathcal{H} (\mathcal{R}(\mu_Y, \epsilon )) \geq 0, 
	\end{equation}
	where $\rho_{X|Y} \cdot \mu_Y$ is a joint probability in $\mathcal{P}(  \mathcal{X} \times \mathcal{Y})$ and 
	$\mathcal{R}(\mu_Y, \epsilon) \in \mathcal{P}(\mathcal{Y})$ is a well-defined probability function of $\mu_Y$ and $\epsilon$. \footnote{The closed-form expression of the probability $\mathcal{R}(\mu_Y, \epsilon)$ is presented in \cite{ho_2010} and in  Appendix \ref{appendix_rate_distortion_function}.}
\end{lemma}
This result offers a tight (achievable) lower bound on the minimum MI achieved by a family of joint models in $\mathcal{P}(  \mathcal{X} \times \mathcal{Y})$ that satisfies two conditions: i) they meet an MPE restriction parametrized by $\epsilon \in [0,  1- \max_{y \in \mathcal{Y}}\mu_Y(y)]$
and ii) they have a marginal distribution on $Y$ given by a fixed model $\mu_Y \in \mathcal{P}(\mathcal{Y})$. Importantly, for the non-trivial case when $\epsilon < (1- \max_{y \in \mathcal{Y}}\mu_Y(y))$, \cite{ho_2010} shows that  $\mathcal{H} (\mu_Y) > \mathcal{H} (\mathcal{R}(\mu_Y, \epsilon )) \Rightarrow f(\mu_Y, \epsilon) >0$, while for the trivial case when $\epsilon = (1- \max_{y \in \mathcal{Y}}\mu_Y(y))$ this work shows that $\mathcal{R}(\mu_Y, \epsilon) =\mu_Y \Rightarrow f(\mu_Y, \epsilon) =0$ \cite{ho_2010}. \footnote{In information theory, the function $f(\mu_Y, \epsilon)$ in (\ref{eq_main_1}) is a special case of the celebrated  rate distortion function of a memoryless source (i.i.d.) with marginal distribution $\mu_{Y}$ and distortion function given by the hamming distance (or the $0$-$1$ loss) \cite{gray_1990}.}

\begin{remark}\label{rmk_2}
Considering a discrete observation $X$ such that $(X,Y)\sim \mu_{X,Y}$,   Lemma \ref{th_interplay_cond_entropy_error_prob} can be used directly to obtain a lower bound for $I(X;Y) =  \mathcal{I}(\mu_{X,Y}) $ as a function of $\ell(\mu_{X;Y})$. More precisely, from  (\ref{eq_main_1}) we have that
\begin{equation}\label{eq_main_1b}
	I(X;Y)=  \mathcal{I}(\mu_{X,Y}) \geq f(\mu_Y, \ell(\mu_{X,Y})) =  H(Y) - \mathcal{H} (\mathcal{R}(\mu_Y, \ell(\mu_{X,Y}))).
\end{equation}
Importantly, the bound in (\ref{eq_main_1b}) recovers the known fact that if $\ell(\mu_{X;Y}) < (1- \max_{y \in \mathcal{Y}}\mu_Y(y))$ then $I(X;Y)>0$,  or,  conversely,  $I(X;Y)=0$ (zero information) implies  $\ell(\mu_{X;Y}) = (1- \max_{y \in \mathcal{Y}}\mu_Y(y)) $, i.e., there is a zero gain in MPE (from the prior) when observing $X$.
\end{remark}

\subsection{A Non-Asymptotic Result}
\label{sub_sec_finite_non_asymtotic}
Returning to our original mixed continuous-discrete setting (Section \ref{sec_II}), the application of Lemma \ref{th_interplay_cond_entropy_error_prob} in the context of our weak information loss vs. operation loss analysis is the following result: 
\begin{theorem}\label{th_2}
	Let us consider our model $\mu_{X,Y}$ and a finite alphabet lossy representation $U_i$ (induced by $\eta_i(\cdot)$) of $X$,
	then
	\begin{equation} \label{eq_main_3}
		\mathcal{I}( \mu_{(\tilde{U}, U_i),Y}) -  \mathcal{I}(\mu_{U_i,Y})  \geq   \sum_{B_{i,j} \in \pi_i } \mu_X(B_{i,j}) \cdot  \left[ \mathcal{H}(\mu_{Y|X}(\cdot |B_{i,j})) -  \mathcal{H} (\mathcal{R}(\mu_{Y|X}(\cdot |B_{i,j}), \epsilon_{i,j} )\right] \geq 0,
	\end{equation}
	where 
	\begin{align*}
	\epsilon_{i,j}      &\equiv  \left[   1- \max_{y\in \mathcal{Y}} \mu_{Y|X}(y | B_{i,j}) \right]  - g(\mu_{X,Y}, B_{i,j})\\ 
				&=\sum_{ A^*_u \in \pi^*} \frac{\mu_X(A^*_u \cap B_{i,j} )}{\mu_X(B_{i,j})} \left[   1- \max_{y\in \mathcal{Y}} \mu_{Y|X}(y | A^*_u \cap B_{i,j}) \right]
	\end{align*}
	and $g(\mu_{X,Y}, B_{i,j})$ as in (\ref{eq_sec_II_8b}). 
\end{theorem}

Remarks and implications of Theorem \ref{th_2}:
\begin{enumerate}
\item  
The lower bound on the information loss in  (\ref{eq_main_3}) is an explicit function of the decomposition of the operation loss presented below in (\ref{eq_main_2}). On the proof of Theorem \ref{th_2}, the expression in (\ref{eq_main_3}) comes from interpreting the operation loss  in (\ref{eq_main_2}) as the sum of some posterior minus prior MPE gains (see Remark \ref{rmk_1}) and the application of Lemma \ref{th_interplay_cond_entropy_error_prob} in this context (see the proof of this result in Section \ref{proof_TH1}).
\item  
\begin{corollary} \label{cor_th_2}
Let us assume a positive operation loss, i.e.,  
	\begin{equation} \label{eq_main_2}
	\ell(\mu_{U_i,Y}) -  \ell(\mu_{X,Y})= \sum_{B_{i,j} \in \pi_i } \mu_X(B_{i,j}) \cdot g(\mu_{X,Y}, B_{i,j})>0, 
	\end{equation}
then from 	Theorem \ref{th_2} it follows that $\mathcal{I}( \mu_{(\tilde{U}, U_i),Y}) -  \mathcal{I}(\mu_{U_i,Y}) >0$.
\end{corollary}
{\bf Proof of Corollary \ref{cor_th_2}}: Assuming that $\ell(\mu_{U_i,Y}) -  \ell(\mu_{X,Y})>0$ in (\ref{eq_main_2}),  
this implies that at least  one component $j$ of the sum 
satisfies that $g(\mu_{X,Y}, B_{i,j})>0 \Leftrightarrow$  $\epsilon_{i,j} < \left[   1- \max_{y\in \mathcal{Y}} \mu_{Y|X}(y | B_{i,j}) \right]$ (see (\ref{eq_main_2})). Then  Lemma \ref{th_interplay_cond_entropy_error_prob} implies that $\mathcal{H}(\mu_{Y|X}(|B_{i,j})) -  \mathcal{H} (\mathcal{R}(\mu_{Y|X}(|B_{i,j}), \epsilon_{i,j} )>0$. This last bound and (\ref{eq_main_3}) suffice to show that 
\begin{equation} \label{eq_main_4}
\mathcal{I}( \mu_{X,Y}) -  \mathcal{I}(\mu_{U_i,Y}) \geq \mathcal{I}( \mu_{(\tilde{U}, U_i),Y}) -  \mathcal{I}(\mu_{U_i,Y})>0.
\end{equation}
The first inequality in (\ref{eq_main_4}) comes from the fact that $(\tilde{U},U_i)$ is a deterministic function of $X$ and the second comes  from (\ref{eq_main_3}). $\blacksquare$\\  
Therefore,  a non-zero  operation loss on using $U_i$ instead of $X$ (stated in (\ref{eq_main_2})) implies a respective non-zero weak information loss as stated in (\ref{eq_main_4}). 
\item 
Corollary \ref{cor_th_2} implies that if $U_i=\eta_i(X)$ (for some finite $i$) is weakly information sufficient 
in the sense that $\mathcal{I}( \mu_{(\tilde{U}, U_i),Y}) -  \mathcal{I}(\mu_{U_i,Y})=I(\tilde{U};Y|U_i)=0$,  then  
$\ell(\mu_{U_i,Y}) =  \ell(\mu_{X,Y})$, i.e.,  
$\eta_i$ (and $U_i$) is operational sufficient for $\mu_{X,Y}$. 
\item It is worth noting that  for a large class of models (continuos in nature), $U_i$ being weakly information sufficient, i.e., $I(\tilde{U};Y|U_i)=0$,  
is strictly weaker than asking that $U_i$ is information sufficient for $\mu_{X,Y}$ in the sense that $I(X;Y|U_i)=0$.  
In fact,  $I(X;Y|U_i)=0$  implies that $I(\tilde{U};Y|U_i)=0$ from the observation that 
$\tilde{U}$ is a deterministic function of $X$ and the chain rule of the MI \cite{cover_2006}, but the converse result is not true in general.\footnote{In contrast,  examples for  $\mu_{X,Y}$ can be constructed, which are discrete in nature,  where $I(X;Y)=I(\tilde{U};Y)$. Here, $\tilde{U}$ (a discrete variable of size $M$) is IS for $\mu_{X,Y}$. In this trivial discrete context, it is simple to verify that $I(X;Y|U_i)$=$I(\tilde{U};Y|U_i)$ independent of $U_i$.}  
\item The difference between the pure information loss (IL), i.e., $I(X;Y|U_i)$, and  the weak information loss (WIL), i.e., $I(\tilde{U};Y|U_i)$, is further discussed in Section \ref{subsec_wis_vs_is} and its non-zero discrepancy is illustrated by an example in Section \ref{ilustration_th1}. 
\end{enumerate}

\subsection{The Asymptotic Result}
\label{sub_sec_finite_asymtotic}
The following is the main asymptotic result of this section that shows that a family of weakly IS  
representations for $\mu\in \mathcal{P}(\mathcal{X} \times \mathcal{Y})$  is operation sufficient for $\mu$. 
This result can be interpreted as a non-trivial asymptotic extension of Corollary \ref{cor_th_2} (of Theorem \ref{th_2}). 

\begin{theorem}\label{th_1}
	Let  $\left\{ U_i\right\}_{i\geq 1}$ be a sequence of representations for $X$ obtained from $\left\{ \eta_i (\cdot) \right\}_{i\geq 1}$. 
	If $\left\{ U_i \right\}_{i\geq 1}$ is  WIS for $\mu_{X,Y}$ (Definition \ref{def_wis}), then $\left\{ U_i, i\geq 1 \right\}$ is OS for  
	$\mu_{X,Y}$ (Definition \ref{def_os}). 
\end{theorem}

Remarks about the statement of Theorem \ref{th_1} and its interpretation:
\begin{enumerate}
\item 
The proof of Theorem \ref{th_1} presented in Section \ref{sec_continuous} shows that if a family of representations $\left\{ U_i \right\}_{i\geq 1}$ is not operation sufficient, i.e., 
$\lim\inf_{i \rightarrow \infty} \ell(\mu_{U_i,Y}) -  \ell(\mu_{X,Y})>0$,  then 
\begin{equation} \label{eq_sec_II_9d}
\lim \inf_{i \rightarrow \infty} \mathcal{I}( \mu_{(\tilde{U}, U_i),Y}) -  \mathcal{I}(\mu_{U_i,Y})= \lim \inf_{i \rightarrow \infty} I(\tilde{U}; Y|U_i) >0.
 \end{equation}
\item  The condition (\ref{eq_sec_II_8d}) (WIS in Definition \ref{def_wis}) means that as $i$ tends to infinity,  $U_i$ captures all the information (in the Shannon sense) that $\tilde{U}$ has to offer to resolve  the uncertainty of $Y$.  In general for continuous models in $\mathcal{P}(\mathcal{X}\times \mathcal{Y})$, we have that $I(\tilde{U}; Y) < I(X; Y)$  because $\tilde{U}$ is an $M$ size quantized version of $X$ (see Eq.(\ref{eq_sec_II_7})). Then,  the sufficient condition stated in (\ref{eq_sec_II_8d}) is strictly weaker (for a large class of models) than asking for informational sufficiency (Definition \ref{def_is}). 
\item The condition in (\ref{eq_sec_II_8d}) and Theorem \ref{th_1} further emphasize the fact that achieving pure sufficiency in the Shannon sense is very conservative if the operational objective is classification, as a strictly weaker notion does exist that guarantees operational sufficiency. An example is presented 
in Section \ref{ilustration_th1} that illustrates this point. 

\item Complementing the previous point, 
it is evident that  $\left\{ U_i \right\}_{i\geq 1}$ being operationally sufficient for $\mu_{X,Y}$  does not imply that  $\left\{ U_i \right\}_{i\geq 1}$ is information sufficient for $\mu_{X,Y}$, in general. We illustrate this with an example in Section \ref{ilustration_th1}. Conversely, if $\left\{ U_i \right\}_{i\geq 1}$ is not OS,   then  $\lim \sup_{i \rightarrow \infty} \mathcal{I}(\mu_{U_i,Y}) <  \mathcal{I}( \mu_{X,Y})$ considering that  $\mathcal{I}( \mu_{(\tilde{U}, U_i),Y}) \leq \mathcal{I}(\mu_{X,Y})$ for any $i\geq 1$ and (\ref{eq_sec_II_9d}).
\item
Finally, it is worth mentioning that WIS, as a condition on $\left\{ \eta_i (\cdot) \right\}_{i\geq 1}$, is theoretically interesting for  the reasons mentioned in previous points, but it is unnatural in terms of its practical adoption for feature design in a learning context.  The reason is that the reference representation $\tilde{U}$ (used in (\ref{eq_sec_II_8d})) is a function of the model $\mu_{X,Y}$, which is by nature unavailable in learning. This practical issue motivates the extensions of Theorem \ref{th_1} into a learning setting presented in Section \ref{sec_wis_applied_in_learning}.
\end{enumerate}

\subsection{How much weaker is WIS than IS?}
\label{subsec_wis_vs_is}
On the significance of our main result (Theorem \ref{th_1}), a key aspect is to analyze and evaluate how much weaker  the WIS condition  used in Theorem \ref{th_1} is with respect to the traditional IS.
The analysis implies looking at the differences between the information losses, i.e., the difference between $I(X;Y)$ and $I((\tilde{U},U_i);Y)$. On 
this, we could say that:
\begin{itemize}
	\item The analysis of IL-WIL $=I(X;Y) - I((\tilde{U},U_i);Y)\geq 0$ depends on the model $\mu_{X,Y}$ and the representation $U_i$. 
	The weak information loss uses $\tilde{U}$ (a quantized version of $X$ of size $M$) as a reference, while IL uses $X$, which is a continuous random variable in the context of our general model $\mu_{X;Y}$. 
	\item In information theory, it is known that $I(X;Y)$ is the supremum of the discrete MI between $\eta(X)$ and $Y$ over all possible finite-size quantizers $\eta(\cdot)$ \cite{liese_2006,silva_2010,vajda_2002} (see results in Section \ref{subsec_information_sufficiient}). 
	We could say that a scenario (a model $\mu_{X,Y}$) where MI is not achieved by any finite size version of $X$ makes the model $\mu_{X,Y}$ continuous from a MI point of view \cite{liese_2006,silva_2010,vajda_2002}. In contrast, a model where a quantized version of $X$ achieves the MI between $X$ and $Y$ makes the model $\mu_{X,Y}$ discrete from a MI point of view \cite{cover_2006}.
	\item Assuming the non-trivial case that $\mu_{X;Y}$ (the model) is continuous, i.e.,  $I(X;Y)$ is not achieved by any finite-alphabet function (or vector quantization) of $X$, we have that for any representation $\eta_i(\cdot)$ ($U_i$) that is a VQ: $I((\tilde{U},U_i);Y) < I(X;Y)$. Consequently, WIL is strictly smaller than IL for the rich case where we have a continuous model and finite alphabet representations. 
	\item On the previous point, the continuous scenario for $\mu_{X;Y}$ is an essential case study for the analysis presented in this paper as we do not impose any  structural assumptions on $\mu_{X,Y}$. Also, in many practical domains with continuous observations (images, audio, acoustic sources, etc), it is reasonable to consider that a quantized (digital) version of $X$ induces a non-zero loss of mutual information about $Y$.
\end{itemize}

\subsection{An illustrative Example}
\label{ilustration_th1}
To illustrate the gap between WIS and IS and the significance of our oracle result, here we present a simple construction
to analyze the interplay between IS vs. OS and WIS vs. OS. 
\begin{itemize}
	\item $Y$ takes two values with $\mu_Y(1)=\mu_Y(2)=1/2$.
	\item $X$ given $Y$ follows a Gaussian distribution: $X \sim Normal(K,\sigma)$ when $Y=1$ and $X \sim Normal(- K,\sigma)$ when $Y=2$. $K>0$ and $\sigma>0$ (the parameters). 
	\item The MPE decision is: $\tilde{U}=1$ if $X\geq 0$ and $\tilde{U}=2$ if $X<0$.
	\item Let us consider the following collection of indexed partitions: 
		$\pi_1=\left\{(-\infty,-1/2),[-1/2,1/2], (1/2,\infty) \right\}$; 
		$\pi_2=\left\{(-\infty,-1/4),[-1/4,1/4], (1/4,\infty) \right\}$; 
… $\pi_i=\left\{(-\infty,-1/2^i),[-1/2^i,1/2^i], (1/2^i,\infty) \right\}$,....
	 \item If we denote by $A^1_i$, $A^2_i$ and $A^3_i$ the cells of $\pi_i$, this produces a VQ of $X$ determined by: $U_i=1$ if $X\in A^1_i$, $U_i=2$ if $X\in A^2_i$, and $U_i=3$ if $X\in A^3_i$.
 	\item It is simple to show that $I(U_i;Y)<I(X;Y)$ (as the model is continuous) \cite{liese_2006,silva_2010,cover_2006} and that  $\lim_{i \rightarrow \infty} I(U_i;Y) < I(X;Y)$. In other words, the collection $\left\{U_i\right\}_{i\geq 1}$ is not information sufficient: i.e, $I(X;Y)- I(U_i;Y)= I(X;Y|U_i)$ is not vanishing as $i$ tends to infinity.
	\item In contrast, by the construction of this family, $U_i$ determines $\tilde{U}$ in the limit (it follows that  $\lim_{i \rightarrow \infty} H(\tilde{U}|U_i)=0$) and, consequently, we have that $\lim_{i \rightarrow \infty} I(\tilde{U};Y|U_i)=0$ \cite{cover_2006}. Therefore, this family of representations $\left\{U_i, i\geq 1\right\}$ is WIS. 
	\item Finally, from  Theorem \ref{th_1}, $\left\{U_i, i\geq 1\right\}$ is OS (Def. \ref{def_os}) but not IS (Def.\ref{def_is}).
\end{itemize}

This example illustrates  an scenario where the difference between IL and WIL is strictly positive and relevant for any $i$. This discrepancy is non-vanishing when $i$ grows: WIL tends to zero, but IL does not. Indeed, WIS (Def. \ref{def_wis}) is strictly weaker than IS (Def.\ref{def_is}) in this example. Furthermore in the context of this example, we observe that IL as fidelity indicator is blind on predicting the quality that the collection $\left\{U_i, i\geq 1\right\}$ has to achieve the MPE  in (\ref{eq_sec_II_1}). 

In our final numerical analysis in Section \ref{sec_numerical:tradeoff}, we show other constructions (VQs) and models where the same finding illustrated in this example is experimentally observed.

\section{On the Optimality of WIS} 
\label{sub_sec_OS_imply_WIS}
On this section we show that WIS is equivalent to OS if we consider a uniqueness condition of the Bayes rule in (\ref{eq_sec_II_6}) of a given model $\mu_{X,Y}$. More precisely, 
\begin{definition}
	\label{def_unique_map_rule}
	A model $\mu_{X,Y}$ is said to have a unique MPE decision rule,  if for any rule $r: \mathcal{X}  \rightarrow  \mathcal{Y}$ such that $\mathbb{P}(r(X)\neq Y)=\ell(\mu_{X,Y})$ then $r(\cdot)$ is equal to the MAP rule $\tilde{r}_{\mu_{X,Y}} (\cdot)$ in (\ref{eq_sec_II_6}) almost surely. \footnote{More preciasly, $\mathbb{P}(r(X)\neq \tilde{r}_{\mu_{X,Y}} (X))=\mu_X(\left\{x\in \mathcal{X}: r(x)\neq \tilde{r}_{\mu_{X,Y}}(x) \right\})=0$.}
\end{definition}

\begin{theorem} \label{th_OS_imply_WIS}
	Let  $\left\{ U_i\right\}_{i\geq 1}$ be a sequence of representations for $X$ and let us assume that $\mu_{X,Y}$ has a unique MPE decision rule (Def. \ref{def_unique_map_rule}). Then,  $\left\{ U_i \right\}_{i\geq 1}$  is OS for $\mu_{X,Y}$ if, and only if, $\left\{ U_i \right\}_{i\geq 1}$ is WIS for $\mu_{X,Y}$.
\end{theorem}
Therefore, under this uniqueness condition of the optimal rule for $\mu_{X,Y}$, WIS is the weakest condition on the representations needed to achieve OS. From that perspective, our main result in Theorem \ref{th_1} can be considered optimal for this family of models.

The proof of Theorem \ref{th_OS_imply_WIS} uses the uniqueness condition in Definition \ref{def_unique_map_rule} to show that OS implies WIS. The argument 
is presented in Section \ref{proof_th_OS_imply_WIS}.

\section{Connections with Information Sufficient Schemes}
\label{sec_learning_empirical}
Looking at the application of the results of Section \ref{sec_main} in a learning context, we present connections with information sufficient analysis and results on MI estimation using partitions. On the approximation error side of this last learning problem, there are interesting results in the literature guaranteeing that a collection of discrete representations (partitions) are asymptotically sufficient for approximating $I(X;Y)$ \cite{berlinet_2005,liese_2006,vajda_2002}, or information sufficient in the language of this paper. 
These results offer concrete conditions for finite-size representations that,  in light of Theorem \ref{th_1} (see also Theorem \ref{th_3}),  will be operationally sufficient for classification (details presented in Section \ref{subsec_information_sufficiient}). In this context, it is worth highlighting the family of data-driven partitions studied in \cite{silva_2010,silva_isit_2007,silva_2010b,vajda_2002,darbellay_1999,gonzales_2020}. These data-driven representations 
are information sufficient (with probability one) and, consequently, they are operationally sufficient (with probability one) as a direct corollary of Theorem \ref{th_1} (details presented in Section \ref{subsec_data_driven_representations}).

\subsection{Information Sufficient Partitions}
\label{subsec_information_sufficiient}
Here we revisit important results on asymptotic sufficient partitions to approximate the divergence (Kullback-Leibler divergence or information divergence) between two distributions in an abstract measurable space \cite{gray_1990_b,csiszar_2004}.   

\begin{lemma}\label{lemma_basic_sufficiency} [\cite{liese_2006}]
	Let us consider $P,Q$ probability measures in the measurable space $(\Omega, \mathcal{F})$ such that $D(P||Q)<\infty$. Let us consider $\left\{ \pi_n, n\geq 1 \right\}$ a family of embedded finite size measurable partitions of $\Omega$, in the sense that  $\sigma(\pi_1)\subset \sigma(\pi_2) \subset ....$.  Let us denote $\mathcal{S} \equiv \sigma (\pi_1\cup \pi_2,...)$\footnote{$\sigma(\mathcal{A})$ denotes the smallest sigma field that contains $\mathcal{A} \subset \mathcal{F}$ \cite{gray_2009}.},  then it follows that 
	\begin{equation}\label{eq_information_sufficiient_1}
		 \lim_{n \rightarrow \infty} D_{\sigma(\pi_n)} (P||Q) = D_{\mathcal{S}} (P||Q) \leq D(P||Q). 
	\end{equation} 
\end{lemma}
In this result, $D_{\sigma(\pi_n)} (P||Q) \equiv \sum_{A\in \pi_n} P(A) \log \frac{P(A)}{Q(A)}$
	denotes the divergence of $P$ with respect to $Q$ restricted over the sigma-field induced by $\pi_n$,
	and  
	 $$D_{\mathcal{S}}(P||Q) \equiv \sup_{\pi \in  \mathcal{Q}(\mathcal{S})} D_{\sigma(\pi)}(P||Q),$$
	 where $\mathcal{S}$ is a general sub-sigma field of $\Omega$ (i.e., $\mathcal{S} \subset \mathcal{F}$)\cite{gray_2009, breiman_1968}, $\mathcal{Q}(\mathcal{S})$ denotes the collection of measurable finite partitions in $\mathcal{S}$, and 
	 $$D(P||Q) \equiv D_{\mathcal{F}}(P||Q).$$
	 
By the assumptions of Lemma \ref{lemma_basic_sufficiency}, we have that $\mathcal{S}\subset \mathcal{F}$, then $ D_{\mathcal{S}} (P||Q) \leq  D (P||Q)$ in (\ref{eq_information_sufficiient_1}). Consequently, if there exists $\left\{ \pi_n, n\geq 1 \right\}$ with full precision in the sense that $\sigma (\pi_1\cup \pi_2,...)=\mathcal{F}$, this family of partitions is information sufficient for the divergence  in the sense that for any pair $P$, $Q$ such that $D(P||Q) <\infty$, it follows that 
\begin{equation}\label{eq_information_sufficiient_1b}
	\lim_{n \rightarrow \infty} D_{\sigma(\pi_n)} (P||Q) = D (P||Q)
\end{equation} 
from Lemma \ref{lemma_basic_sufficiency}.
A special case of the result in (\ref{eq_information_sufficiient_1b}) in our mixed continuous-discrete setting, $\Omega= \mathbb{R}^d\times \mathbb{Y}$,  
with the mutual information\footnote{$I(\mu_{X,Y}) = D(\mu_{X,Y} ||  \mu_{X} \mu_{Y})$ \cite{cover_2006,gray_1990_b}.} is  the following: 
\begin{lemma}\label{lm_MI_embedded_sufficient_partitions} [\cite{liese_2006}]
	Let us consider a joint vector $(X,Y)$ with probability $\mu_{X,Y}$ and a collection of finite embedded partitions $\left\{\pi_n, n\geq 1 \right\}$ in $\mathcal{B}(\mathbb{R}^d)$ such that $\sigma(\pi_1)\subset \sigma(\pi_2) \subset ....$. Then 
	\begin{equation}\label{eq_information_sufficiient_2}
		 \lim_{n \rightarrow \infty} \mathcal{I}_{\sigma(\pi_n)} (X;Y) = \mathcal{I}_{\mathcal{S}} (X;Y),  
	\end{equation} 
	where $\mathcal{S} \equiv  \sigma (\pi_1\cup \pi_2,...)\subset \mathcal{B}(\mathbb{R}^d)$. 
\end{lemma}

Then, we have the following corollary: 	
\begin{corollary}\label{cor_MI_embedded_sufficient_partitions}
	If $\mathcal{S} = \sigma(\pi_1\cup \pi_2,...) = \mathcal{B}(\mathbb{R}^d)$, then $\left\{\pi_n, n\geq 1 \right\}$ is information sufficient (Def. \ref{def_is}) in the sense that 
	\begin{equation}\label{eq_information_sufficiient_3}
		 \lim_{n \rightarrow \infty} \mathcal{I}_{\sigma(\pi_n)} (X;Y) = \mathcal{I}(X;Y),  
	\end{equation} 
	for any model $\mu_{X,Y}$ in $(\mathbb{R}^d\times \mathbb{Y}, \sigma(\mathcal{B}(\mathbb{R}^d) \times 2^{\mathbb{Y}}))$).
\end{corollary}
In this case 
$$\mathcal{I}_{\sigma(\pi)} (X;Y) = \sum_{A\in \pi,y\in \mathcal{Y}} \mu_{X,Y}(A \times \left\{ y \right\})  \frac{ \mu_{X,Y}(A \times \left\{ y \right\} }{ \mu_{X}(A) \mu_{Y}(\left\{ y \right\})},$$ 
which is equivalent to $\mathcal{I}(\mu_{U,Y})=I(U;Y)$
where $U=\eta_\pi(X)$ ($\eta_\pi(\cdot)$ is the finite representation induced by $\pi\in \mathcal{Q}(\mathcal{F})$ in  (\ref{eq_sec_continuous_2})).

\subsubsection{The Universal (distribution-free) Construction}
\label{subsec_universal_quantization}
We present a universal partition scheme introduced in \cite{liese_2006} for the observation space $(\mathbb{R}^d,\mathcal{B}(\mathbb{R}^d))$ that is information sufficient (distribution-free). The construction is the following:  
\begin{equation}\label{eq_subsec_inf_divergence_17b}
	\tilde{\pi}_m=  \left\{ B_{m,0} \right\} \cup \left\{B_{m,\bar{j}}, \bar{j}=(j_1,..,j_d) \in \mathcal{J}_m \right\}
\end{equation} 
where the index set is $\mathcal{J}_m = \left\{-m 2^m,...,m2^m-1\right\}^d$ and 
\begin{align}\label{eq_subsec_inf_divergence_17c}
	B_{m,0}&= \mathbb{R}^d \setminus [-m,m)^d, \\ 
	B_{m,j_1,...,j_d} &= \bigotimes^d_{k=1}  \left[\frac{j_k}{2^m},\frac{j_{k}+1}{2^m} \right), \  \forall (j_1,..,j_d)\in \mathcal{J}_m.
\end{align}
This construction is universal for the Borel sigma-field $\mathcal{B}(\mathbb{R}^d)$, as any interval can be approximated (arbitrarily closely) by the union  of cells of $\tilde{\pi}_m$ as $m$ goes to infinity \cite{liese_2006}.  Consequently,  we have that $\sigma(\cup_{m\geq 1} \tilde{\pi}_m)=\mathcal{B}(\mathbb{R}^d)$ \cite{liese_2006}. 

Then from Corollary \ref{cor_MI_embedded_sufficient_partitions} 
we have the following zero-information loss condition:
	\begin{equation}\label{eq_subsec_inf_divergence_18}
		 \lim_{m \rightarrow \infty} \mathcal{I} (X;Y| \eta_{\tilde{\pi}_m}(X)) = 0,  
	\end{equation} 
where the representation $\eta_{\tilde{\pi}_m}(\cdot)$  (encoder induced by $\tilde{\pi}_m$) is given by the mapping (quantizer)
	\begin{equation}\label{eq_subsec_inf_divergence_19}
		 \eta_{\tilde{\pi}_m}(x)= (m2^m,...,m2^m) \cdot {\bf 1}_{B_{m,0}}(x) + \sum_{\bar{j}  \in \mathcal{J}_m}   \bar{j} \cdot {\bf 1}_{B_{m,\bar{j}}}(x)
	\end{equation} 
from $\mathbb{R}^d$ to $\left\{ (m2^m,...,m2^m) \right\} \cup \mathcal{J}_m$ of size $m2^m+1$.  

\subsubsection{Final Remarks}
Therefore from (\ref{eq_subsec_inf_divergence_18}), there is a collection of finite-size partitions (of size $m2^m+1$) that asymptotically captures all the information of $(X,Y)$ in a distribution-free manner (independent of $\mu_{X,Y}$). In fact, Theorem \ref{th_1} and Corollary \ref{cor_MI_embedded_sufficient_partitions} imply that $\left\{\tilde{\pi}_n, n\geq 1 \right\}$ is operationally sufficient distribution-free: i.e., $\forall \mu_{X,Y}$,  $\lim_{n \rightarrow \infty} \ell(\mu_{U_n,Y})= \ell(\mu_{X,Y})$, where $U_n=\eta_{\tilde{\pi}_m}(X)$. 

In summary, this section shows that digitalization (in the form of a vector quantization of $X$) offers a universal  representation scheme 
with the capacity to retain an arbitrary amount of the information that $X$ has about $Y$ in the strong Shannon mutual information sense (Def. \ref{def_is}). 

\subsection{Data-Driven Partions}
\label{subsec_data_driven_representations}
The universal construction presented in the previous section is an interesting feasibility result; however, its construction is  distribution-free (universal) and data-independent. Complementing this feasibility analysis, here we present results that show that unsupervised data-driven representations of $\mathcal{X}$ (with the $X$-property \cite{devroye1996}) are also information sufficient (with probability one) and, consequently, they have the potential to provide a better tradeoff between complexity (size of the partition) and information loss for a given model $\mu_{X,Y}$.
Results presented in this section contextualize some work on data-driven partitions for mutual information estimation \cite{silva_2010,silva_2010b,silva_2012}. 
In addition,  we will show two concrete examples of discrete data-driven schemes that are information sufficient (Definition \ref{def_is}) and, consequently, they are also operationally sufficient for classifying $Y$ (in the MPE sense) in light of Theorem \ref{th_1}.

\subsubsection{Preliminaries}
Let us introduce the concept of data-dependent partition.  An $n$-sample partition rule $\pi_n(\cdot)$ is a mapping from the data space $(\mathbb{R}^d\times \mathcal{Y})^n$ to $\mathcal{Q}(\mathbb{R}^d)$, where $\mathcal{Q}(\mathbb{R}^d)$ denotes the space of finite size (measurable) partitions of $\mathbb{R}^d$.   A special case of this family is when $\pi_n$ has the $X$-property (unsupervised rule) \cite[Ch. 20.2]{devroye1996} meaning that $\pi_n(\cdot)$ is a mapping from the unsupervised data space $\mathbb{R}^{dn}$ to $\mathcal{Q}(\mathbb{R}^d)$. Finally, a scheme $\Pi= \left\{\pi_1(\cdot),\pi_2(\cdot),\ldots \right\}$ is a collection of rules of different lengths.  

\begin{definition}\label{information_sufficient_data_driven_partition}
	A partition scheme $\Pi$ with the $X$-property (unsupervised) is said to be information sufficient, if for any $\mu_{X,Y}\in \mathcal{P}(\mathcal{X} \times \mathcal{Y})$, it follows that 
	\begin{equation}\label{eq_subsec_data_driven_representations_1}
		\lim_{n \infty} I(\eta_{\pi_n(X_1,..,X_n)}(X); Y) = I(X;Y),  \ \text{with probability one}
	\end{equation}
	where $X_1,..,X_n$ are i.i.d. samples following $\mu_X$, and $\eta_{\pi_n(X_1,..,X_n)}(\cdot)$ is the finite-size representation induced by the data-driven (random) partition $\pi_n(X_1,..,X_n)$.\footnote{It is worth noting that $\pi_n(X_1,..,X_n)$ is a random partition in $\mathcal{Q}(\mathbb{R}^d)$, because it is a function of the random vector $X^n_1=(X_1,..,X_n)$.}
\end{definition}

Adopting Theorem \ref{th_1} in this context, we have the following:
\begin{corollary}\label{cor_th1_data_driven}
	Let $\Pi$ be a partition scheme with the $X$-property driven by the i.i.d. process $(X_n)_{n\geq 1}$ with $X_i\sim \mu_X$. If $\Pi$ is information sufficient (Def. \ref{information_sufficient_data_driven_partition}), then $ \left\{\eta_{\pi_n(X^n_1)}(\cdot) \right\}_{n\geq 1}$ is operationally sufficient (almost surely) in the sense that\footnote{The probability one is with respect to the process distribution of $(X_n)_{n\geq 1}$.}
	\begin{equation}\label{eq_subsec_data_driven_representations_2}
		\lim_{n  \rightarrow \infty} \ell(\mu_{U_n(X^n_1),Y}) = \ell(\mu_{X,Y}),  \ \text{with probability one}, 
	\end{equation}
	where $U_n(X^n_1) \equiv \eta_{\pi_n(X^n_1)}(X)$ is the (data-driven) representation of $X$, which is a function of the 
	unsupervised training data $X^n_1$.
\end{corollary}

\subsubsection{A Shrinking-cell Condition for Informational Sufficiency}
\label{sub_shrinking_cell}
Here we present a sufficient condition that guarantees that a scheme with the $X$-property is information sufficient (Def. \ref{information_sufficient_data_driven_partition}). Before that, we need to introduce a few definitions. The diameter of an event $B\subset \mathcal{B}(\mathbb{R}^d)$ is 
 \begin{equation} \label{eq_sub_shrinking_cell_0}
 	diam(B) \equiv \sup_{x,y \in B} \left|\left|x-y\right|\right|, 
 \end{equation}
where $\left|\left|\cdot\right|\right|$ is the Euclidian norm in $\mathbb{R}^d$. Considering a partition rule $\pi_n:\mathbb{R}^{dn} \rightarrow \mathcal{Q}(\mathbb{R}^d)$, a point $z \in \mathbb{R}^d$ and data-sequence $x^n_1\in \mathbb{R}^{dn}$,  we use $\pi_n(z|x^n_1)$ to denote the cell in $\pi_n(x^n_1)$ that contains $z$.

\begin{lemma}\label{lm_shrinking_cell}(\cite[Th. 2]{silva_2010})
	Let us consider $\mu_{X,Y}$ and  $\Pi= \left\{\pi_1(\cdot),\pi_2(\cdot),\dots \right\}$ a partition scheme  with the $X$-property driven by $X_1, X_2,...,$  where $X_i\sim \mu_X$ for any $i\geq 1$. If $\mu_X$ has a density function\footnote{$\mu_X$ is absolutely continuous with respect to the Lebesgue measure in $(\mathbb{R}^d, \mathcal{B}(\mathbb{R}^d))$.} and 
	$\Pi$ satisfies that for any $\delta>0$
	\begin{equation}\label{eq_sub_shrinking_cell_1}
		\lim_{n  \rightarrow \infty} \mu_{X}\left(  \left\{x\in \mathbb{R}^d, diam(\pi_n(x|X^n_1)) > \delta \right\} \right) = 0,  \text{ with probability one},
	\end{equation}
	then 
	\begin{equation}\label{eq_sub_shrinking_cell_2}
		\lim_{n  \rightarrow \infty} I(\eta_{\pi_n(X_1,..,X_n)}(X); Y) = I(X;Y),  \ \text{with probability one}.
	\end{equation}
\end{lemma}
This result derives from the proof of  \cite[Th. 2]{silva_2010}.

Lemma \ref{lm_shrinking_cell} offers a {\em shrinking-cell sufficient condition}  (in (\ref{eq_sub_shrinking_cell_1})) for information sufficiency (Def. \ref{def_is}). In particular, this result states that if the diameter of the random  partition $(\pi_n(X^n_1))_{n\geq 1}$ tends to zero in a specific probabilistic sense,  this high-resolution condition implies that its  discrete representations are information sufficient in (\ref{eq_sub_shrinking_cell_2}). Different flavors of this high-resolution condition have been presented in the literature of learning \cite{lugosi_1996_ann_sta, devroye1996,nobel_1996b}. The one adopted in this work comes from results on histogram-based estimation for information measures \cite{silva_2010,silva_2012,silva_2010b}.

The next subsections present two constructions that satisfy the $X$-property (unsupervised representations learned from data)
and are information sufficient (Def. \ref{information_sufficient_data_driven_partition}). 
\subsubsection{Statistically Equivalent Blocks}
\label{stat_equiv_blocks}
Here, we present a scheme implementing the principle of statistically equivalent partition \cite{devroye1996,gessaman_1970}.
 Let $X_1,..,X_n$ be i.i.d. samples of $\mu_X \in \mathcal{P_X}$ for which we assume that $\mu_X\ll \lambda$. The idea is to partition the space $\mathbb{R}^d$ by axis-parallel hyperplane in such a way that at the end of 
the process we have almost the same number of sample points per cell. For that, let $l_n$ be the number of samples (a non zero integer) that ideally we want to have at the end of the process in every cell of $\pi_n$. The method chooses an arbitrary order of the axis-coordinate, let us say the order $(1,2,..,d)$, and considers $T_n= \lfloor (n/l_n)^{1/d} \rfloor$ as the number of partitions to produce in every axis. The method goes as follows: choose the first coordinate and project the data in that direction $Z_1,..,Z_n \in \mathbb{R}$; compute the order statistics that we denote by $Z^{(1)}<....<Z^{(n)}$. From this sequence,
define the following axis-parallel partition of the real line:
\begin{align}\label{eq_stat_equiv_blocks_1}
&\left\{I^1_i:i =1,..,T_n\right\}=
\left\{(-\infty,Z^{(s_n)}],(Z^{(s_n)},Z^{(2\cdot s_n)}],..,(Z^{\left((T_n-1)\cdot s_n\right)},\infty)\right\} \subset \mathbb{R}, 
\end{align}
where $s_n=\lfloor n/T_n \rfloor$. Then assigning the vector samples $X_1,..,X_n$ to the cells of $\pi^{(1)}_n=  \left\{I^1_i \times \mathbb{R}^{d-1}, i=1,..,,T_n \right\}$ concludes the first iteration that produces $\pi^{(1)}_n$. The second iteration applies the same principle (statistically equivalent partition with axis-parallel hyperplanes) over the cells of $\pi^{(1)}_n$  but in the second coordinate, for which the original samples $X_1,..,X_n$ are assigned to each individual cell of $\pi^{(1)}_n$, accordingly.  At the end of the second iteration, we produce $\pi^{(2)}_n$. Iterating this algorithm until the last coordinate, $d$, we obtain $\pi^{(d)}_n(X^n_1)$. Importantly, it can be shown that for any cell $A\in \pi^{(d)}_n(X^n_1)$ the number of training samples that belong to this cell is greater than or equal to $l_n$ \cite{silva_2010}. This (data-driven) assignment is critical to derive the following result:
\begin{lemma}\label{lm_stat_bllock_shrinking_cell} \cite[Th.4]{silva_2010}
	Let $\mu_X$ be a continuous probability in $\mathbb{R}^d$ ($\mu_X \ll \lambda$) 
	and let $(X_n)_{n\geq 1}$ be i.i.d. samples driven by $\mu_X$.  If $(l_n)$ is $o(n)$,\footnote{$(a_n)$ being $o(b_n)$ means that $\lim_{n \rightarrow \infty}\frac{a_n}{b_n}=0$.} it follows that for any $\delta>0$
	\begin{equation}\label{eq_stat_equiv_blocks_2}
		\lim_{n  \rightarrow \infty} \mu_{X}\left(  \left\{x\in \mathbb{R}^d, diam(\pi^{(d)}_n(x|X^n_1)) > \delta \right\} \right) = 0,  	\text{ with probability one},
	\end{equation}
	which implies from Lemma \ref{lm_shrinking_cell} that the scheme $\Pi=  \left\{  \pi^{(d)}_n, n\geq 1 \right\}$ is information sufficient (Def. \ref{information_sufficient_data_driven_partition}).
\end{lemma}
The argument to prove Lemma \ref{lm_stat_bllock_shrinking_cell} derives from the proof of  \cite[Th.4]{silva_2010}.

\subsubsection{Balanced Search Tree}
\label{subsec_tree_partition}
Here we present a version of a balanced search tree \cite[Ch. 20]{devroye1996}, in particular the binary case introduced in \cite[Sec.V.B]{silva_2010} for the problem of MI estimation. Given $X_1,...,X_n$ i.i.d. samples (unsupervised data) driven by $\mu_X$, the partition rule selects a dimension (or coordinate) of the set $\left\{1,..,d \right\}$ in a given sequential order, let us say the $i$-coordinate, and the axis-parallel hyperplane 
\begin{equation}\label{eq_subsec_tree_partition_1}
		H_i(X^n_1)= \left\{{x\in \mathbb{R}^d:x(i)\leq X^{(\lceil{n/2}\rceil)}(i)} \right\},
\end{equation}
where $X^{(1)}(i) < X^{(2)}(i) < ,..,< X^{(n)}(i)$ is the order statistics of $X^n_1$ projected over the coordinate $i$.
Then we create a binary partition of $\mathbb{R}^d$ given by $\bar{\pi}_n^{(1)}=  \left\{ H_i(X^n_1), H_i(X^n_1)^c\right\}$.
Assigning $X^n_1$ to its respective cells in $\bar{\pi}_n^{(1)}$, the process continues with the following coordinate, let us say the  $j$-coordinate,  in the mentioned sequential order applying the median statistically equivalent principle in (\ref{eq_subsec_tree_partition_1}) in each cell of $\bar{\pi}_n^{(1)}$ by projecting the data over the $j$-coordinate. Importantly, the axis-parallel binary partition in (\ref{eq_subsec_tree_partition_1}) is conduced if the number of samples associated with the resulting cells is greater than or equal to $l_n$ (a positive parameter of the method); otherwise, the algorithm stops the splitting process for this cell. $l_n$ is designed  to guarantee that the statistically equivalent splitting approach has a sufficient number of samples\footnote{A systematic exposition of the statistical properties of this scheme and its implementation and use can be found in \cite{devroye1996} for pattern recognition and in \cite{silva_2010,silva_2010b} for estimating  information measures.}. Therefore, after iterating this principle and meeting the stopping criterion in every resulting cell, we have a partition $\bar{\pi}^{(l_n)}_n$ with a binary tree-structure that has almost the same number of samples in every cell (balanced), and in addition the number of samples per cell is lower bounded  by $l_n$.
The stopping criterion (parametrized by $l_n$) and the statistically equivalent principle of this scheme are critical to prove the following result (the shrinking-cell condition):
\begin{lemma}\label{lm_balanced_shrinking_cell} \cite[Th.5]{silva_2010}
	Let $\mu_X$ be a continuous probability on $\mathbb{R}^d$  ($\mu_X \ll \lambda$) and let $(X_n)_{n\geq 1}$ be i.i.d. samples driven by $\mu_X$. If $(l_n)$ is $\mathcal{O}(n^p)$ with $p\in (0,1)$, it follows that for any $\delta>0$
	\begin{equation}\label{eq_subsec_tree_partition_2}
		\lim_{n  \rightarrow \infty} \mu_{X}\left(  \left\{x\in \mathbb{R}^d, diam(\bar{\pi}^{(l_n)}_n(x|X^n_1)) > \delta \right\} \right) = 0,  	\text{ with probability one},
	\end{equation}
	which implies from Lemma \ref{lm_shrinking_cell} that  the scheme $\bar{\Pi} =  \left\{  \pi^{(l_n)}_n, n\geq 1 \right\}$ is information sufficient (Def. \ref{information_sufficient_data_driven_partition}).
\end{lemma}
The argument to prove Lemma \ref{lm_balanced_shrinking_cell} derives from the proof of  \cite[Ths.5 and 6]{silva_2010}.

\subsubsection{Remarks}
\label{sub_sec_data_driven_represent_final_remarks}
In conclusion, we show a general condition to meet IS for data-driven partitions and two concrete schemes that meet this 
strong IS requirement and that in light of Corrolary \ref{cor_th1_data_driven} (from Theorem \ref{th_1}) are OS for classification.  

\section{Weak Informational Sufficiency (WIS) for a Class of Models}
\label{sec_wis_applied_in_learning}
As we discussed in Section \ref{sub_sec_finite_asymtotic}, the WIS condition on $\left\{ \eta_i(\cdot)\right\}_{i\geq 1}$ used in Theorem \ref{th_1} cannot be adopted as a criterion in a learning setting. The reason for this is that the reference representation $\tilde{U}$ used in Definition \ref{def_wis} is a function of the true model $\mu_{X,Y}$, which is unavailable in  learning. Then, adapting this pure WIS condition in learning, for instance for the task of learning representations from data, is not possible. 

To give  practical significance to Theorem \ref{th_1} in a learning setting, in this section we move in the direction of considering a family of indexed models $\Lambda =  \left\{\mu^\theta_{X,Y}, \theta\in \Theta \right\} \subset \mathcal{P}(\mathcal{X} \times \mathcal{Y})$ (or hypotheses) to formalize on $\Lambda$ the structure of a learning problem. More precisely, we assume that the unknown model $\mu_{X,Y}$ belongs to $\Lambda$, which can be seen as a form of prior knowledge. The objective here is to use this prior knowledge to come up with an information sufficient condition that implies OS but is strictly weaker than pure IS (in Definition \ref{def_is}). Importantly, this new condition will be derived from Theorem \ref{th_1} and specific assumptions on the structure (or richness) of $\Lambda$.

\subsection{Formalizing Operational Structure}
\label{subsec_operational_structure}
For any $\mu_{X,Y}^\theta \in \Lambda$, let us consider its {\em MPE decision rule} given by\footnote{This rule is not unique. In general, we could select any function that is a solution of (\ref{eq_wis_learning_1}) and achieves the MPE.}: 
\begin{equation}\label{eq_wis_learning_1}
	r_{\theta}(x) \equiv  \arg \max_{y\in \mathcal{Y}} \mu^\theta_{Y|X}(y|x)
\end{equation}
and its induced optimal (in MPE sense) partition:
\begin{equation}\label{eq_wis_learning_2}
	\pi^{*,\theta} \equiv    \left\{{r_{\theta}}^{-1}(\left\{ y \right\}), y \in \mathcal{Y} \right\} \subset \mathcal{B}(\mathbb{R}^d). 
\end{equation}
Importantly, we can introduce 
\begin{equation}\label{eq_wis_learning_3}
	\sigma(\Lambda) \equiv  \bigcap_{\theta\in \Theta} \sigma(\pi^{*,\theta})
\end{equation}
to be the smallest sub-sigma field that makes all the decision rules $ \left\{r_{\theta} (\cdot), \theta \in \Theta \right\}$ of $\Lambda$ measurable from $\mathcal{X}$ to $\mathcal{Y}$. In (\ref{eq_wis_learning_3}), $\sigma(\pi^{*,\theta})$ denotes the sigma field induced by the finite size partition $\pi^{*,\theta}$ \cite{gray_2004}. $\sigma(\Lambda)$ is capturing the richness of $\Lambda$, where by construction it follows that $\sigma(\Lambda)\subset \mathcal{B}(\mathbb{R}^d)$. The object in (\ref{eq_wis_learning_3}) can be seen as an operational form of structure, in the sense that $\sigma(\Lambda)$  is the smallest $\sigma$-field that makes all MPE rules in (\ref{eq_wis_learning_1}) measurable and, consequently, it is a function of both $\Lambda$ and the operational (decision) problem in (\ref{eq_sec_II_1}). In general, the smaller (or the simpler) $\sigma(\Lambda)$ is relative to the Borel sigma field of  $\mathbb{R}^d$, the more (prior) information we have from assuming that our model belongs to $\Lambda$. 

\subsection{Finite-Size Families}\label{subsec_finite_size_class_models}
An important case to consider is when $\Lambda$ has an intrinsic finite-size (discrete) structure of size $K>0$. This happens when $\bigcap_{\theta\in \Theta} \pi^{*,\theta}$ forms a partition of finite size (let say $K>0$) of $\mathbb{R}^d$ that we denote by $\pi=\left\{A_i, i=1,..,K \right\}\subset \mathcal{B}(\mathbb{R}^d)$ where,  
consequently, $\sigma(\Lambda) = \sigma(\pi)$. Under this finite-size  assumption, we can construct a finite-size operationally sufficient representation for the whole family $\Lambda$. In particular, we can choose a prototype $p_i\in A_i$ for every cell of $\pi$ and the following lossy mapping (vector quantizer): $\forall x\in \mathcal{X}$
\begin{equation}\label{eq_wis_learning_4}
	\eta(x) \equiv \sum_{j=1}^K \mathbf{1}_{A_j}(x) \cdot p_j \in \mathbb{R}^d.
\end{equation}
Importantly, from (\ref{eq_wis_learning_2}), (\ref{eq_wis_learning_3}) and the construction $\eta(\cdot)$ in (\ref{eq_wis_learning_4}), it follows that for any $\theta \in \Theta$
\begin{equation}\label{eq_wis_learning_5}
	r_{\theta}(x) = r_{\theta}(\eta(x))	,\ \forall x\in \mathbb{R}^d.
\end{equation}
Therefore,  all the MPE rules on our class of models are insensitive to this lossy operator $\eta(\cdot)$.
From this invariant (robustness) property in (\ref{eq_wis_learning_5}), we have the following representation result: 
\begin{proposition}\label{pro_OS_from_invariance}
	Let us consider a lossy mapping $\eta: \mathcal{X} \rightarrow \mathcal{X}$. If the family $\Lambda$
	is invariant with respect to  $\eta(\cdot)$, in the sense of (\ref{eq_wis_learning_5}), then for any $\mu^\theta_{X,Y} \in \Lambda$
	\begin{equation}\label{eq_wis_learning_6}
		\ell(\mu^\theta_{\eta(X),Y}) = \ell(\mu^\theta_{X,Y}).
	\end{equation} 
\end{proposition}
\begin{proof} The proof derives from the observation that the condition in (\ref{eq_wis_learning_5}) implies that 
$r_{\theta}(\cdot)$ is a deterministic function of $\eta(\cdot)$ for any $\theta \in \Theta$, and the fact that by construction 
$r_{\theta}(\cdot)$ is operationally sufficient for $\mu^\theta_{X,Y}$ (i.e., $\ell(\mu^\theta_{r_{\theta}(X),Y}) = \ell(\mu^\theta_{X,Y})$).
\end{proof}

Proposition \ref{pro_OS_from_invariance} illustrates an operational structure intrinsic in $\Lambda$, where 
there is a lossy representation (vector quantizer of size $K$) of $X$ in (\ref{eq_wis_learning_4}) that is operationally sufficient for all members of the family. 
This (intrinsic) lossy structure for $\Lambda$ is captured more generally in $\sigma(\Lambda)$, as long as $\sigma(\Lambda)$ is  strictly contained in $\mathcal{B}(\mathbb{R}^d)$. 

The general idea of this section is to use this structure (and the existence of a lossy mapping satisfying (\ref{eq_wis_learning_6})) to extend Theorem \ref{th_1} in this learning setting. This will be the focus of the next subsections. 

\subsection{The Main Result}\label{sub_sec_wis_applied_in_learning_result}
Here, we present a weak form of informational sufficiency for a family of models $\Lambda \subset \mathcal{P}(\mathcal{X} \times \mathcal{Y})$ that implies operational sufficiency. Before that, let us introduce the following definition:

\begin{definition}\label{def_oper_suff_family}
	We say that a lossy mapping $\eta: \mathcal{X} \rightarrow \mathcal{X}$ is operationally sufficient (OS) 
	for $\Lambda =  \left\{\mu^\theta_{X,Y}, \theta\in \Theta \right\}$ if $H(r_\theta(X)|\eta(X))=0$ when $X\sim \mu^\theta_X$ for any $\theta \in \Theta$.
\end{definition}
Definition \ref{def_oper_suff_family} implies that $r_\theta(\cdot)$ is a deterministic function of $\eta(\cdot)$ for any $\theta\in \Theta$ ($\mu^\theta_X$ almost surely) and, consequently,  we have that $\ell(\mu^\theta_{\eta(X),Y}) = \ell(\mu^\theta_{X,Y})$, from the same argument used to prove Proposition \ref{pro_OS_from_invariance}.
Then, we can state the following result: 
\begin{theorem}\label{th_wis_over_a_family}
	Let $\left\{ U_i \right\}_{i\geq 1}$  be  a family of continuous representations 
	of $X$ obtained from the mappings $\left\{\eta_i\right\}_{i\geq 1}$, and let us assume that there is a lossy mapping $\eta^*: \mathcal{X} \rightarrow \mathcal{X}$ that is OS 
	for $\Lambda$ (see Def. \ref{def_oper_suff_family}). If for any of the models $\mu^\theta_{X,Y} \in \Lambda$
	\begin{equation}\label{eq_wis_learning_7}
		\lim_{i \rightarrow \infty}  \underbrace{\mathcal{I}(\mu^\theta_{(\eta^*(X),U_i),Y}) - \mathcal{I}(\mu^\theta_{U_i,Y})}_{I(\eta^*(X);Y|\eta_i(X))} = 0,
	\end{equation} 
	then 
	\begin{equation}\label{eq_wis_learning_8}
		\lim_{i \rightarrow \infty} \ell(\mu^\theta_{\eta_i(X),Y}) = \ell(\mu^\theta_{X,Y}).
	\end{equation} 
\end{theorem}
The proof is presented in Section \ref{proof_th_wis_over_a_family}.

Analysis and interpretation of Theorem \ref{th_wis_over_a_family}:
\begin{enumerate}	
	\item Theorem \ref{th_wis_over_a_family} can be considered as a non-oracle extension of Theorem \ref{th_1} because we do not assume that we know the true model in (\ref{eq_wis_learning_7}) to establish the sufficient condition to achieve OS in (\ref{eq_wis_learning_8}), but rather we asume that the true model belongs to a class with a non-trivial structure represented by $\eta^*(\cdot)$. 
	
	\item Here, we use $\eta^*(\cdot)$ (known a priori for the class $\Lambda$) to verify a sufficient condition for the representations $\left\{\eta_i (\cdot) \right\}_{i\geq 1}$ that is strictly weaker (if $\eta^*(\cdot)$ is a lossy mapping) than IS in Definition \ref{def_is}. 
	
	\item To make this result truly relevant, it is critical to find a 
	lossy mapping $\eta^*$ that is OS for a class $\Lambda \subset \mathcal{P}(\mathcal{X} \times \mathcal{Y})$ or ideally to find the simplest mapping of $\Lambda$ that satisfies the condition in Definition \ref{def_oper_suff_family}. This last representation problem for a class $\Lambda$ is not evident in principle. On the existence of a lossy mapping for $\Lambda$,  Section \ref{sub_sec_wis_applied_in_learning_cases} addresses this problem by illustrating some relevant examples.
	
	\item Complementing the previous point, one might note that the identity function from $\mathcal{X}$ to $\mathcal{X}$ is operationally sufficient for any class of models  $\Lambda \subset \mathcal{P}(\mathcal{X} \times \mathcal{Y})$.  Using this trivial function in (\ref{eq_wis_learning_7}), this condition reduces to verify the pure information sufficient condition in Definition \ref{def_is}, which recovers the result that if $\left\{ U_i \right\}_{i\geq 1}$ IS for $\mu^\theta_{X,Y}$ then $\left\{ U_i \right\}_{i\geq 1}$ is OP for $\mu^\theta_{X,Y}$.
	
	\item Importantly, the finite-size family introduced in Section \ref{subsec_finite_size_class_models}  has a non-trivial lossy function in (\ref{eq_wis_learning_4}) that meets the requirement of being OS (Def. \ref{def_oper_suff_family}) for $\Lambda$ and, consequently, Theorem \ref{th_wis_over_a_family} applies in this case in a non-trivial way. More details about this class of models can be found in Section \ref{sub_finite_size:revisiting}. 
	
	\item Finally, the proof of this result is constructed from the WIS condition of Theorem \ref{th_1}.
\end{enumerate}

\subsection{Application of Theorem \ref{th_wis_over_a_family}} 
\label{sub_sec_wis_applied_in_learning_cases}
At this point, we could use $\sigma(\Lambda)$ to determine a lossy mapping $\eta(\cdot)$ that meets Definition \ref{def_oper_suff_family} for $\Lambda$. Along this line, we revisit the case of finite-size families and introduce a class of models that satisfies some invariant properties to illustrate two relevant cases where Theorem \ref{th_wis_over_a_family} applies.

\subsubsection{Finite Size Families}\label{sub_finite_size:revisiting}
In the special case when $\Lambda$ has an intrinsic finite-size structure (see Section \ref{subsec_finite_size_class_models}), we have a lossy quantizer $\eta (\cdot)$ in (\ref{eq_wis_learning_4}) that satisfies the conditions in (\ref{eq_wis_learning_5}) and, consequently, is operationally sufficient for $\Lambda$ (Definition \ref{def_oper_suff_family}). Therefore, Theorem \ref{th_wis_over_a_family} applies in this case using the finite-size vector quantizer in (\ref{eq_wis_learning_4}) as the sufficient representation for the class $\Lambda$ in (\ref{eq_wis_learning_7}).  In this case, condition (\ref{eq_wis_learning_7}) is strictly weaker than pure IS and, consequently, the result in Theorem \ref{th_wis_over_a_family} is meaningful.  

A simple example of this class of problems is when $\mu_{X,Y}$ belongs to a finite class of hypotheses $\Lambda=\left\{\mu^i_{X,Y}, i=1,..,L \right\}$, where it is direct to show from (\ref{eq_wis_learning_3}) that  $\sigma(\Lambda)=\sigma(\pi)$ where $\pi \equiv \bigcap_{i=1}^L \pi^{*,i}$ and $\pi^{*,i}$ is the $M$-size partition induced  by the MPE decision rule of $\mu^i_{X,Y}$ in (\ref{eq_wis_learning_2}), and, consequently, $\Lambda$ has an intrinsic finite size $K \leq M^L < \infty$.

\subsubsection{Invariant Models}\label{sub_invariant_models}
Another interesting class of models to consider in this analysis is when $\Lambda$ has some invariances and probabilistic symmetries \cite{Bloem_2019}. We will consider the case where $\Lambda$ is invariant with respect to the action of a compact group $\mathcal{G}$ of measurable transformations on $(\mathcal{X},  \mathcal{B}(\mathcal{X}))$\footnote{A compact group $\mathcal{G}$  acting on $\mathcal{X}$ is a collection of Borel measurable functions from $(\mathcal{X}, \mathcal{B}(\mathcal{X}))$ to $(\mathcal{X}, \mathcal{B}(\mathcal{X}))$ such that: for any pair $g,h\in \mathcal{G}$,  $g \circ h\in \mathcal{G}$; for any $g\in \mathcal{G}$,  $g^{-1}\in \mathcal{G}$; and the identity mapping belongs to $\mathcal{G}$ \cite{eaton_1889,rotman_1995}.}. For this purpose, we introduce a new form of operational invariances of $\Lambda$ relative to the action of a group $\mathcal{G}$, which matches the operationally sufficient conditions needed to apply Theorem \ref{th_wis_over_a_family} in this context.

\begin{definition}\label{def_inv_rules} (Functional invariance)
	A measurable transformation $f:(\mathcal{X}, \mathcal{B}(\mathcal{X}))  \rightarrow  (\mathcal{U}, \mathcal{B}(\mathcal{U}))$ is $\mathcal{G}$-invariant if for any $g\in \mathcal{G}$
	\begin{equation*}
		f \circ g (x) = f(x), \ \forall x \in \mathcal{X}.
	\end{equation*}
\end{definition}

\begin{definition}\label{def_inv_models} (Operational invariance of a model)
	A model $\mu_{X,Y}\in \mathcal{P}(\mathcal{X} \times \mathcal{Y})$ is said to be operational invariant with respect to $\mathcal{G}$ (in short $\mathcal{G}$-invariant) if  there is a MPE decision rule,  solution of (\ref{eq_wis_learning_1}), which is $\mathcal{G}$-invariant in the sense of Def. \ref{def_inv_rules}.
\end{definition}

\begin{definition}\label{def_inv_classes}(Operational invariance of a class)
	A class $\Lambda= \left\{\mu^\theta_{X,Y}, \theta\in \Theta \right\} \subset \mathcal{P}(\mathcal{X} \times \mathcal{Y})$ is said to be operational invariant with respect to $\mathcal{G}$ (in short $\mathcal{G}$-invariant) if $\mu^\theta_{X,Y}$ is $\mathcal{G}$-invariant (Def. \ref{def_inv_models}) for any $\mu^\theta_{X,Y} \in \Lambda$. 
\end{definition}

{\bf A lossy mapping obtained from $\mathcal{G}$:}
Let us consider the orbit of $\mathcal{G}$ at the point $x\in \mathcal{X}$ given by 
$$\mathcal{G}(x) \equiv \left\{g(x), g \in \mathcal{G} \right\} \in \mathcal{B}(\mathcal{X}).$$   
We can induce an equivalent relationship in $\mathcal{X}$ where $x \longleftrightarrow y$ if $\mathcal{G}(x)=\mathcal{G}(y)$ and from this a measurable partition of $\mathcal{X}$ given by $\pi_\mathcal{G} \equiv \left\{ \mathcal{G}(x), x \in \mathcal{X} \right\}  \subset \mathcal{B}(\mathcal{X})$. $\pi_\mathcal{G}$ is the collection of orbits induced by the application of $\mathcal{G}$ in every point of $\mathcal{X}$. Importantly, there exists a measurable cross section $\mathcal{C}\subset \mathcal{X}$ of $ \mathcal{G}$ \cite{eaton_1889}\footnote{A systematic exposition of this result can be found in \cite{Bloem_2019} and references therein.}, which is a collection of prototypes for every orbit of $\pi_\mathcal{G}$ satisfying that $\forall x\in \mathcal{X}$
\begin{equation}
    \mathcal{C} \cap \mathcal{G}(x) \text{ is a singleton (i.e. $\mathcal{C}$ selects one prototype for every cell of $\pi_\mathcal{G}$)},
\end{equation}
and we denote this element by $\mathcal{C}(x)$. Then, we can construct the following lossy mapping $\eta^*_\mathcal{G}(x) \equiv \mathcal{C}(x)$ from $\mathcal{X}$ to $\mathcal{C}\subset \mathcal{X}$ similar to  $\eta(\cdot)$ in (\ref{eq_wis_learning_4}). From construction, $\eta^*_\mathcal{G}(\cdot)$ is Borel measurable and $\mathcal{G}$-invariant (Def. \ref{def_inv_rules}). Importantly, $\eta^*_\mathcal{G}(\cdot)$ is also {\em maximal-invariant} in the sense that for any pair $x,y\in \mathcal{X}$ that belongs to different orbits of $\pi_\mathcal{G}$ (i.e., $\mathcal{G}(x) \cap \mathcal{G}(y)= \emptyset$) then $\eta^*_\mathcal{G}(x) \neq \eta^*_\mathcal{G}(y)$. This last discrimination property over disjoint orbits of $\pi_{\mathcal{G}}$ is central to show that $\eta^*_\mathcal{G}(\cdot)$ is operationally sufficient (Def. \ref{def_oper_suff_family}) for the collection of $\mathcal{G}$-invariance models (see Def \ref{def_inv_classes}):
\begin{proposition}\label{prop_representation_for_invariant_models}
	Let $\Lambda = \left\{\mu^\theta_{X,Y}, \theta\in \Theta \right\} $ be $\mathcal{G}$-invariant (Def. \ref{def_inv_classes}) w.r.t. a compact measurable group $\mathcal{G}$ acting on $(\mathcal{X}, \mathcal{B}(\mathcal{X}))$. Then $\eta^*_\mathcal{G}(\cdot)$ is OS for $\Lambda$ (Def. \ref{def_oper_suff_family}) meaning that $\forall \theta\in \Theta$ there is $r_\theta()$, solution of (\ref{eq_wis_learning_1}), such that  $H(r_\theta(X)|\eta^*_{\mathcal{G}}(X))=0$ when $X\sim \mu^\theta_X$. 
\end{proposition}
The proof is presented in Appendix \ref{proof_prop_representation_for_invariant_models}.\footnote{The proof presented in  Appendix \ref{proof_prop_representation_for_invariant_models} is made for $\eta^*_{\mathcal{G}}(\cdot)$ but can be extended to any maximal-invariant mapping.}

The proof of Proposition \ref{prop_representation_for_invariant_models} also shows the sufficient property of $\eta^*_\mathcal{G}(\cdot)$ over $\Lambda$, in the sense that $\forall \theta\in \Theta$ 
	\begin{equation*}
		\ell(\mu^\theta_{\eta^*_\mathcal{G}(X),Y}) = \ell(\mu^\theta_{X,Y}).
	\end{equation*}
More  importantly, we have the following:
\begin{theorem}\label{th_wis_for_invariant_models}
	Let $\Lambda = \left\{\mu^\theta_{X,Y}, \theta\in \Theta \right\} $ be operational $\mathcal{G}$-invariant (Def. \ref{def_inv_classes}), and let  $\left\{ U_i \right\}_{i\geq 1}$  be a family of continuous representations 
	of $X$ obtained from $\left\{\eta_i (\cdot)\right\}_{i\geq 1}$. For any $\mu^\theta_{X,Y} \in \Lambda$, if 
	\begin{equation}\label{eq_wis_learning_9}
		\lim_{i \rightarrow \infty}  \underbrace{\mathcal{I}(\mu^\theta_{(\eta^*_\mathcal{G}(X),U_i),Y}) - \mathcal{I}(\mu^\theta_{U_i,Y})}_{I(\eta^*_\mathcal{G}(X);Y|\eta_i(X)) \text{ when }(X,Y)\sim \mu^\theta_{X,Y}} = 0,
	\end{equation} 
	then 
	\begin{equation}\label{eq_wis_learning_10}
		\lim_{i \rightarrow \infty} \ell(\mu^\theta_{\eta_i(X),Y}) = \ell(\mu^\theta_{X,Y}).
	\end{equation} 
\end{theorem} 
\begin{proof} Follows from Proposition \ref{prop_representation_for_invariant_models} and Theorem \ref{th_wis_over_a_family}. 
\end{proof}

\subsubsection{$\mathbb{S}_d$-invariant models}
\label{sub_sec_exchangable_models}
An important class of models is the one that is invariant to the action of permutations 
of the coordinates of $\mathcal{X}=\mathbb{R}^d$ \cite{Bloem_2019}.\footnote{This class was systematically studied in the excellent paper by Bloem-Reddy and Teh \cite{Bloem_2019}.}  We use this rich class to illustrate the relevant
use of Theorem \ref{th_wis_for_invariant_models}.  In this case, the compact group $\mathcal{G}$ is denoted
by $\mathbb{S}_d$ where for any $g\in \mathbb{S}_d$ there is a permutation of $[d]=\left\{1,,.,d\right\}$ $p:[d] \rightarrow [d]$
such that $g({\bf x})=(x_{d(1)},x_{d(2)},..,x_{d(d)})$  $\forall {\bf x} \in \mathbb{R}^d$. Therefore, if a function $f(\cdot)$ is 
$\mathbb{S}_d$-invariant (see Def. \ref{def_inv_rules}), it means that its output is invariant to the action of any permutation of ${\bf x}=(x_1,..,x_d)$, and therefore $f({\bf x})$ depends on the set $\left\{x_1,..,x_d\right\} \subset \mathbb{R}$ induced by ${\bf x}$.\footnote{A complete characterization of this family of permutation invariant functions is presented in  
\cite{zaheer_2017} and revisited and extended for a family of probabilistic models in \cite{Bloem_2019}.} 

Here we consider the family of operational $\mathbb{S}_d$-invariant models $\mathcal{P}^{\mathbb{S}_d}\subset \mathcal{P}(\mathcal{X} \times \mathcal{Y})$ from Definition \ref{def_inv_classes}. For this group, it is well known that the {\em empirical distribution} $\mathcal{M}:\mathbb{R}^d \rightarrow \mathcal{P}(\mathbb{R})$ \footnote{$\mathcal{M}({\bf x}) = \frac{1}{d} \sum_{i=1}^d \delta_{x_i}(\cdot) \in \mathcal{P}(\mathbb{R})$ denotes the empirical distribution induced by ${\bf x}$ in $(\mathbb{R}, \mathcal{B}(\mathbb{R}))$.} is invariant to the actions of $\mathbb{S}_d$, but, more importantly (for the adoption of  Theorem \ref{th_wis_for_invariant_models}) this lossy mapping is {\em maximal-invariant} for $\mathbb{S}_d$ \cite{Bloem_2019}.  Consequently, we could consider $\mathcal{M}(\cdot)$ as the maximal-invariant mapping $\eta^*_{\mathbb{S}_d}(\cdot)$ of $\mathbb{S}_d$ in the statement of Theorem \ref{th_wis_for_invariant_models}, where this result applies naturally for the family of $\mathbb{S}_d$-invariant models.

\subsubsection{Concluding Remarks}

\begin{remark} It is important to mention that our definition of operational invariances (Def. \ref{def_inv_classes}) for $\Lambda$ only implies that $\eta^*_{\mathcal{G}}(\cdot)$ is operationally sufficient (i.e., the requirement that $\ell(\mu^\theta_{\eta^*_\mathcal{G}(X),Y}) = \ell(\mu^\theta_{X,Y})$ for any $\mu^\theta_{X,Y} \in \Lambda$). This operational condition does not imply that $\eta^*_{\mathcal{G}}(\cdot)$ is information sufficient for all the models in $\Lambda$. Therefore, we could have that $\mathcal{I}(\mu^\theta_{X,Y}) > \mathcal{I}(\mu^\theta_{\eta^*_{\mathcal{G}}(X),Y})$ where $(X,Y)\sim \mu^\theta_{X,Y}$ for some model $\mu^\theta_{X,Y} \in \Lambda$. This last non-zero information loss condition makes this result non-trivial and interesting because, under this scenario,  (\ref{eq_wis_learning_9}) is strictly weaker than pure IS in Definition \ref{def_is}.
\end{remark}

\begin{remark} Bloem-Reddy and Teh in \cite{Bloem_2019} studied a stronger notion of invariance for $\Lambda$ under the action of a compact group $\mathcal{G}$. They consider the case where $\Lambda$ is $\mathcal{G}$-invariant if for any $g\in \mathcal{G}$ and any model $\mu^\theta_{X,Y}\in \Lambda$,  it follows that $(X,Y)=(g(X),Y)$ in distribution when $(X,Y)\sim \mu^\theta_{X,Y}$. This means that the complete joint model of $(X,Y)$ is invariant to the actions of $\mathcal{G}$.  Importantly, this invariant model-based assumption over $\Lambda$  is stronger than the operational $\mathcal{G}$-invariant assumption we used in Definition \ref{def_inv_classes} to derive Theorem \ref{th_wis_for_invariant_models}.\footnote{In fact, it is simple to verify that  if $\mu_{X,Y}$ is $\mathcal{G}$-invariant in the sense that $(X,Y)=(g(X),Y)$ in distribution for any $g\in \mathcal{G}$, then $\mu_{X,Y}$ is operational $\mathcal{G}$-invariant (Def. \ref{def_inv_models}). For completeness, this is shown in Appendix \ref{app_pro_strong_invariace_weak_invariance}.}
Furthermore,  under the model-based invariance used in \cite{Bloem_2019}, the authors showed that any maximal invariant transformation $\eta^*(\cdot)$ of $\mathcal{G}$ offers a $D$-separation of the model $\mu^\theta_{X,Y} \in \Lambda$, in the sense that $I(X;Y|\eta^*(X))=0$ when $(X,Y) \sim \mu^\theta_{X,Y}$.  Then $\eta^*(X)$ is an information sufficient representation of $Y$, which is strictly stronger than the concepts of OS (in Def. \ref{def_oper_suff_family}) used to derive Theorem \ref{th_wis_for_invariant_models}. Indeed, under this stronger model-based invariant assumption for $\Lambda$, it can be shown that condition (\ref{eq_wis_learning_9}) reduces to the IS condition in Definition \ref{def_is}, and the application of Theorem \ref{th_wis_for_invariant_models} reduces to restate the known 
condition that if $\left\{ U_i \right\}_{i\geq 1}$ is IS for every model in $\Lambda$, then $\left\{ U_i \right\}_{i\geq 1}$ is OP for every model in $\Lambda$.
\end{remark}


\section{Numerical Examples}
\label{sec_numerical}
To conclude this work, we design some simple settings (the model $\mu_{X,Y}$ and family of representations $ \left\{U_i,i\geq 1 \right\}$) 
to evaluate numerically the interplay between information loss and operation loss. 
For a given model $\mu_{X,Y}$, we will consider a collection of different representations (data-driven and non data-driven). We focus on discrete (digital) representations obtained from a vector quantization of $X$. On this, we  consider  universal information sufficient partitions (from the results in Section \ref{subsec_information_sufficiient}), and information sufficient data-driven partition (from the results in Section \ref{subsec_data_driven_representations}). 
The idea is to have a diverse range of representations of $X$ in terms of  information loss (expressiveness in the information sense) and see  how this diversity translates in the operation loss (expressiveness in the MPE sense). From this, we could observe if the order of representations in the information loss is an adequate predictor of the order obtained with the operation loss for regimes where the information loss is non-zero. In the regime of vanishing information loss, we could validate the result obtained in Theorem \ref{th_1} that says that a vanishing IL implies a vanishing OL and that vanishing WIL implies a vanishing OL. 

\subsection{Settings and Experimental Design}
\label{sec_numerical:setting}
We consider three classes of models ($\mu_{X,Y}$) for this analysis.  Each of them simple enough to compute the true values of information loss and 
operation loss with good precision, and each of them expressing some interesting structure that could be used as a prior information to obtain more effective representations for $X$. In the three model classes, some prior knowledge of the class will inform how we could partition $X$ more effectively for  MPE decision.

\subsubsection{The Models}
\label{sec_numerical:setting_model}
For this analysis, we consider a simple setting with $\mathcal{X}=\mathbb{R}^2$ and $\mathcal{Y}= \left\{1,\ldots,4 \right\}$ using a uniform marginal (i.e., $\mu_Y( \left\{ y \right\})=1/4$).  We consider three classes of models: 
\begin{itemize}
	\item {\em Scale Invariant Models}: the joint distribution $\mu_{X,Y}$ of $(X,Y)$ has the following structure: $\mu_Y$ is uniform in $\mathcal{Y}$ (i.e., $\mu_Y( \left\{y \right\})=1/4$ for any $y\in \mathcal{Y}$).  The probability of $X$ given $Y=y$ (i.e., $\mu_{X|Y}(\cdot | y)$) follows a normal distribution ($\sim \mathcal{N}({\bf m}_{X|Y}(y), {\bf K}_{X|Y})$) with the same isotropic covariance ${\bf K}_{X|Y}=\sigma \cdot {\bf I}_{2}$, where ${\bf I}_{2}$ denotes the identity, 
	and  mean  ${\bf m}_{X|Y}(y) \equiv \mathbb{E}(X|Y=y) \in \mathbb{R}^2$ given by: ${\bf m}_{X|Y}(1)^{t}=(\alpha, \alpha)$, ${\bf m}_{X|Y}(2)^{t}=(-\alpha, \alpha)$, ${\bf m}_{X|Y}(3)^{t}=(\alpha, - \alpha)$ and ${\bf m}_{X|Y}(4)^{t}=(-\alpha, -\alpha)$. To illustrate the symmetric pattern of the model,  the zones of equal-density for the 4 conditional distributions $ \left\{\mu_{X|Y}(\cdot | y), y\in \mathcal{Y} \right\}$ are presented in Fig.\ref{fig0}.
	
	Finally, we have two scalar parameters (degrees of freedom) $\sigma>0$ and $\alpha>0$ that determine the joint distribution 
	of $(X,Y)$, which we denote by $\mu^S_{X,Y}(\sigma,\alpha)$. Importantly, from the symmetry of the class $ \left\{ \mu^S_{X,Y}(\sigma,\alpha), \sigma>0, \alpha>0 \right\} \subset \mathcal{P}(\mathcal{X} \times \mathcal{Y})$, 
	the MPE rule is independent of $\alpha$ and $\sigma$, and induces the following  operationally sufficient partition
	\begin{equation}\label{eq_sec_numerical_1}
	\pi_S^{*}= \left\{ \underbrace{[0,\infty) \times [0,\infty)}_{A^*_{S,1}}, \underbrace{(-\infty,0) \times [0,\infty)}_{A^*_{S,2}}, \underbrace{(-\infty,0] \times (-\infty,0)}_{A^*_{S,3}},  \underbrace{(0,\infty] \times (-\infty,0)}_{A^*_{S,4}} \right\}.
	\end{equation}
	where the MPE rule is  $\tilde{r}_S(x)=\sum_{j=1}^{4}1_{A^*_{S,j}}(x) \cdot j$.
	\item {\em Translation Invariant Models}: This 2D joint distribution $\mu_{X,Y}$ follows the same Gaussian parametric structure for $\mu_{X|Y}(\cdot|y)$ 
	of the previous model. In this case, the mean vectors of the 5 equiprobable clases are oriented in one (1D) direction, see Fig. \ref{fig6}. This 1D linear disposition of the mean vectors makes the MPE rule invariant (and $\mu_{X,Y}$, see Def. \ref{def_inv_models}) to any translations in the direction that is orthogonal to the direction used to place the mean vectors of $\left\{ \mu_{X|Y}(\cdot|y), y\in \mathcal{Y} \right\} $.
	
	\item {\em Rotation Invariant Models}: We use the same 2D Gaussian parametric structure of the previous two cases.  In this scenario,  $\left\{ \mu_{X|Y}(\cdot|y), y\in \mathcal{Y} \right\} $ shares the same mean vector (the zero vector for simplicity), where $\sigma_y$ in ${\bf K}_{X|Y=y}=\sigma_y \cdot {\bf I}_{2}$  is a function of the class $y\in \mathcal{Y}$. This centered structure model makes its MPE rule 
	 invariant to any rotation of the space (see Defs. \ref{def_inv_rules} and \ref{def_inv_models}). Samples of this model are illustrated in Fig. \ref{fig8}.
\end{itemize}

\subsubsection{The Partitions}
\label{sec_numerical:setting_partition}
For the three models presented in Section \ref{sec_numerical:setting_model}, we will use the following partitions of $\mathcal{X}$: 
\begin{itemize}
	\item {\em Product partition (PP)}: we consider $K>0$ sufficiently large, and we produce a uniform quantization of the bounded space $[-K,K] \times [-K,K]$
	following the product structure presented in Section \ref{subsec_universal_quantization}. We denote these partitions by $ \left\{ \pi^P_n, n=1,..,N \right\}\subset \mathcal{Q}(\mathcal{X})$ with sizes $k_n= \left| \pi^P_n \right|$ and its representations in (\ref{eq_subsec_inf_divergence_19}) (VQs) by $ \left\{ \eta^P_n(\cdot),  n=1,..,N \right\}$. An illustration of this partition is presented in Fig.\ref{fig3_a}. 
	
	\item {\em Gessaman partition (GP)}:	 Using empirical samples of the model $\mu_{X,Y}$, i.e., $(X_1,Y_1).....(X_m,Y_m)$, we implement 
	the statistically equivalent partition presented in Section \ref{stat_equiv_blocks}. Therefore, we have a family of data-driven  partitions 
	that we denote by $ \left\{ \pi^G_n, n=1,..,N \right\} \subset \mathcal{Q}(\mathcal{X})$  with $k_n= \left| \pi^G_n \right|$
	where the different sizes (cardinalities) were obtained  by changing the number of samples $m$ used to construct $\pi^G_n$  (see Eq.(\ref{eq_stat_equiv_blocks_1})). Using Eq.(\ref{eq_stat_equiv_blocks_1}), we design $m_n$ (the number of samples) such that $k_n= \left| \pi^G_n \right|$. 
	An illustration of this non-product data-driven partition is presented in Fig. \ref{fig3_b}.
	
	\item {\em Tree-structured partition (TSP)}: Using empirical samples of the model $\mu_{X,Y}$, i.e., $(X_1,Y_1).....(X_m,Y_m)$, we implement 
	the TSP in Section \ref{subsec_tree_partition}. This scheme produces a family of data-driven  partitions 
	that we denote by $ \left\{ \pi^T_n, n=1,..,N \right\} \subset \mathcal{Q}(\mathcal{X})$  with $k_n= \left| \pi^T_n \right|$
	where the different sizes (cardinalities) were obtained  by changing the number of samples $m$. We design $m_n$ (the number of samples) 
	such that $k_n= \left| \pi^T_n \right|$.  An illustration of this data-driven partition is presented in Fig. \ref{fig3_c}. 
\end{itemize}

For each model $\mu_{X,Y}$ to be analized, we produce at  least three partition schemes: $ \left\{ \pi^P_n, n=1,..,N \right\}$, $ \left\{ \pi^G_n, n=1,..,N \right\}$, and $ \left\{ \pi^T_n, n=1,..,N \right\}$. The last two are data-driven schemes obtained from a i.i.d. samples of the true model. We designed them in increasing order of complexity (representation quality), i.e., $\left|  \pi^P_n\right| <  \left|  \pi^P_{n+1}\right| $, and consistent in size, i.e.,  $ \left|  \pi^P_n\right| =  \left|  \pi^G_n\right| =  \left|  \pi^T_n\right|$ for any $n\in  \left\{1,..,N \right\}$. 
In addition, and depending on the model case, we include an additional partition scheme using prior knowledge of the class of models in Section \ref{sec_numerical:setting_model}.

\subsubsection{Estimation of Information and Operation losses}
\label{sub_sec_computation_il_ol}
For a given model $\mu_{X,Y}$  and  any partition $\pi_i$ in our setting,  
we consider its representation $\eta_{\pi_i}(\cdot)$, $U_i=\eta_{\pi_i}(X)$ and the induced distribution 
of $(U_i,Y)$, i.e., $\mu_{U_i,Y}$. For the representation, we have two scenarios: non data-driven $\pi_i$ and data-driven $\pi_i(x_1,..,x_{m_i})$,  where for this last 
case is a function of the unsupervised samples $x_1,..,x_{m_i} \in \mathcal{X}^{m_i}$ that follow the true marginal $\mu_X$.   
For what follows, the dependency on $x_1,..,x_{m_i} $ of the data-dependent partition will be considered implicit and the cells of the partition will be denoted by  $\pi_i= \left\{A_j,j=1,..,k_i \right\}$ in both cases.

To derive expression for the information loss and operation loss associated with $U_i$,  we use i.i.d. samples of $(X,Y)\sim \mu_{X,Y}$ to 
estimate $\mathcal{I}(\mu_{X,Y})$, $\mathcal{I}(\mu_{U_i,Y})$, $\ell({\mu_{X,Y}})$ and $\ell({\mu_{U_i,Y}})$. In particular using $(X_1,Y_1),...,(X_n,Y_n)$, we use a strongly consistent estimator of each of the mentioned terms. 

For the information loss of $U_i$ given a model $\mu_{X,Y}$, we know that $\mathcal{I}(\mu_{X,Y})=I(X;Y)=\mathbb{E}_{(X,Y)}\left\{\log \frac{\mu_{Y|X}(Y|X)}{\mu_{Y}(Y)} \right\}$ and, consequently, its empirical estimator (assuming knowledge of the model) is
\begin{equation}\label{eq_comp_il_ol_1}
	\hat{\mathcal{I}}_n(\mu_{X,Y}) \equiv  \frac{1}{n} \sum_{j=1}^n \log \frac{\mu_{Y|X}(Y_j|X_j)}{{\mu_{Y}(Y_j)}}.
\end{equation}
Concerning the discrete model $\mu_{U_i,Y}\in \mathcal{P}([k_i]\times \mathcal{Y})$, we have that 
\begin{equation}\label{eq_comp_il_ol_2}
	\mathcal{I}(\mu_{U_i,Y})=I(U_i;Y) = \sum_{j=1}^{k_i}\sum_{y\in \mathcal{Y}} \mu_{X,Y}(A_j\times  \left\{y \right\}) \log \frac{\mu_{X,Y}(A_j\times  \left\{y \right\}) }{\mu_{X}(A_j) \cdot \mu_{Y}( \left\{y \right\})}.
\end{equation}
Then, we can use the empirical version (known as the plug-in MI estimator) of (\ref{eq_comp_il_ol_2}) by 
\begin{equation}\label{eq_comp_il_ol_3}
	\mathcal{I}(\hat{\mu}^n_{U_i,Y}) \equiv \sum_{j=1}^{k_i}\sum_{y\in \mathcal{Y}} \hat{\mu}^n_{X,Y}(A_j\times  \left\{y \right\}) \log \frac{ \hat{\mu}^n_{X,Y}(A_j\times  \left\{y \right\}) }{ \hat{\mu}^n_{X}(A_j) \cdot  \hat{\mu}^n_{Y}( \left\{y \right\})}, 
\end{equation}
where $\hat{\mu}^n_{X;Y}$ denotes the empirical joint distribution given by $\hat{\mu}^n_{X;Y}(A \times  \left\{y \right\})= \frac{1}{n} \sum_{j=1}^n {\bf 1}_{A\times  \left\{y \right\}} (X_i,Y_i)$.  Finally, our empirical estimation of the information loss of $U_i$ is
\begin{equation}\label{eq_comp_il_ol_4}
	\hat{\mathcal{I}}_n(\mu_{X,Y}) - \mathcal{I}(\hat{\mu}^n_{U_i,Y}).
\end{equation}
From the law of large numbers, it follows that $\lim_{n \rightarrow \infty} \hat{I}_n(\mu_{X,Y})  = \mathcal{I}(\mu_{X,Y})$ and  
$ \lim_{n \rightarrow \infty} \mathcal{I}(\hat{\mu}^n_{U_i,Y})=\mathcal{I}(\mu_{U_i,Y})$  with probability one. Therefore, 
our empirical estimation of ${\mathcal{I}}(\mu_{X,Y}) - \mathcal{I}({\mu}_{U_i,Y})$ in (\ref{eq_comp_il_ol_4}) is strongly consistent. 

For the operation loss of $U_i$ given a model $\mu_{X,Y}$, we could determine the MPE decision $\tilde{r}_{\mu_{X,Y}}(\cdot)$ and $\tilde{r}_{\mu_{U_i,Y}}(\cdot)$ analytically as we know $\mu_{X,Y}$ and $\mu_{U_i,Y}$. Again using the i.i.d. realizations $(X_1,Y_1),..,,(X_n,Y_n)$ of $\mu_{X,Y}$,  we use the empirical risks: 
\begin{equation}\label{eq_comp_il_ol_5}
	\hat{\ell}_n(\mu_{X,Y}) \equiv \frac{1}{n} \sum_{j=1}^n {\bf 1}_{\left\{Y_i \right\}}(\tilde{r}_{\mu_{X,Y}}(X_i))
\end{equation}
\begin{equation}\label{eq_comp_il_ol_6}
	\hat{\ell}_n(\mu_{U_i,Y}) \equiv \frac{1}{n} \sum_{j=1}^n {\bf 1}_{\left\{Y_i \right\}}(\tilde{r}_{\mu_{U_i,Y}}(\eta_{\pi_i}(X_i)),
\end{equation}
and the empirical operation loss of $U_i$ is
\begin{equation}\label{eq_comp_il_ol_7}
	\hat{\ell}_n(\mu_{U_i,Y}) -  \hat{\ell}_n(\mu_{X,Y}).
\end{equation}
Importantly, by the law of large numbers, $\lim_{n \rightarrow \infty} \hat{\ell}_n(\mu_{U_i,Y}) -  \hat{\ell}_n(\mu_{X,Y}) = {\ell}(\mu_{U_i,Y}) -  {\ell}(\mu_{X,Y})$
with probability one.  

For the following sections, we will use these consistent estimators to have  precise indicators of the true losses for which we consider 
a sufficiently large $n$ in the computation of (\ref{eq_comp_il_ol_4})  and (\ref{eq_comp_il_ol_7}).

\subsection{Information loss vs. Operation loss}
\label{sec_numerical:tradeoff}
\begin{figure}
\centering
 \subfloat[Product partition $\pi_n^P$]{\label{fig0_a}\includegraphics[width=0.50\textwidth]{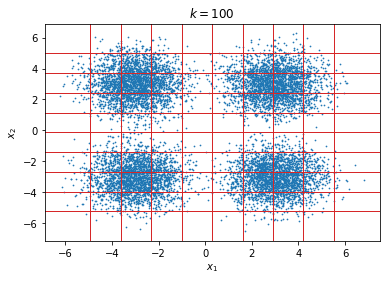}} \subfloat[Gessaman Partition $\pi_n^G$]{\label{fig0_b}\includegraphics[width=0.50\textwidth]{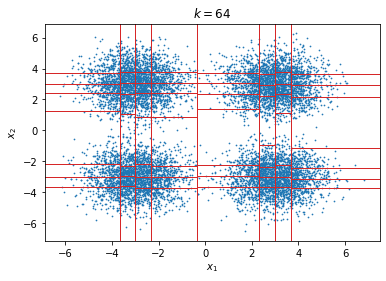}}\\
\subfloat[TSP partition $\pi_n^T$]{\label{fig0_c}\includegraphics[width=0.50\textwidth]{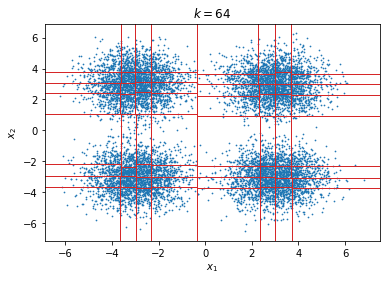}} \subfloat[Asymmetric  Partiton $\pi_n^B$]{\label{fig0_d}\includegraphics[width=0.50\textwidth]{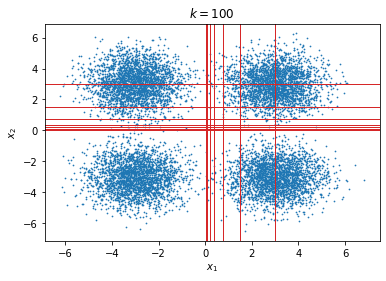}}
\caption{Illustration of the data-driven and non data-driven partitions (with $k=100$) used to obtain lossy representations (vector quantizers) of $X$.  For the data-driven methods, we show samples of the distribution used for this analysis: the Scale Invariant Model in Section \ref{sec_numerical:setting_model}.}
\label{fig0}
\end{figure}

We begin with the scale invariant model for which we consider ${\alpha=1.5}$ and $\sigma=1$. We use $10.000$ i.i.d. realizations of $(X,Y)$
to obtain accurate estimations of the information losses and operation losses. We consider the product  (uniform) partitions $\left\{ \pi^P_n\right\}_n$, the Gessaman partitions $\left\{ \pi^G_n\right\}_n$, and TSP $\left\{ \pi^T_n\right\}_n$. In addition, we include in the analysis an asymmetric partition scheme, denoted by  $\left\{ \pi^B_n\right\}_n$,  which is non-data driven and approximates (in the limit of large size) the optimal quantization in (\ref{eq_sec_numerical_1}). 
This last scheme refines (in a dyadic way) only the zone of low values of $\mathbb{R}^2$ in an asymmetric way as can be seen  in Fig.\ref{fig0_d}. 
The idea here was to produce a family of lossy representations with the capacity to approximate (in the limit) the optimal partition $\pi^*_S$ in (\ref{eq_sec_numerical_1}) for this class of models. For each of the four representation strategies, we produce a collection of partitions of different sizes (number of cells).   Figure \ref{fig0} illustrates the model (how the i.i.d. samples for each of the four classes are distributed in the space) and the different partitions strategies adopted (data-driven and non data-driven) considering specific number of cells for illustration. These figures show the diversity of ways in which $X$ is quantized in this analysis.

\begin{figure}
\centering
\includegraphics[width=1.00\textwidth]{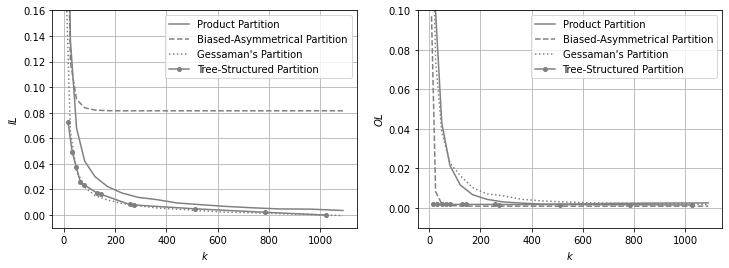}
\caption{Information loss (the left panels) and operation losses (the right panel) curves for different number of cells ($k$) of the partitions: Product, Gessaman, Tree-structured and Biased-Asymmetrical. These curves are produced using  samples from the model illustrated in Fig.\ref{fig0}.}
\label{fig1}
\end{figure}

Figure \ref{fig1} presents the information losses (the left panels) and the operation losses (the right panel) for the 4 collections of partitions across a range of partition sizes from $10$ to $1,000$. The trends of all curves follow the expected decreasing pattern: as the partition size increases so does the quality to represent the information that $U_i$ has about $Y$.  However, on the information loss side, there are two clear groups of curves. The first group shows a vanishing information loss (Product, Gessaman, and TSP) expressing the fact that these families are information sufficient: these representations capture all the information that $X$ has about $Y$ ($\mu_{X,Y}$) in the MI sense as $k$ increases. 
This confirms the results presented in Section \ref{sec_learning_empirical}. In addition, for $k$ in the range $[50-400]$, there are representation differences among these three cases. The data-driven schemes (TSP and Gessaman) offer better information losses  than the product (non-adaptive) partition. This discrepancy is attributed to the fact that data-driven approaches are more flexible in their capacity to adapt its structure to the problem as it can be seen in Fig.\ref{fig0}. 

On the expressiveness of these three schemes (Product, Gessaman, and TSP) in terms of operation loss (OL) (the right panel), the exact order observed in the information loss side is not preserved. Indeed, for sample size in the range $[50,400]$, the product scheme (non-adaptive) shows marginally better OL than the Gessaman scheme (data-driven) in opposition of what is observed in the IL indicator. However, the differences are very marginal as the magnitude of the OL in this regime is very close to zero.  On the other hand, the TSP shows a clear advantage in  OL  compared with the other two methods: the Gessaman and the product partition. For this specific comparison (Product, Gessaman, and TSP), the IL  indicador predicts the best solution in the operational side.  On the expressiveness of the TSP, this solution captures with very few cells ($k <10$) the optimal partition in (\ref{eq_sec_numerical_1}), which can be attributed to the symmetry observed in the unsupervised samples (see Fig.\ref{fig0}) mimicking the symmetry of the optimal rule in (\ref{eq_sec_numerical_1}).

\begin{figure}
\centering
\includegraphics[width=0.90\textwidth]{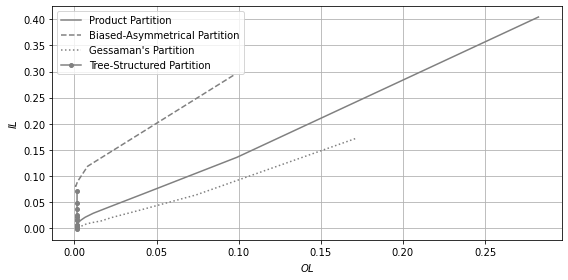}
\caption{Information loss vs. operation loss for the partitions schemes: Product, Gessaman, Three-structured and Biased-Asymmetrical.}
\label{fig2}
\end{figure}

If we look at the biased-asymmetrical partitions $ \left\{\pi_n^B \right\}$, the analysis is insightful. This family was designed from the idea of only approximating $\pi^*_S$ in the spirit of the weak information loss indicator in (\ref{eq_sec_II_8d}). Importantly, we observe that this family is not information sufficient (left panel in Fig.\ref{fig1}). However, on the operation loss  (right panel of Fig.\ref{fig1}) this family shows to be operationally sufficient. Furthermore, this  scheme has an almost zero OL in all the size regimes (the only exception is when $k<20$). Indeed, these representations show to be very more competitive in operation loss when the partition size increases and it is better  than the Gessaman and the Product partitions in all regimes, which is very interesting considering that these last two schemes are information sufficient. This concrete example illustrates how prior knowledge of the task (in this case the symmetry of the optimal solution) can be used to meet OS very efficiently without the need to achieve the strong IS requirement.  In addition, this asymmetric  construction shows an example (based on prior knowledge of the task) where pure IL is not adequate as a predictor of the quality of representations if the objective is selecting representations based on the operational performance (probability of error) in classification. 

To summarize these results, Fig. \ref{fig2} presents the information loss (IL) vs operation loss (OL) for our four schemes.

\subsubsection{Rotated version of the Scale Invariant Model}
\label{sec_numerical:tradeoff_symetric}
\begin{figure}
\centering
 \subfloat[Product partition $\pi_n^P$]{\label{fig3_a}\includegraphics[width=0.50\textwidth]{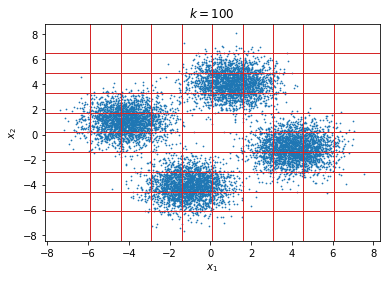}} \subfloat[Gessaman Partition $\pi_n^G$]{\label{fig3_b}\includegraphics[width=0.50\textwidth]{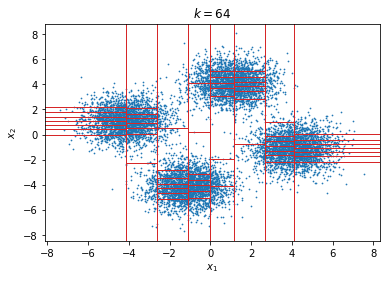}}\\
\subfloat[TSP partition $\pi_n^T$]{\label{fig3_c}\includegraphics[width=0.50\textwidth]{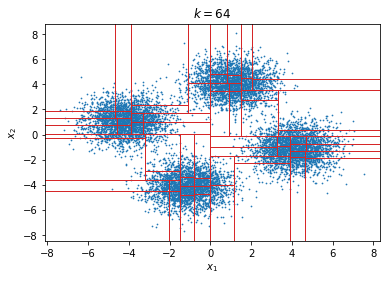}} \subfloat[Asymmetric  Partiton $\pi_n^B$]{\label{fig3_d}\includegraphics[width=0.50\textwidth]{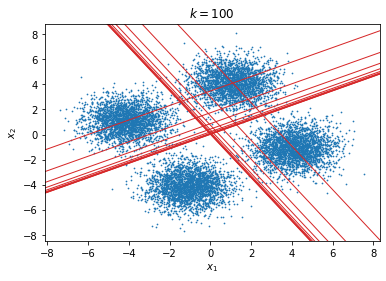}}
\caption{Illustration of the data-driven and non data-driven partitions (with $k=100$) used to obtain lossy representations (vector quantizers) of $X$.  For the data-driven methods, we show samples of the distribution used for this analysis: a rotated version of the Scale Invariant Model in Section \ref{sec_numerical:setting_model}.}
\label{fig3}
\end{figure}

To make the previous problem setting a bit more challenging for the partitions schemes that are coordinate oriented,  we consider a rotated version of the scale invariant model. The samples for this rotated model and the obtained partitions used for this analysis are illustrated in Fig. \ref{fig3}.  We consider the same family of data-driven partitions (TSP and Gessaman)  and non-data driven partitions (Product and Biased-Asymmetric).  All partitions (with the exception of the asymmetric that used prior knowledge of the optimal decision region) are produced in a coordinate based manner. Then, there is a mismatch with the orientation of the data that makes this scenario a bit more challenging from the information loss and operation loss perspective. 

\begin{figure}
\centering
\includegraphics[width=1.00\textwidth]{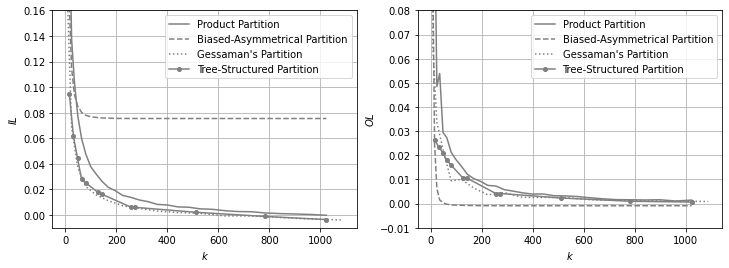}
\caption{Information loss (the left panels) and operation losses (the right panel) curves for different number of cells (k) of the partitions: Product, Gessaman, Tree-structured and Biased-Asymmetrical. These curves are produced using  samples from the model illustrated in Fig.\ref{fig3}.}
\label{fig4}
\end{figure}

Fig. \ref{fig4} shows the information loss and the operation loss curves. 
These curves show again two groups of decreasing curves on the information loss side: three schemes are information sufficient (Gemmanan, TSP and Product),   while the asymmetric scheme $ \left\{\pi_n^B \right\}$ does not show a vanishing information loss.  Interestingly, not only $ \left\{\pi_n^B \right\}$ has a 
vanishing operation loss but is the scheme with the best operational performance (see the right panel in Fig.\ref{fig4}).  Consequently, as in the previous example, the information loss is not adequate predictor of the ranking in operation loss among these four schemes. Furthermore, the IL indicator is blind in expressing the vanishing operation loss of $ \left\{\pi_n^B \right\}$.  As a final comment, we observe that achieving a close to zero operation loss in this case is quite more difficult than what is  observed on the previous non-rotated example as we anticipated.  Indeed, the data-driven methods achieves close to zero operation loss only after $k>800$ in clear contrast with what is presented in the right panel of Fig. \ref{fig1}.

\subsubsection{Translation Invariant Model}
\label{sec_numerical:tradeoff_trans}
\begin{figure}
\centering
 \subfloat[Product partition $\pi_n^P$]{\label{fig6_a}\includegraphics[width=0.50\textwidth]{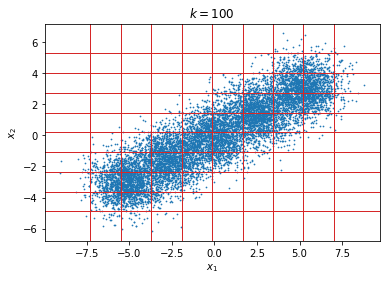}} \subfloat[Gessaman Partition $\pi_n^G$]{\label{fig6_b}\includegraphics[width=0.50\textwidth]{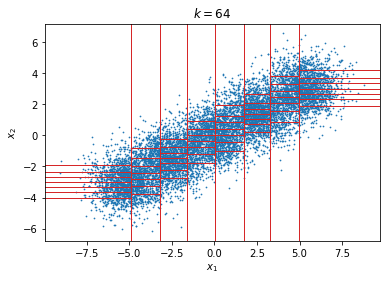}}\\
\subfloat[TSP partition $\pi_n^T$]{\label{fig6_c}\includegraphics[width=0.50\textwidth]{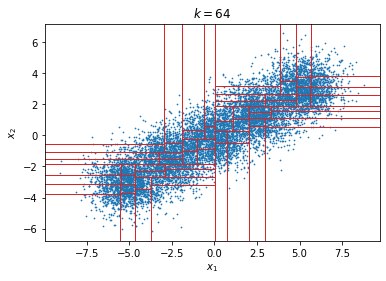}} \subfloat[1D-Uniform Partiton $\pi_n^S$]{\label{fig6_d}\includegraphics[width=0.50\textwidth]{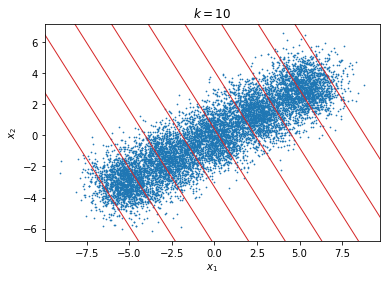}}
\caption{Illustration of the data-driven and non data-driven partitions (with $k=100$) used to obtain lossy representations (vector quantizers) of $X$.  For the data-driven methods, we show samples of the distribution used for this analysis: a rotated version of the Translation Invariant Model in Section \ref{sec_numerical:setting_model}.}
\label{fig6}
\end{figure}

In this case, we consider a rotated version of the translation invariant model introduced in Section \ref{sec_numerical:setting_partition} to make the problem 
non-coordinate oriented and more interesting for our analysis.  
The data distribution of this model is presented in Fig.\ref{fig6}. 
In addition to the partitions presented in Section \ref{sec_numerical:setting_partition},  we include a partition that project  the data in the direction of  symmetry of the problem (1D projection) to then perform a uniform partition in this projected scalar domain. This informed (with prior knowledge of the task) representation of $X$ is illustrated  in Fig.\ref{fig6_d} and denoted by $\pi^S_n$. 

As this problem is translation invariant,  there is a lossy mapping $\eta: \mathbb{R}^2 \rightarrow \mathbb{R}$ that is operationally sufficient for $\mu_{X,Y}$, in the sense that $\ell(\mu_{X,Y})=\ell(\mu_{\eta(X),Y})$ (see the statement of Theorem \ref{th_wis_for_invariant_models}).  Because of this operational structure, we consider the projected information loss $I((\eta{(U)},U_i); Y) - I(U_i; Y) = I(\eta(U);Y|U_i)$ for our analysis (see Eq. (\ref{eq_wis_learning_7})) to evaluate in this  task if this refined fidelity indicator better predict the operation loss in this example. 
\begin{figure}
\centering
\includegraphics[width=1.00\textwidth]{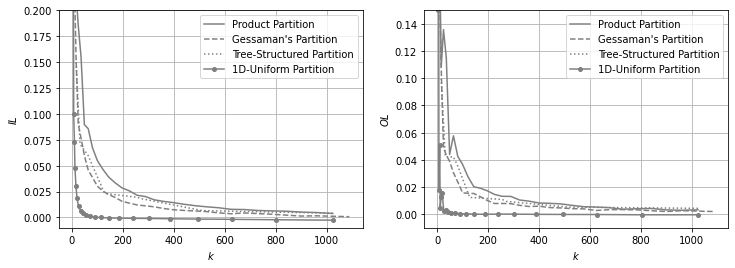}
\caption{Information loss (the left panels) and operation losses (the right panel) curves for different number of cells ($k$) of the partitions: Product, Gessaman, Tree-structured and 1D-Uniform projection. These curves are produced using  samples from the model illustrated in Fig.\ref{fig6}.}
\label{fig7}
\end{figure}

Fig.\ref{fig7} shows the task-dependent information loss $I(\eta(U);Y|U_i)$ and the operation loss for our family of four representations schemes.
In the information loss,  all partitions perform very well in capturing the information that $\eta(X)$ has about $Y$ and they offer 
the expected monotonic and vanishing decreasing pattern.  In particular, data-driven partitions are slightly better than the product one (that is consistent with their more efficient adaptation illustrated in Fig. \ref{fig6}). Importantly, the projected partition (that use the operationally sufficient representation $\eta(\cdot)$ in its
construction) takes advantage of this prior knowledge, which is expressed in a clearly superior performance in the projected information loss.
In fact, the decreasing trend of its loss curve is very drastic showing that this representation is very effective in capturing the latent information 
structure of $I(\eta(X);Y)$. For example, the informed scheme (1D-Uniform) achieves a loss with $10$ cells that is obtained with the TSP and Gessaman (partitioning the whole space) with more than $500$ cells. Therefore, the effect of projecting the problem to a smaller dimension is very radical 
in terms of the projected information loss $I(\eta(U);Y|U_i)$. Importantly, these patterns are preserved in the operation loss domain (right panel of Fig.\ref{fig7}). 
Here, the drastic difference in performance of the 1D-Uniform representation with respect to the rest is also observed.
Also the order, or the ranking, from the less informative to the more informative representation is consistently observed in the operation side (right panel), 
showing for this example that $I(\eta(U);Y|U_i)$ provides a good prediction of the relative (operational) performance of the schemes.
\subsubsection{Rotation Invariant Model}
\label{sec_numerical:tradeoff_rot}
\begin{figure}
\centering
 \subfloat[Product partition $\pi_n^P$]{\label{fig8_a}\includegraphics[width=0.50\textwidth]{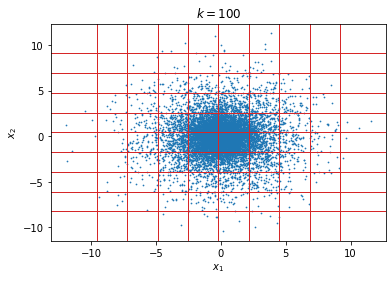}} \subfloat[Gessaman Partition $\pi_n^G$]{\label{fig8_b}\includegraphics[width=0.50\textwidth]{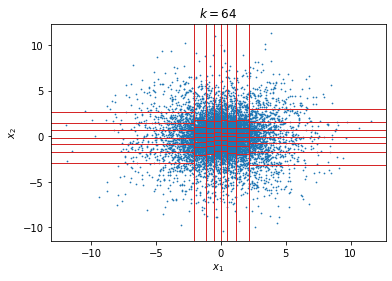}}\\
\subfloat[TSP partition $\pi_n^T$]{\label{fig8_c}\includegraphics[width=0.50\textwidth]{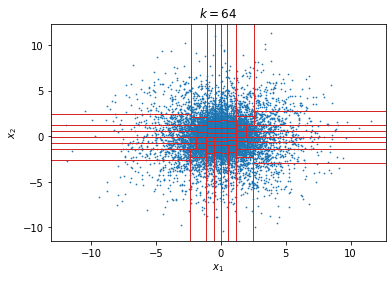}} \subfloat[1D-Uniform Partiton $\pi_n^R$]{\label{fig8_d}\includegraphics[width=0.50\textwidth]{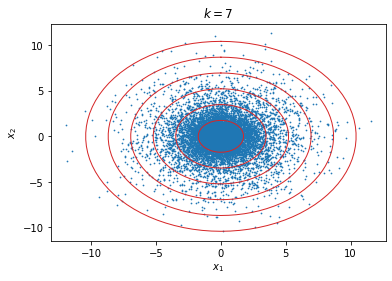}}
\caption{Illustration of the data-driven and non data-driven partitions (with $k=100$) used to obtain lossy representations (vector quantizers) of $X$.   For the data-driven methods, we show samples of the distribution used for this analysis: the Rotation Invariant Model in Section \ref{sec_numerical:setting_model}.}
\label{fig8}
\end{figure}

Finally, we consider the rotation invariant model introduced in Section \ref{sec_numerical:setting_partition} that is illustrated in Fig. \ref{fig8}. This problem is quite 
more challenging for the representations in Section \ref{sec_numerical:setting_partition} that are coordinate oriented. 
As in the previous example,  this model has a 1D projection that is operationally sufficient $\eta: \mathbb{R}^2  \rightarrow \mathbb{R}$, meaning that $\ell(\mu_{X,Y})=\ell(\mu_{\eta(X),Y})$. Here, we also consider the projected information loss for our analysis, i.e.  $I(\eta(U);Y|U_i)$, and a specific representation (1D-Uniform) induced by projecting $X$ using $\eta(\cdot)$ and then performing a uniform quantization in this scalar domain (see Fig. \ref{fig8_d}). These collection of partitions including this last informed partition scheme (1D-Uniform) are illustrated in Fig.\ref{fig8}.  

\begin{figure}
\centering
\includegraphics[width=1.00\textwidth]{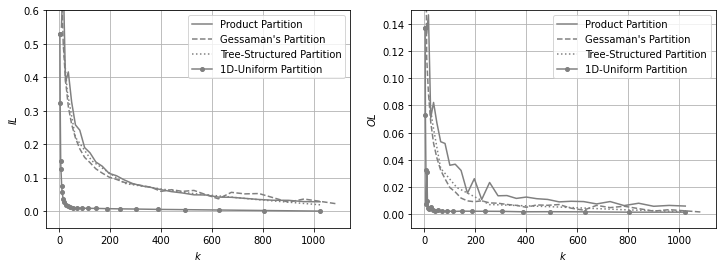}
\caption{Information loss (the left panels) and operation losses (the right panel) curves for different number of cells ($k$) of the partitions: Product, Gessaman, Tree-structured and 1D-Uniform projection. These curves are produced using  samples from the model illustrated in Fig.\ref{fig8}.}
\label{fig9}
\end{figure}

Fig.\ref{fig9} shows the curves associated to the projected information loss and the operation loss. As in the previous case, the difference in information loss for the projected representation is radical, meaning that prior knowledge translates in a significant boost in expressiveness.  In this scenario, the discrepancy is even more drastic than the previous example in Section \ref{sec_numerical:tradeoff_trans}, where the 1D-Uniform representation has a projected information loss with 10 cells that is better than what all the other alternatives partitions can achieve with $1,000$ cells, which is a very impressive difference. As in the previous example in Section \ref{sec_numerical:tradeoff_symetric}, these differences translate in the operation loss showing a clear advantage of the 1D-Uniform with respect to the rest. In addition, a relatively consistent performance trend is observed from information loss to operation loss as observed in the previous example in Section \ref{sec_numerical:tradeoff_trans}.


\section{Summary and Final Discussions}
\label{sec_summary}
This work offers new results that shed light on the interplay between information loss (in the Shannon sense) and operation loss (in the classical MPE sense) when considering a general family of lossy representations of an observation vector $X$  in $\mathbb{R}^d$.  Our main asymptotic result (Theorem \ref{th_1}) supports the idea that creating a family of information sufficient representations is an adequate criterion in the sense that these representations have a vanishing residual error with respect to the MPE decision acting on $X$ to classify (a class) $Y$.  On the other hand, Theorem \ref{th_1} shows that pure informational sufficiency (in the sense of Def. \ref{def_is}) is a conservative criterion. Indeed, Theorems \ref{th_1}, \ref{th_wis_over_a_family}  and  \ref{th_wis_for_invariant_models} 
show that a  weaker notion of informational sufficiency (in the form of  Def. \ref{def_wis}) suffices  to obtain the required operational result.

In the context of a learning setting, we worked on an extension of Theorem \ref{th_1}  where $\mu_{X,Y}$ belongs to a class of models $\Lambda$.  
We studied how the structure of $\Lambda$ can be used as prior information to propose less conservative (and non-oracle) weak forms of informational  sufficiency.  In this direction, we look at the family of invariant models (invariant to the action of a compact group) \cite{Bloem_2019}, where it is possible to determine ``{\em a non-oracle}" surrogate of $r^*(\cdot)$ (in Theorem \ref{th_1}) that extends our main result (WIS implies OS) in this learning setting. This new result is stated in Theorem \ref{th_wis_over_a_family}  and then applied to the class of invariant models in Theorem \ref{th_wis_for_invariant_models}.

Finally, our experimental analysis supports some of our theoretical findings, illustrating that the WIS condition is strictly weaker than IS for a given model and collection of representations. In addition,  we validate that pure information loss is not always an adequate predictor of operation loss in classification. Our empirical findings also show that in the presence of prior knowledge about the task, this knowledge can be used to design representations that offers better operation losses (approximation error properties). In the presented scenarios with prior information, the projected version of the IL, adopted from  Theorem \ref{th_wis_over_a_family}, shows to be an adequate predictor of the operation loss.

\subsection{Applications of our Results}
\label{broad}
The analysis presented in this work about the interplay between vanishing information and operation loss offers relevant insight in learning settings.  First, our  results support the universality of approximating (or learning) compressed representations that capture the mutual information between $X$  and $Y$, for example, via minimization of the conditional entropy $H(Y|U)$, or  maximization of $I(U;Y)$. This is a widely adopted criterion in representation learning,  in the form of maximizing empirical versions of the mutual information (info-max problems) or  minimizing empirical versions of the conditional entropy over a family of compressed representations of $X$ \cite{amjad_2019,alemi_2017,achille_2018b,vera_2018,strouse_2017,tegmark_2020}.
On the other hand, our results show that IS is a conservative criterion if the objective is designing OS representations. On this dimension, we introduce weaker (strictly weaker in some cases) information sufficient conditions that imply OS for different classification settings.

Theorem \ref{th_1} (WIS implies OS in its oracle form based on $\tilde{U}$) does not offer a practical strategy that can be used to learn or select representations from data. However, the extensions in Theorems \ref{th_wis_over_a_family} and  \ref{th_wis_for_invariant_models} could be used, as shown in Section \ref{sub_sec_wis_applied_in_learning_cases}, and motivate practical avenues of research: for instance, designing new "non-oracle" information losses inspired by WIS, weaker than pure IS,  that could be adopted to rank, select or learn representations from data when some prior knowledge of the task is available.

On the application of Theorem \ref{th_1} for the analysis of representations in learning problems, we show in  Section \ref{sec_learning_empirical} an interesting connection with the problem of mutual information estimation.
Here, it is worth emphasizing the connection made for the analysis of data-driven partitions  \cite{silva_2010,silva_isit_2007,silva_2010b,vajda_2002,darbellay_1999,gonzales_2020}. These data-driven representations (representations learned from data) are information sufficient with probability one. 
Then Theorem \ref{th_1} is used to prove that these representations are operational sufficient (with probability one) for the task of MPE decision, as presented in Lemmas \ref{lm_stat_bllock_shrinking_cell} and \ref{lm_balanced_shrinking_cell} in Section \ref{sec_learning_empirical}. Therefore, we demonstrate that Theorem \ref{th_1} could be a useful tool for the analysis of practical representation strategies in classification.

\subsection{Connections with Learning Algorithms}
In the context of Theorems \ref{th_wis_over_a_family} and \ref{th_wis_for_invariant_models},  it is worth mentioning that a sufficient condition to achieve  $I(\eta_{\Lambda}(X);Y|U_i) \rightarrow 0$ is asking that $H(\eta_{\Lambda}(X)|U_i) \rightarrow 0$ when $\eta_{\Lambda}(\cdot)$ is finite or countable mapping.  Interestingly, this last  learning criterion is non-supervised and practical.  On this,  it is worth mentioning a natural connection with the the recent work 
 by Dubois {\em et.~al} \cite{Dubois_2021}.  They proposed a new unsupervised representation learning task (over a collection of invariant models $\Lambda$) that finds an optimal tradeoff between {\em compression} $I(X;U)$ and {\em fidelity} $H(\eta_{\Lambda}(X)|U)$.  They show  that finding a lossy encoder (compressor) $U_\delta$ of $X$ as the solution of 
 $$\min_{U} I(X;U)  \text{ s.t. } H(\eta_{\Lambda}(X)|U) \leq \delta,$$
 in theory (see in \cite[ Th.1]{Dubois_2021}) offers a collection of lossy representations $ \left\{ U_{\delta_n} \right\}_{n\geq 1}$, where $H(\eta_{\Lambda}(X)|U_{\delta_n})\leq \delta_n$ and if $\delta_n \rightarrow 0$ then $ \left\{ U_{\delta_n} \right\}_{n\geq 1}$ meets the conditions $I(\eta_{\Lambda}(X);Y|U_i) \rightarrow 0$ that implies that $ \left\{ U_{\delta_n} \right\}_{n\geq 1}$ is OS from Theorem \ref{th_wis_for_invariant_models}.  Therefore,  there is an unsupervised learning scheme solution of $\min_{U} I(X;U) - \beta(\delta) H(\eta_{\Lambda}(X)|U) \leq \delta$\footnote{There is a similarity with the {\em information bottleneck} (IB) problem: when $\eta(X)$ is replaced by $Y$, we recover the unconstrained version of the IB method.} that extracts all the information about $\eta_\Lambda(X)$  and produces representations that meet (in theory as $\delta$ vanishes) the weak information sufficient condition stated in Theorem \ref{th_wis_for_invariant_models}. 

\subsection{Data-Driven Representations and Mutual Information Estimation}
\label{sub_sec_data_driven_represent_final_remarks}
As a side observation of our analysis in Section \ref{sec_learning_empirical}, we show that it is feasible to use unsupervised data-driven representations 
to capture in discrete forms (digital representations) all the information that $X$ has to offer to discriminate $Y$ (with arbitrarily large probability). Interestingly, the statistically equivalent principle offers solutions that meet the very
strong information sufficient expressiveness condition in Def. \ref{def_is} almost surely (see Lemma \ref{lm_balanced_shrinking_cell} and Lemma \ref{lm_stat_bllock_shrinking_cell}), which translates into being operation sufficient almost surely from Corollary \ref{cor_th1_data_driven} and Theorem \ref{th_1}. More generally,  the results in Section \ref{sec_learning_empirical} offer a path that connects methods and results used for the problem of mutual information estimation into the design of unsupervised representations for classification. 

\newpage

\section{Proofs of the Main Results}
\label{sec_proof}

\subsection{Proof of Theorem \ref{th_2}}
\label{proof_TH1}
\begin{proof}
Let us first look at the definition of $g(\mu_{X,Y}, B)$ in  (\ref{eq_sec_II_8b}). This is a function of the model $\mu_{X,Y}$, the partition 
$\pi^*=\left\{ A^*_y, y \in \mathcal{Y} \right\}$ in (\ref{eq_sec_II_7}) and a set $B \subset \mathcal{X}$. In particular, we have that 
\begin{equation}\label{eq_TH1_1}
	g(\mu_{X,Y}, B) = \left[   1- \max_{y\in \mathcal{Y}} \mu_{Y|X}(y | B) \right]- \sum_{ A^*_u \in \pi^*} {\mu_X(A^*_u | B )} \cdot  \left[   1- \max_{y\in \mathcal{Y}} \mu_{Y|X}(y | A^*_u \cap B ) \right]. 
\end{equation}
The first term on the RHS of (\ref{eq_TH1_1}) can be seen as the prior minimum error probability of a random variable $\tilde{Y}$ in $\mathcal{Y}$ with marginal probability $(v_{\tilde{Y}}(y) \equiv \mu_{Y|X}(y|B))_{y\in \mathcal{Y}} \in \mathcal{P}(\mathcal{Y})$. On the other hand, the second term on the RHS of (\ref{eq_TH1_1})  can be seen as the MPE of a joint vector $(\tilde{X}, \tilde{Y})$ in $\mathcal{Y}\times \mathcal{Y}$ with probability $v_{\tilde{X}, \tilde{Y}}$ in $\mathcal{P}(\mathcal{Y} \times \mathcal{Y})$ defined by 
\begin{equation}\label{eq_TH1_3}
	v_{\tilde{X}, \tilde{Y}}(u,y) \equiv  \frac{\mu_{X,Y}(A^*_u\cap B \times \left\{ y \right\} )} {\mu_X(B)}, \ \forall (u,y)\in \mathcal{Y}^2. 
\end{equation}
The second term in (\ref{eq_TH1_1}) is precisely $\ell(v_{\tilde{X}, \tilde{Y}})$.  Adopting Lemma \ref{th_interplay_cond_entropy_error_prob} in this context, we can use its corollary in (\ref{eq_main_1b}) to obtain  that 
\begin{align}\label{eq_TH1_4}
	\mathcal{I}(v_{\tilde{X}, \tilde{Y}})=I(\tilde{X};\tilde{Y}) &\geq  H(\tilde{Y}) - \mathcal{H} (\mathcal{R}(v_{\tilde{Y}}, \ell(v_{\tilde{X}, \tilde{Y}}))) \nonumber\\ 
				     &= \mathcal{H}(v_{\tilde{Y}}) - \mathcal{H} (\mathcal{R}(v_{\tilde{Y}}, \ell(v_{\tilde{X}, \tilde{Y}})))\nonumber\\ 
				     &= \mathcal{H}(\mu_{Y|X}(\cdot |B)) - \mathcal{H} (\mathcal{R}(\mu_{Y|X}(\cdot |B), \ell(\mu_{\tilde{X}, \tilde{Y}})))
\end{align}
where $\ell(v_{\tilde{X}, \tilde{Y}})=\left[   1- \max_{y\in \mathcal{Y}} \mu_{Y|X}(y|B) \right] - g(\mu_{X,Y}, B)$ from (\ref{eq_TH1_1}) and 
the construction of $v_{\tilde{X},\tilde{Y}}$  in (\ref{eq_TH1_3}). The inequality in (\ref{eq_TH1_4}) is obtained as a function of $B\subset \mathcal{X}$,  as it is used to construct $v_{\tilde{X}, \tilde{Y}}$ in (\ref{eq_TH1_3}).

Returning to the main object of interest of this result,  we have that 
\begin{align}\label{eq_TH1_5}
\mathcal{I}( \mu_{(\tilde{U}, U_i),Y}) -  \mathcal{I}(\mu_{U_i,Y}) 
		= I(\tilde{U}; Y|U_i)
		&= \sum_{B_{i,j} \in \pi_i } \mu_X(B_{i,j}) \cdot  I(\tilde{U}; Y|X=B_{i,j}).
\end{align}
The first equality is by the chain rule of MI and the second is by definition of the conditional MI \cite{cover_2006}.  Finally 
we recognize that $I(\tilde{U}; Y|X=B)=\mathcal{I}(\mu_{\tilde{U}; Y|X}(\cdot |B))$, where $\mu_{\tilde{U}; Y|X}(\cdot |B)$
is precisely the distribution $v_{\tilde{X}, \tilde{Y}}$ defined in (\ref{eq_TH1_3}).  Consequently,  applying (\ref{eq_TH1_4}) in each $B_{i,j} \in \pi_i$,  we have that
\begin{equation*} 
\mathcal{I}( \mu_{(\tilde{U}, U_i),Y}) -  \mathcal{I}(\mu_{U_i,Y})  \geq \sum_{B_{i,j} \in \pi_i } \mu_X(B_{i,j}) \cdot \left[   \mathcal{H}(\mu_{Y|X}(\cdot |B_{i.j})) - \mathcal{H} (\mathcal{R}(\mu_{Y|X}( \cdot |B_{i,j}), \epsilon_{i,j} )) \right]
\end{equation*}
where $\epsilon_{i,j}=\left[   1- \max_{y\in \mathcal{Y}} \mu_{Y|X}(y|B_{i,j}) \right] - g(\mu_{X,Y}, B_{i,j})$.
\end{proof}

\subsection{Proof of Theorem \ref{th_1}: from Discrete to Continuous Representations} 
\label{sec_proof_th1}
\subsubsection{Overview}
The proof of Theorem \ref{th_1} is divided in two main stages. The first stage, presented in Section \ref{sec_discrete_version}, restricts the analysis to the important case of finite alphabet representations, or vector quantizers of $\mathcal{X}$.  In this discrete context, we use results from information theory to show that WIS implies OS: Theorem \ref{th_3} in Section \ref{sec_discrete_version}. The decision to begin studying the case of finite alphabet representations  was essential 
because it offers a path to adopt concrete results on the interplay between probability of error and conditional entropy  
only available for discrete random variables (Section \ref{sub_sec_finite_non_asymtotic}). 

In the second stage 
in Section \ref{sec_continuous}, we make a connection between the discrete and the continuous version of this problem.  Importantly, the finite alphabet result in Theorem \ref{th_3}  is used as a building block to extend the proof argument to the continuous case stated in Theorem \ref{th_1}.  For this objective,  results on information sufficient partitions for mutual information are used \cite{liese_2006}. 

\subsubsection{Discrete version of Theorem \ref{th_1}}
\label{sec_discrete_version}
\begin{theorem}\label{th_3}
	Let  $\left\{ U_i\right\}_{i\geq 1}$ be a sequence of representations for $X$ obtained from $\left\{ \eta_i\right\}_{i\geq 1}$ 
	where $ \left| \mathcal{U}_i \right|< \infty$ for any $i\geq 1$.  If $\left\{ U_i \right\}_{i\geq 1}$ is  WIS for $\mu_{X,Y}$ 
	then $\left\{ U_i, i\geq 1 \right\}$ is OS for $\mu_{X,Y}$. 
\end{theorem}
The proof of Theorem \ref{th_3} is presented in Section \ref{proof_th3}. 

Technical remarks about the proof of Theorem \ref{th_3}:
\begin{enumerate}
	\item 
	The proof of this result uses a sample-wise version of the inequality presented in (\ref{eq_main_3}) (Theorem \ref{th_2}) as a key element in the argument. 
	
	\item Another important technical element of the proof was characterizing and analyzing the following information object: 
	\begin{equation}\label{eq_discrete_version_1}
	\mathcal{I}_{loss}(\epsilon, M) \equiv \min_{v\in \mathcal{P}^\epsilon([M])}  \left\{ \mathcal{H}(v) - \mathcal{H} (\mathcal{R}(v, prior(v) - \epsilon)) \right\},
	\end{equation}
	where $\mathcal{P}^\epsilon([M]) \equiv  \left\{ v\in \mathcal{P}([M]), prior(v) \geq \epsilon \right\}$ and $M= \left| \mathcal{Y}\right|$.
	Indeed, a non-trivial  part of this argument was to prove that $\mathcal{I}_{loss}(\epsilon, M)>0$ for some values of $\epsilon>0$ (see Theorem \ref{lm_interplay_inf_oper_density} and Appendix \ref{append_proof_lm_interplay_inf_oper_density}). To achieve this key result, we derived an explicit lower bound for $\mathcal{I}_{loss}(\epsilon, M)$ function of $\epsilon$ and $M$.
\end{enumerate}

\subsubsection{Proof of Theorem \ref{th_1}}
\label{sec_continuous}
\begin{proof}
Without loss of generality,  let us assume that $\eta_i:  \mathcal{X}  \rightarrow \mathcal{U}_i$ is such that $\mathcal{U}_i\subset \mathcal{U}=\mathbb{R^q}$ for some $q\geq 1$\footnote{The general case derives directly from the argument presented for this case, and it only requires the introduction of additional notations that occludes the proof flow.}. Here we use a result from the seminal work of Liese, Morales and Vajda \cite{liese_2006} on asymptotic sufficient partition for MI.  In particular, in the context of our work we have the following:  

\begin{lemma} (Liese {\em et al.} \cite{liese_2006}) \label{lm_universal_sufficient_partition}
	There is an infinite collection of finite-size embedded partitions $\pi_1 \ll \pi_2 \l \ldots \subset \mathcal{B}(\mathbb{R}^q)$ of $\mathcal{U}=\mathbb{R}^q$ such that 
	for any model $\mu_{X,Y} \in \mathcal{P}(\mathcal{X} \times \mathcal{Y})$ and any measurable function $\eta: \mathcal{X}  \rightarrow \mathcal{U}$
	it follows that 
	\begin{equation}\label{eq_sec_continuous_1}
	\lim_{i \rightarrow \infty} I(Y;m_{\pi_i}(\eta(X))) =  I(Y;\eta(X)),
	\end{equation}
	where 
	\begin{equation}\label{eq_sec_continuous_2}
	m_{\pi_i}(u) \equiv \sum_{A_l\in \pi_i} l \cdot {\bf 1}_{A_l}(u) \in \left\{1,..,  \left| \pi_i \right|  \right\} , \ \forall u \in \mathcal{U}
	\end{equation}
	denotes the lossy function (vector quantizer) induced by the partition $\pi_i= \left\{ A_i, i=1,..,  \left| \pi_i \right|  \right\} $.
\end{lemma}
Lemma \ref{lm_universal_sufficient_partition} is a remarkable implication of the work by Liese {\em et al.} \cite{liese_2006}. This result shows the existence of a finite-size quantization family that approximates (universally) any well-defined MI on a continuous space in the sense presented in (\ref{eq_sec_continuous_1}). More details of this result and the construction of $\left\{ \pi_i \right\}_{i \geq 1}$ are presented in Section \ref{subsec_universal_quantization}.

In the context of this argument, we can use the universal embedded quantization $\left\{ \pi_i \right\}_{i \geq 1}$ of $\mathcal{U}$ stated in Lemma \ref{lm_universal_sufficient_partition} to obtain as a direct corollary of Lemma \ref{lm_universal_sufficient_partition} 
that for any $\eta_j:  \mathcal{X} \rightarrow \mathcal{U}$
	\begin{equation}\label{eq_sec_continuous_3}
		\lim_{i \rightarrow \infty} I((\tilde{U}, m_{\pi_i}(U_j));Y) - I(  m_{\pi_i}(U_j);Y) =   I((\tilde{U},U_j);Y) - I(U_j;Y) = I(\tilde{U};Y| U_j) \geq 0, 
	\end{equation}
where $U_j=\eta_j(X)$ and $\tilde{U} = \tilde{r}_{\mu_{X,Y}}(X)\in \mathcal{Y}$ (see Eq.(\ref{eq_sec_II_6})).
	
On the other hand,  from the hypothesis that assumes that  $\left\{\eta_j (\cdot) \right\}_{j\geq 1}$ is WIS,  we have that
	\begin{equation}\label{eq_sec_continuous_4}
	\lim_{j \rightarrow \infty} I(\tilde{U};Y| U_j=\eta_j(X)) =0.
	\end{equation}
	
Let us consider an arbitrary sequence $(\epsilon_n)_{n\geq 1} \in \mathbb{R}^+\setminus  \left\{ 0 \right\}$ such that 
$\epsilon_n \rightarrow 0$ as $n$ tends to infinity.  Using (\ref{eq_sec_continuous_3}),  
we have that for any $j \geq 1$ there exists $i^*_j(\epsilon_j, \eta_j)\geq 1$ sufficiently large such that\footnote{For what follows, we omitted the dependency on $\epsilon_j, \eta_j$ in $i^*_j$  to simplify the notation.}
	\begin{equation}\label{eq_sec_continuous_5}
	I(\tilde{U};Y| U_j) + \epsilon_j >  
	\underbrace{I((\tilde{U}, m_{\pi_{i^*_j}}(U_j));Y) - I(  m_{\pi_{i^*_j}}(U_j);Y)}_{I(\tilde{U};Y| m_{\pi_{i^*_j}}(U_j)))} > I(\tilde{U};Y| U_j) - \epsilon_j.
	\end{equation}
In (\ref{eq_sec_continuous_5}), it is worth noticing that $m_{\pi_{i^*_j}}(U_j)= m_{\pi_{i^*_j}} \circ  \eta_j (X)$.  
Then, we can define 
	\begin{equation}\label{eq_sec_continuous_5b}
	\tilde{\eta}_j \equiv m_{\pi_{i^*_j}} \circ  \eta_j: \mathcal{X} \rightarrow   \left\{1,..,  \left| \pi_{i^*_j} \right| <\infty \right\}, 
	\end{equation}
which is a finite alphabet representation (vector quantization) of $X$.  Therefore using $ \left\{\eta_j (\cdot) \right\}_{j\geq 1}$  and  $(\epsilon_n)_{n\geq 1}$, we have constructed a family of finite alphabet lossy representations of $X$,  which we denoted by $ \left\{ \tilde{\eta}_j(\cdot) \right\}_{j\geq 1}$ in (\ref{eq_sec_continuous_5b}), satisfying that
	\begin{equation}\label{eq_sec_continuous_6}
		\lim_{j \rightarrow \infty} I(\tilde{U};Y| \tilde{\eta}_{j}(X)))=0, 
	\end{equation}
from (\ref{eq_sec_continuous_5}), (\ref{eq_sec_continuous_4}), and the fact $(\epsilon_n)_{n\geq 1}$ is $o(1)$.
Therefore, (\ref{eq_sec_continuous_6}) means that $ \left\{ \tilde{\eta}_j (\cdot) \right\}_{j\geq 1}$ is {\em weakly information sufficient} (Def.\ref{def_wis}).  Then, Theorem \ref{th_3} implies that
	\begin{equation}\label{eq_sec_continuous_7}
		\lim_{j \rightarrow \infty}  \left[ \ell({\mu_{\tilde{\eta}_{j}(X),Y}}) - \ell({\mu_{X,Y}}) \right] =0. 
	\end{equation}
Finally, by construction, we have that $\tilde{\eta}_{j}(X)= m_{\pi_{i^*_j}} \circ  \eta_j (X)$. Then, $\tilde{\eta}_{j}(X)$  is indeed a deterministic function 
of $\eta_j(X)$ for any $j$. Therefore, from classical results on Bayes decision $\ell({\mu_{\tilde{\eta}_{j}(X),Y}}) \geq \ell({\mu_{\eta_{j}(X),Y}})$, 
which concludes the proof from (\ref{eq_sec_continuous_7}).
\end{proof}

\subsection{Proof of Theorem \ref{th_3}}
\label{proof_th3}
Let us begin introducing some preliminaries that will be used in the main argument in Section \ref{proof_th3_pre_main}. 

\subsubsection{Preliminaries}
\label{proof_th3_pre}
Let us consider a finite alphabet representation $\eta:\mathcal{X} \rightarrow \mathcal{U}$ where $ \left|\mathcal{U} \right| < \infty$.
Using the expressions presented in Propositions \ref{pro_opt_loss} and \ref{pro_inf_loss} and the interplay between 
them, determined in Theorem \ref{th_2}, we define the {\em information loss density} (ILD) and the {\em operation loss density} (OLD)
associated with $\eta (\cdot)$ as follows: 
	\begin{align} 
	\label{eq_proof_th3_1}
	\ell_{\eta}(x) & \equiv  \sum_{A\in \pi_\eta} {\bf 1}_{A}(x) \cdot g(\mu_{X,Y},A) \geq 0, \ \forall x \in \mathcal{X}  \\
	\label{eq_proof_th3_1b}
	\mathcal{I}_{\eta}(x) & \equiv \sum_{A\in \pi_\eta} {\bf 1}_{A}(x) \cdot I(\tilde{U}; Y | X\in A) \geq 0, \ \forall x \in \mathcal{X}.
	\end{align}
It is useful to denote by $\pi_\eta(x)$ the cell in $\pi_\eta$ that contains $x\in \mathcal{X}$. Using this notation, we have that 
$\ell_{\eta}(x)=g(\mu_{X,Y},\pi_\eta(x))$ and $\mathcal{I}_{\eta}(x)=I(\tilde{U}; Y | X\in \pi_\eta(x))$. The names of $\ell_{\eta}(\cdot)$ and $\mathcal{I}_{\eta}(\cdot)$ come from the observation that
	\begin{align} \label{eq_proof_th3_2}
	\mathbb{E}_X \left\{  \ell_{\eta}(X)  \right\} &= \ell(\mu_{U,Y}) -  \ell(\mu_{X,Y})\nonumber\\
	\mathbb{E}_X \left\{  \mathcal{I}_{\eta}(X)  \right\} &= \mathcal{I}( \mu_{(\tilde{U}, U),Y}) -  \mathcal{I}(\mu_{U,Y}), 
	\end{align} 
where $U=\eta(X)$.

From the proof of Theorem \ref{th_2},  we obtain the following sample-wise inequality: for any $A \in \mathcal{B}(\mathcal{X})$
	\begin{align} \label{eq_proof_th3_3}
	I(\tilde{U}; Y | X\in A) \geq \mathcal{H}(\mu_{Y|X}(\cdot |A)) -  \mathcal{H} (\mathcal{R}(\mu_{Y|X}(\cdot |A), prior(\mu_{Y|X}( \cdot |A)) - g(\mu_{X,Y},A))), 
	\end{align} 
where $prior(\mu_Y) \equiv (1-\max_{y\in \mathcal{Y}} \mu_Y(y))$ denotes the prior risk of a prior model $\mu_Y\in \mathcal{P}(\mathcal{Y})$.
Adopting this inequality, it follows that for any $x\in \mathcal{X}$
	\begin{align} \label{eq_proof_th3_4}
		\mathcal{I}_{\eta}(x) \geq  \mathcal{H}(\mu_{Y|X}( \cdot |\pi_\eta(x))) - \mathcal{H} (\mathcal{R}(\mu_{Y|X}(\cdot |\pi_\eta(x)), prior(\mu_{Y|X}( \cdot |\pi_\eta(x))) - \ell_{\eta}(x))).
	\end{align} 
Then the ILD $\mathcal{I}_{\eta}(x)$ is lower bounded by a function of the posterior model $\mu_{Y|X}(\cdot |\pi_\eta(x)) \in \mathcal{P}(\mathcal{Y})$ and the gain of observing $\tilde{U}$ when the prior distribution on $\mathcal{Y}$ is $\mu_{Y|X}(\cdot |\pi_\eta(x))$, i.e.,  
	$$\left[ prior(\mu_{Y|X}(\cdot |\pi_\eta(x))) - \ell_{\eta}(x))  \right] = \sum_{A^*_u\in \pi^*} \mu_X(A^*_u|A) \cdot  \left[ 1-\max_{y\in \mathcal{Y}} \mu_{Y|X}(y| A^*_u \cap A) \right] \geq 0.$$ 
	
Let us assume that we have a family of WIS representations for $\mu_{X,Y}$ (Definition \ref{def_wis}) given by $ \left\{ \eta_i (\cdot) \right\}_{i\geq 1}$ 
where $\eta_i:\mathcal{X} \rightarrow \mathcal{U}_i$ and $ \left|\mathcal{U}_i \right| < \infty$ for any $i$. Using the definition of the ILD in (\ref{eq_proof_th3_1b}) and (\ref{eq_proof_th3_2}), it follows that  
	\begin{align} \label{eq_proof_th3_5}
		\lim_{i \longrightarrow \infty} \mathbb{E}_X \left\{  \mathcal{I}_{\eta_i}(X)  \right\}=0.
	\end{align} 
As $\mathcal{I}_{\eta_i}(x) \leq \log \left| \mathcal{Y} \right|$ (uniformly in $i$ and $x$), the convergence in (\ref{eq_proof_th3_5}) is equivalent 
to the convergence in probability of $(\mathcal{I}_{\eta_i}(X))_{i\geq 1}$, i.e., $\forall \epsilon>0$ it follows that  $\lim_{i \rightarrow \infty} \mathbb{P} \left( \left\{  \mathcal{I}_{\eta_i}(X) > \epsilon \right\} \right) =0$. 

Using again (\ref{eq_proof_th3_2}), the proof reduces to verify that 
	\begin{align} \label{eq_proof_th3_6}
		\lim_{i \longrightarrow \infty} \mathbb{E}_X \left\{ \ell_{\eta_i}(X) \right\}=0.
	\end{align} 
Again $\ell_{\eta_i}(x)$ is uniformly bounded by $1$, then the convergence in (\ref{eq_proof_th3_6}) is equivalent 
to the convergence in probability of the random sequence $(\ell_{\eta_i}(X))_{i\geq 1}$, i.e., for any $\epsilon>0$
	\begin{align} \label{eq_proof_th3_7}
	\lim_{i \rightarrow \infty} \mathbb{P} \left( \left\{  \ell_{\eta_i}(X) > \epsilon \right\} \right) =0. 	
	\end{align}
	
\subsubsection{Main Argument}
\label{proof_th3_pre_main}
\begin{proof}
Let us prove the result by contradiction. Let us assume that $\left\{\eta_i (\cdot) \right\}_{i\geq 1}$ is not OS. Then, there exists $\epsilon>0$
such that $\lim \inf_{i \rightarrow \infty}  \mu_X (B^i_{\epsilon})>0$  where $B^i_{\epsilon} \equiv  \left\{x \in \mathcal{X}, \ell_{\eta_i}(x) > \epsilon \right\} \subset \mathcal{X}$.
Then, we can pick  $\delta>0$ and $N>0$,  such that for any $i\geq N$,  
	\begin{align} \label{eq_proof_th3_7}
		\mu_X(B^i_{\epsilon}) \geq \delta.
	\end{align} 
Using the definition of the function $\mathcal{R}(v,\epsilon)$ (see Appendix \ref{appendix_rate_distortion_function}),  for any $v\in \mathcal{P}(\mathcal{Y})$, it 
follows --- from the expression of $f(v,\epsilon)$ in (\ref{eq_main_1}) --- that $\mathcal{H}(\mathcal{R}(v,\epsilon_1)) \geq \mathcal{H}(\mathcal{R}(v,\epsilon_2))$ when $\epsilon_1\geq \epsilon_2$; therefore, from (\ref{eq_proof_th3_4}),  if $x\in B^i_{\epsilon}$,  we have that 
	\begin{align} \label{eq_proof_th3_8}
		\mathcal{I}_{\eta_i}(x)  \geq   \mathcal{H}(\mu_{Y|X}(\cdot|\pi_i(x))) - \mathcal{H} (\mathcal{R}(\mu_{Y|X}(\cdot|\pi_i(x)), prior(\mu_{Y|X}( \cdot |\pi_i(x))) - \epsilon))
	\end{align}  
	where $\pi_i(x)$ is a shorthand for $\pi_{\eta_i}(x)$.
	
The bound in (\ref{eq_proof_th3_8}) will be central to prove the result: a lower bound on the information loss density function of the operation loss density that is lower bounded by $\epsilon>0$. 
More precisely, given $\epsilon>0$, we proceed by finding a uniform lower bound for 
	\begin{align}  \label{eq_proof_th3_8b}
		\mathcal{H}(v) - \mathcal{H} (\mathcal{R}(v, prior(v) - \epsilon))
	\end{align}  
over all models $v\in \mathcal{P}(\mathcal{Y})$ that are admissible in the sense that $prior(v) \geq \epsilon$. 

In particular, we will consider the following general information vs. operation loss problem:
	\begin{align} \label{eq_proof_th3_9}
		\mathcal{I}_{loss}(\epsilon, M) \equiv \min_{v\in \mathcal{P}^\epsilon([M])}  \left\{ \mathcal{H}(v) - \mathcal{H} (\mathcal{R}(v, prior(v) - \epsilon)) \right\},  
	\end{align}
	where 
	\begin{align} \label{eq_proof_th3_9b}
	\mathcal{P}^\epsilon([M]) \equiv  \left\{ v\in \mathcal{P}([M]), prior(v) \geq \epsilon \right\}.
	\end{align}
	In this notation, we use $\mathcal{Y}=[M] \equiv  \left\{1,..,M \right\}$ to make explicit the role that the cardinality of $\mathcal{Y}$ plays in this analysis. 
	Importantly, we have the following (information loss vs. operation loss) interplay result that shows that a non-zero operation loss ($\epsilon>0$)
	implies a positive information loss for any $M\geq 1$: 
	\begin{theorem}\label{lm_interplay_inf_oper_density}
		$\forall M\geq 1$, and for any $\epsilon\in (0, 1-1/M]$, it follows that $\mathcal{I}_{loss}(\epsilon, M)>0.$
	\end{theorem}
	The proof of this result requires (non-trivial) technical elements that are presented in Section \ref{append_proof_lm_interplay_inf_oper_density}.
	
	Returning to the main proof argument,  by definition of the operation loss density  in (\ref{eq_proof_th3_1}) , we have that $\ell_{n_i}(x) \leq prior(\mu_{Y|X}( \cdot |\pi_i(x)))$, which implies that 
	$\mu_{Y|X}(\cdot |\pi_i(x))\in \mathcal{P}^{\ell_{n_i}(x)}([M])$ in (\ref{eq_proof_th3_9b}). Then using (\ref{eq_proof_th3_8}) and (\ref{eq_proof_th3_9}), 
	for any $x \in B^i_\epsilon$ (considering that $\epsilon< \ell_{\eta_i}(x)$ if $x \in B^i_\epsilon$)
	\begin{align} \label{eq_proof_th3_10}
	\mathcal{I}_{\eta_i}(x)  	&\geq   \mathcal{H}(\mu_{Y|X}( \cdot |\pi_i(x))) - \mathcal{H} (\mathcal{R}(\mu_{Y|X}( \cdot |\pi_i(x)), prior(\mu_{Y|X}(\cdot |\pi_i(x))) - \epsilon)) \nonumber\\
						&\geq   \min_{v\in \mathcal{P}^\epsilon([M])}  \left\{ \mathcal{H}(v) - \mathcal{H} (\mathcal{R}(v, prior(v) - \epsilon)) \right\}= \mathcal{I}_{loss}(\epsilon, M), 			
	\end{align}  
	where the second inequality comes from the observation that $\mu_{Y|X}(\cdot |\pi_i(x))\in \mathcal{P}^{\ell_{n_i}(x)}([M]) \subset \mathcal{P}^{\epsilon}([M])$ from (\ref{eq_proof_th3_9b}). 
	
At this point,  we use Theorem \ref{lm_interplay_inf_oper_density}: we have that for any $x\in B^i_\epsilon$,  $\mathcal{I}_{\eta_i}(x)\geq \mathcal{I}_{loss}(\epsilon, M)>0$. In particular, we have that for any $\bar{\epsilon} \in (0,\mathcal{I}_{loss}(\epsilon, M))$, $B^i_\epsilon \subset A^i_{\bar{\epsilon}} \equiv  \left\{x \in \mathcal{X}, \mathcal{I}_{\eta_i}(x) > \bar{\epsilon} \right\}$.   Then using the hypothesis in (\ref{eq_proof_th3_7}), we have that  for any $i\geq N$  $\mu_x(A^i_{\bar{\epsilon}}) \geq \mu_x(B^i_{\bar{\epsilon}}) \geq \delta>0$. This implies that $(\mathcal{I}_{\eta_i}(x))_{i \geq 1}$ does not converge to zero in probability, which from the argument presented in Section \ref{proof_th3_pre} contradicts the fact that 
	$\left\{\eta_i (\cdot) \right\}_{i\geq 1}$ is WIS. This concludes the proof of Theorem \ref{th_3}. 
\end{proof}	

\subsection{Proof of Theorem \ref{lm_interplay_inf_oper_density}}
\label{append_proof_lm_interplay_inf_oper_density}
\begin{proof}
Given a probability $\mu\in \mathcal{P}^\epsilon([M])$, Ho {\em et al.}  \cite{ho_2010} presented a closed-form analytical expression 
for $\mathcal{R}(\mu, prior(\mu)-\epsilon)$ (see details in Appendix \ref{appendix_rate_distortion_function}) appearing  in the definition of $\mathcal{I}_{loss}(\epsilon, M)$  in (\ref{eq_proof_th3_9}).  To present this induced distribution more clearly, 
we assume, without loss of generality,  that $\mu(1) \geq \mu(2) \geq \ldots \geq \mu(M)$.  Then $\mu^\epsilon \equiv \mathcal{R}(\mu, prior(\mu)-\epsilon)$
has the following structure:\footnote{To simplify notation $\mu(j)$ denotes $\mu(\left\{ j \right\})$, i.e.,$\mu(j)$ is a short-hand of the probability mass function (pmf).}
	\begin{align}
	\label{eq_app_proof_lm_int_inf_oper_density_1a}
	&\mu^\epsilon(1)= \mu(1) + \epsilon \leq 1 \\
	\label{eq_app_proof_lm_int_inf_oper_density_1b}
	&\mu^\epsilon(2)=  \theta\nonumber\\ 
	&\ldots	\nonumber\\ 
	&\mu^\epsilon(K)=  \theta\\ 
	\label{eq_app_proof_lm_int_inf_oper_density_1c}
	&\mu^\epsilon(K+1) = \mu(K+1) \nonumber\\ 
	&\ldots	\nonumber\\ 
	&\mu^\epsilon(M) = \mu(M). 
	\end{align}
where both $K\in  \left\{2,..,M \right\}$ and $\theta \in (0,\mu(1))$ are functions of $\mu$ and $\epsilon>0$ satisfying the following condition: 
	\begin{align}\label{eq_app_proof_lm_int_inf_oper_density_2}
	\sum_{j=2}^K  \left(  \mu(i)-\theta \right) = \epsilon>0, 
	\end{align}
which makes $\mu^\epsilon$ a well-defined probability in $\mathcal{P}([M])$.\footnote{Ho {\em et al.} \cite{ho_2010} show that for any $\epsilon \leq prior(\mu)$, $\exists \theta \in [0,\mu(1))$ and $K\in \left\{2,..,M \right\}$ that meet the condition in (\ref{eq_app_proof_lm_int_inf_oper_density_2}).}

Therefore,  using (\ref{eq_app_proof_lm_int_inf_oper_density_1a}), (\ref{eq_app_proof_lm_int_inf_oper_density_1b}) and (\ref{eq_app_proof_lm_int_inf_oper_density_1c}), we have that for any $\mu\in \mathcal{P}^\epsilon([M])$:
	\begin{align}\label{eq_app_proof_lm_int_inf_oper_density_3}
	\mathcal{H}(\mu) -  \mathcal{H}(\mu^\epsilon) 	&=  \mu(1) \log \frac{1}{ \mu(1)} - (\mu(1) + \epsilon) \log \frac{1}{ \mu(1) + \epsilon}  \nonumber\\
										&+ \sum^{K(\mu,\epsilon)}_{j=2} \mu(j) \log \frac{1}{ \mu(j)} -  (K(\mu,\epsilon)-1) \theta(\mu,\epsilon) \log \frac{1}{\theta(\mu,\epsilon)}, 
	\end{align}
	where here we make explicit the dependency of $K$ and $\theta$ on $\mu$ and $\epsilon$. 
	It is important 
	to note that by construction $\theta(\mu,\epsilon) < \mu(K) \leq \mu(K-1) \ldots \leq \mu(1)$. At this point, we will use the following result: 
\begin{lemma}\label{lm_maximum_entropy}
$\forall \epsilon>0$ and for any $\mu \in \mathcal{P}^\epsilon([M])$, it follows that 
	\begin{align}\label{eq_app_proof_lm_int_inf_oper_density_4}
	\sum^{K(\mu,\epsilon)}_{j=2} \mu(j) \log \frac{1}{ \mu(j)} \geq (\theta(\mu,\epsilon) + \epsilon) \log \frac{1}{\theta(\mu,\epsilon) + \epsilon} + (K(\mu,\epsilon)-2)  \theta(\mu,\epsilon) \log \frac{1}{\theta(\mu,\epsilon)}.
	\end{align}	
\end{lemma}
The proof is presented in Appendix \ref{appendix_lm_maximum_entropy}. 
\begin{remark}
	The proof of Lemma \ref{lm_maximum_entropy} comes from the use of some information-theoretic inequalities, similar to the 
	argument used to prove that the Shannon entropy (over a finite alphabet) is minimized with a degenerated distribution \cite{cover_2006,gray_1990_b}.
\end{remark}

Applying Lemma \ref{lm_maximum_entropy}, we have that for all $\mu\in \mathcal{P}^\epsilon([M])$:
	\begin{align}\label{eq_app_proof_lm_int_inf_oper_density_5}
	\mathcal{H}(\mu) -  \mathcal{H}(\mu^\epsilon) 	&\geq \mu(1) \log \frac{1}{ \mu(1)} - (\mu(1) + \epsilon) \log \frac{1}{ \mu(1) + \epsilon}  \nonumber\\
										&+\left[  (\theta(\mu,\epsilon) + \epsilon)\log \frac{1}{\theta(\mu,\epsilon) + \epsilon} - \theta(\mu,\epsilon) \log \frac{1}{\theta(\mu,\epsilon)} \right]. 
	\end{align}
Using the fact that $\theta(\mu,\epsilon) < \mu(K) \leq \mu(K-1) \ldots \leq \mu(1)$, and that $\sum_{j=2}^{K(\mu,\epsilon)} (\mu(j)-\theta(\mu,\epsilon))=\epsilon$, it is simple to verify that\footnote{This because $\mu(2)- \theta(\mu,\epsilon) \geq \mu(3)- \theta(\mu,\epsilon) \geq  \ldots \geq \mu(K)- \theta(\mu,\epsilon)>0$.} 
	\begin{align}\label{eq_app_proof_lm_int_inf_oper_density_5b}
	\mu(2)- \theta(\mu,\epsilon) \geq \frac{\epsilon}{K-1},
	\end{align}
which implies that $\theta(\mu,\epsilon) \leq \mu(2) - \epsilon/(K-1).$

On the other hand, if we consider the following function used in (\ref{eq_app_proof_lm_int_inf_oper_density_5}): 
	\begin{align} \label{eq_app_proof_lm_int_inf_oper_density_6}
	f_1(\theta, \epsilon) \equiv (\theta + \epsilon)\log \frac{1}{\theta + \epsilon} - \theta \log \frac{1}{\theta}, 
	\end{align}
$\frac{\partial f_1(\theta, \epsilon)}{\partial \theta} (\theta, \epsilon) = \log \frac{\theta}{\theta + \epsilon} <0$, then $f_1(\theta, \epsilon)$
is strictly decreasing in the domain $\theta >0$, for any $\epsilon>0$. Therefore from (\ref{eq_app_proof_lm_int_inf_oper_density_5b}), we have
that $f_1(\theta(\mu,\epsilon), \epsilon) \geq f_1(\mu(2)- {\epsilon}/{(K-1)}, \epsilon)$. Applying this last inequality in (\ref{eq_app_proof_lm_int_inf_oper_density_5}), we have that
	\begin{align}\label{eq_app_proof_lm_int_inf_oper_density_7}
	\mathcal{H}(\mu) -  \mathcal{H}(\mu^\epsilon) 	&\geq  - f_1(\mu(1), \epsilon) + f_1(\theta(\mu,\epsilon), \epsilon)\nonumber\\
										&\geq   - f_1(\mu(1), \epsilon) + f_1(\mu(2)- {\epsilon}/{(K-1)}, \epsilon).
	\end{align}
Furthermore, $\mu(2) - \epsilon/(K-1) \leq \mu(2) - \epsilon/(M-1)$, which offers a bound that is independent of $K(\mu, \epsilon)$. 
Finally,  we have that 
	\begin{align}\label{eq_app_proof_lm_int_inf_oper_density_8}
	\mathcal{H}(\mu) -  \mathcal{H}(\mu^\epsilon) 
				\geq   - f_1(\mu(1), \epsilon) + f_1(\mu(2)- {\epsilon}/{(M-1)}, \epsilon).
	\end{align}

At this point, we return to our main problem:
	\begin{align}\label{eq_app_proof_lm_int_inf_oper_density_9}
	&\mathcal{I}_{loss}(\epsilon, M) = \min_{\mu \in \mathcal{P}^\epsilon([M])} \mathcal{H}(\mu) - \mathcal{H}(\mu^\epsilon) \nonumber\\
							&\geq \min_{\mu(1)\in [1/M,1-\epsilon]} \left(   - f_1(\mu(1), \epsilon) + \min_{{\mu(2) \in [0, \min \left\{\mu(1), 1-\mu(1) \right\}]}} \left( f_1(\mu(2)- {\epsilon}/{(M-1)}, \epsilon) \right)  \right),  	
	\end{align}
where the lower bound in (\ref{eq_app_proof_lm_int_inf_oper_density_9}) comes from (\ref{eq_app_proof_lm_int_inf_oper_density_8}) and the fact that $\mu(1) = \max \left\{ \mu(j), j \in [M] \right\} \in [1/M,1-\epsilon]$ if $\mu \in \mathcal{P}^\epsilon([M])$. For the rest of the proof, we concentrate on the analysis of the RHS of (\ref{eq_app_proof_lm_int_inf_oper_density_9}), where we recognize for the second optimization in (\ref{eq_app_proof_lm_int_inf_oper_density_9}) two scenarios.

{\bf Case 1} (the restriction $\mu(2)\leq \mu(1)$ is active in (\ref{eq_app_proof_lm_int_inf_oper_density_9})): If we restrict the second optimization problem in (\ref{eq_app_proof_lm_int_inf_oper_density_9}) to the case where $\mu(1)\leq 1- \mu(1)$,  this scenario implies that $\mu(1)\leq \frac{1}{2}$. In addition, we have that  $\mu(1) \geq 1/M$ (achieved for the case of a uniform distribution in $[M]$). Then under this hypothesis, it follows that 
	\begin{align}\label{eq_app_proof_lm_int_inf_oper_density_10}
	\mathcal{I}_{loss}(\epsilon, M) \geq \min_{\mu(1)\in [1/M, 1/2]} 	- f_1(\mu(1), \epsilon)  + f_1(\mu(1) - \epsilon/(M-1), \epsilon), 
	\end{align}
	the last bound from (\ref{eq_app_proof_lm_int_inf_oper_density_9}) using the fact that $f_1(x, \epsilon)$ is strictly decreasing for $x\in (0,\infty)$ for any $\epsilon>0$. Let us define $\tilde{f}(x, \epsilon) \equiv - f_1(x, \epsilon) + f_1(x - \epsilon/(M-1), \epsilon)$. It is simple to verify that $\frac{\partial \tilde{f}(x, \epsilon) }{\partial x}< 0$ for any $x>0$\footnote{$\frac{\partial \tilde{f}(x, \epsilon) }{\partial x}=\log \frac{\psi_\epsilon(x)}{\psi_\epsilon(x-\epsilon/(M-1))}<0$ for any $x>0$,  where $\psi_\epsilon(x) \equiv (1+\epsilon/x)$.}.
	This implies that
	\begin{align}\label{eq_app_proof_lm_int_inf_oper_density_11}
	\mathcal{I}_{loss}(\epsilon, M) \geq \tilde{f}(1/2,\epsilon) = f_1\left(1/2-\frac{\epsilon}{M-1}, \epsilon \right) - f_1\left(1/2, \epsilon \right) >0, 
	\end{align}
	using again that $(f_1\left(x, \epsilon \right))_{x>0}$ is strictly decreasing for any $\epsilon>0$.
	
{\bf Case 2} (the restriction $\mu(2) \leq 1- \mu(1)$ is active in (\ref{eq_app_proof_lm_int_inf_oper_density_9})): If we restrict the second optimization problem in (\ref{eq_app_proof_lm_int_inf_oper_density_9}) to the case where $1- \mu(1) <  \mu(1)$,  this scenario implies that $\mu(1)>\frac{1}{2}$. In addition, 
as $\mu \in \mathcal{P}^\epsilon([M])$, it follows that $\mu(1) \leq 1 -\epsilon$.  Therefore, under this hypothesis, 
	\begin{align}\label{eq_app_proof_lm_int_inf_oper_density_12}
	\mathcal{I}_{loss}(\epsilon, M) \geq \min_{\mu(1)\in (1/2,1 - \epsilon]} - f_1(\mu(1), \epsilon)  + f_1((1- \mu(1)) - \epsilon/(M-1), \epsilon), 
	\end{align}
	the last bound from (\ref{eq_app_proof_lm_int_inf_oper_density_9}) using the fact that $(f_1(x, \epsilon))_{x\in (0,\infty)}$ is strictly decreasing for any $\epsilon>0$. In this case, we consider $\tilde{\phi}(x, \epsilon) \equiv  - f_1(x, \epsilon)  + f_1((1- x) - \epsilon/(M-1), \epsilon)$.  It is simple to verify  
	that $\frac{ \partial \tilde{\phi}(x, \epsilon)}{\partial x} >0$ for any $x>0$. Consequently, we have that 
	\begin{align}\label{eq_app_proof_lm_int_inf_oper_density_13}
	\mathcal{I}_{loss}(\epsilon, M) \geq \tilde{\phi}( 1/2, \epsilon) = f_1\left(1/2-\frac{\epsilon}{M-1}, \epsilon \right) - f_1\left(1/2, \epsilon \right) >0. 
	\end{align}
	Interestingly in (\ref{eq_app_proof_lm_int_inf_oper_density_13}) and (\ref{eq_app_proof_lm_int_inf_oper_density_11}), we arrived to the 
	same positive closed-form lower bound for $\mathcal{I}_{loss}(\epsilon, M)$, which concludes the proof of Theorem \ref{lm_interplay_inf_oper_density}. 
\end{proof}

\subsection{Proof of Theorem \ref{th_OS_imply_WIS}}
\label{proof_th_OS_imply_WIS}
The proof of this result divided in two stages. First, we show that under the uniqueness assumption of $\tilde{r}_{\mu_{X,Y}} (\cdot)$ (Def. \ref{def_unique_map_rule}), the OS condition implies that $\lim_{i \rightarrow \infty} \ell(\mu_{U_i,\tilde{U}})=0$, where $\tilde{U}=\tilde{r}_{\mu_{X,Y}} (X)\in \mathcal{Y}$ in the MPE predictor of $Y$.  The second stages used a refined version of  {\em Fano's inequality}  stated in \cite[Th.1]{feder_1994} to prove that $\lim_{i \rightarrow \infty} \ell(\mu_{U_i,\tilde{U}})=0$ implies that $\lim_{i \rightarrow \infty}  I(\tilde{U};Y|U_i)=0$. Finally, the equivalence stated in Theorem \ref{th_OS_imply_WIS} is obtained from our result in Theorem \ref{th_1}.

\subsubsection{Stage 1: $\lim_{i\rightarrow \infty} \ell(\mu_{U_i,Y})=\ell(\mu_{X,Y}) \Rightarrow \lim_{i\rightarrow \infty} \ell(\mu_{U_i,\tilde{U}})=0$}
\begin{proof}  
	For the MPE decision rule, we use the expression of $\tilde{r}_{\mu_{X,Y}} (\cdot)$ in (\ref{eq_sec_II_6}) and its induced partition $\pi^*= \left\{A^*_y,y\in \mathcal{Y} \right\}$ (with $M$ cells) in (\ref{eq_sec_II_7}). On the same line, we can introduce: 
\begin{equation} \label{eq_proof_th_OS_WIS_1}
	\tilde{r}_{\mu_{U_i,Y}} (u) \equiv  \arg \max_{y\in \mathcal{Y}} \mu_{Y|U_i}(y|u).
\end{equation}
and
\begin{equation} \label{eq_proof_th_OS_WIS_2}
	\pi^i \equiv  \left\{ A^i_y \equiv \eta_i^{-1}(\tilde{r}_{\mu_{U_i,Y}}^{-1}(\left\{ y \right\})), y \in \mathcal{Y} \right\} \subset \mathcal{B}(\mathcal{X}).
\end{equation}
From these, we have that $\ell(\mu_{X,Y})=\sum_{j=1}^M (1-\mu_{X|Y}(A^*_j|j)) \mu_{Y}(j)$ and  $\ell(\mu_{U_i,Y})=\sum_{j=1}^M (1-\mu_{X|Y}(A^i_j|j)) \mu_{Y}(j)$. Then, the operation loss of $U_i$ (or $\eta_i(\cdot)$) can be expressed by: 
\begin{align} \label{eq_proof_th_OS_WIS_3}
	\ell(\mu_{U_i,Y}) - \ell(\mu_{X,Y}) 	&=\sum_{j=1}^M (\mu_{X|Y}(A^*_j|j) -  \mu_{X|Y}(A^i_j|j)) \cdot  \mu_{Y}(j) \nonumber\\
								&= \sum_{j=1}^M (\mu_{X,Y}(A^*_j \times  \left\{ j \right\}) -  \mu_{X,Y}(A^i_j \times  \left\{ j \right\})
\end{align}
On the other hand, our object of interest is $\ell(\mu_{U_i,\tilde{U}})$, where we have that 
\begin{align} \label{eq_proof_th_OS_WIS_4}
	\ell(\mu_{U_i,\tilde{U}}) 	&\leq \mathbb{P}(\tilde{r}_{\mu_{U_i,Y}} (U_i)  \neq \tilde{U}) = \mathbb{P}(\tilde{r}_{\mu_{U_i,Y}} (\eta_i(X)) \neq \tilde{U}) \nonumber\\
						&=1 - \mathbb{P}(\tilde{r}_{\mu_{U_i,Y}} (\eta_i(X)) = \tilde{r}_{\mu_{X,Y}} (X)) = 1- \sum_{j=1}^M \mu_X (A^*_j\cap A^i_j ) \nonumber\\
						&= \sum_{j=1}^M (\mu_X(A_j^*)- \mu_X (A^*_j\cap A^i_j ))= \sum_{j=1}^M \mu_X (A^*_j \setminus  A^i_j ),
\end{align}
where the first inequality comes from the definition of the MPE rule and the third equality from the fact that $\tilde{r}_{\mu_{U_i,Y}}(\eta_i(x))=\sum_{j=1}^M {\bf 1}_{A^i_j}(x) \cdot j$ and $\tilde{r}_{\mu_{X,Y}}(x)=\sum_{j=1}^M {\bf 1}_{A^*_j}(x) \cdot j$. 

To upper bound (\ref{eq_proof_th_OS_WIS_4}), let us work with the information loss expression in (\ref{eq_proof_th_OS_WIS_3}). It follows that
\begin{align} \label{eq_proof_th_OS_WIS_5}
	\ell(\mu_{U_i,Y}) - \ell(\mu_{X,Y}) =  \sum_{j=1}^M  \mu_{X,Y}(A^*_j \setminus A^i_j  \times  \left\{ j \right\}) -  \sum_{\tilde{j}=1}^M \mu_{X,Y}(A^i_{\tilde{j}} \setminus A^*_{\tilde{j}}  \times  \left\{ \tilde{j} \right\}).
\end{align} 
Using the fact that $\bigcup_{j=1}^M A^*_j \setminus A^i_j = \bigcup_{\tilde{j}=1}^M A^i_{\tilde{j}} \setminus A^*_{\tilde{j}}$,
then for any $j\in \mathcal{Y}$, it follows that  $A^i_{\tilde{j}} \setminus A^*_{\tilde{j}} = (\bigcup^M_{j=1} A^*_j \setminus A^i_j) \cap A^i_{\tilde{j}} \setminus A^*_{\tilde{j}}$.
From this last identity, we have that:
\begin{align} \label{eq_proof_th_OS_WIS_6}
	\ell(\mu_{U_i,Y}) - \ell(\mu_{X,Y}) = \sum_{j=1}^M   \left[  \mu_{X,Y}(A^*_j \setminus A^i_j  \times  \left\{ j \right\}) - \sum^M_{\tilde{j}=1, \tilde{j}\neq j}  \mu_{X,Y}(A^i_{\tilde{j}} \setminus A^*_{\tilde{j}} \cap A^*_j \setminus A^i_j  \times  \left\{ \tilde{j} \right\}) \right] 
\end{align} 
Let us analize one of the terms in the RHS of (\ref{eq_proof_th_OS_WIS_6}), 
\begin{align} \label{eq_proof_th_OS_WIS_7}
&\mu_{X,Y}(A^*_j \setminus A^i_j  \times  \left\{ j \right\}) - \sum^M_{\tilde{j}=1, \tilde{j}\neq j}  \mu_{X,Y}(A^i_{\tilde{j}} \setminus A^*_{\tilde{j}} \cap A^*_j \setminus A^i_j  \times  \left\{ \tilde{j} \right\})\nonumber\\ 
&= \int_{A^*_j \setminus A^i_j } f_{X,Y}(x,j) dx - \sum_{\tilde{j}=1, \tilde{j}\neq j}^M \int_{A^i_{\tilde{j}} \setminus A^*_{\tilde{j}} \cap A^*_j \setminus A^i_j} f_{X,Y}(x, \tilde{j}) dx \nonumber\\
	&\geq  \int_{A^*_j \setminus A^i_j }  \left[ \underbrace{ f_{X,Y}(x,j) - \max_{\tilde{j} \in \mathcal{Y}, \tilde{j} \neq j }f_{X,Y}(x, \tilde{j})}_{\geq 0} \right]   dx \geq 0,
\end{align} 
where $f_{X,Y}(x,y)$ denotes the density of $\mu_{X,Y}$.  The last inequality  in  (\ref{eq_proof_th_OS_WIS_7}) comes from the definition of the MPE, 
the fact that for any $x\in A^*_j \setminus A^i_j$, $f_{X,Y}(x,j)=\max_{y \in \mathcal{Y}} f_{X,Y}(x,y)$, the fact that for any $x\in A^i_{\tilde{j}} \setminus A^*_{\tilde{j}} \cap A^*_j \setminus A^i_j$, $f_{X,Y}(x, \tilde{j}) \leq  \max_{y \in \mathcal{Y}, y \neq j }f_{X,Y}(x, y)$, and the fact that $A^*_{j} \setminus A^i_{j} = \bigcup^M_{\tilde{j}=1} A^*_j \setminus A^i_j \cap A^i_{\tilde{j}} \setminus A^*_{\tilde{j}}$.  Finally, using (\ref{eq_proof_th_OS_WIS_7}) in (\ref{eq_proof_th_OS_WIS_6}),  we have that 
\begin{align} \label{eq_proof_th_OS_WIS_8}
	\ell(\mu_{U_i,Y}) - \ell(\mu_{X,Y}) \geq \sum_{j=1}^M \int_{A^*_j \setminus A^i_j } (f_{X,Y}(x,j) - \max_{\tilde{j} \in \mathcal{Y}, \tilde{j} \neq j }f_{X,Y}(x, \tilde{j}))  dx.  
\end{align} 

{\bf Proving the result under a strong condition on $\mu_{X,Y}$:}\\
For simplicity and clarity, let us assume for the moment the following discrimination condition on $\mu_{X,Y}$: 
$\exists K>0$ and $\exists A\subset \mathcal{X}$ s.t. $\forall x\in A$
\begin{align} \label{eq_proof_th_OS_WIS_9}
	\max_{y\in \mathcal{Y}} f_{X,Y}(x,y) - \max_{y\in \mathcal{Y}, y \neq \tilde{r}_{\mu_{X,Y}}(x)} f_{X,Y}(x,y)\geq K\cdot f_X(x)
\end{align} 
where $f_X(x)=\sum_{y\in \mathcal{Y}}f_{X,Y}(x,y)$ is the marginal density of $X$ and $\mu_X(A)=1$. This strong 
discrimination assumption on $\mu_{X,Y}$ is instrumental to directly prove our result.  Indeed, under this assumption, we have that 
\begin{align} \label{eq_proof_th_OS_WIS_10}
	\ell(\mu_{U_i,Y}) - \ell(\mu_{X,Y}) &\geq \sum_{j=1}^M \int_{(A^*_j \setminus A^i_j )\cap A} (f_{X,Y}(x,j) - \max_{\tilde{j} \in \mathcal{Y}, \tilde{j} \neq j }f_{X,Y}(x, \tilde{j}))  dx \\
	&\geq K \cdot  \sum_{j=1}^M \int_{(A^*_j \setminus A^i_j )\cap A} f_{X}(x) dx = K \cdot \sum_{j=1}^M \mu_X((A^*_j \setminus A^i_j )\cap A)\\
	&= K \cdot  \sum_{j=1}^M \mu_X(A^*_j \setminus A^i_j ) \geq K\cdot  \ell(\mu_{U_i,\tilde{U}}).
\end{align}
The first inequality comes from (\ref{eq_proof_th_OS_WIS_8}) and the fact that $(A^*_j \setminus A^i_j )\cap A \subset A^*_j \setminus A^i_j$ for every $j\in \mathcal{Y}$ and the observation that the function been integrated is non-negative.  The second inequality comes from the discrimination condition stated in (\ref{eq_proof_th_OS_WIS_9}).  The second equality comes from the assumption that  $\mu_X(A)=1$, and the last inequality from (\ref{eq_proof_th_OS_WIS_4}). 
Therefore, $\lim_{i \rightarrow \infty } \ell(\mu_{U_i,Y}) - \ell(\mu_{X,Y})=0$ implies that $\lim_{i \rightarrow \infty} \ell(\mu_{U_i,\tilde{U}})=0$ under the discrimination condition on $\mu_{X,Y}$ stated in (\ref{eq_proof_th_OS_WIS_9}). 

{\bf Relaxing the discrimination condition in (\ref{eq_proof_th_OS_WIS_9}): }\\
The argument used above to prove that $\lim_{i \rightarrow \infty} \ell(\mu_{U_i,\tilde{U}})=0$ can be extended when we relax the condition stated in (\ref{eq_proof_th_OS_WIS_9}).  For that let us introduce the following set:
\begin{align} \label{eq_proof_th_OS_WIS_11}
	A_{\mu_{X,Y}}^\epsilon \equiv  \left\{ x \in  \mathcal{X}, f_{X,Y}(x,y_{(1)}) - f_{X,Y}(x,y_{(2)}) > \epsilon \cdot f_X(x) \right\}
\end{align}
where $y_{(1)} \equiv \arg \max_{y\in \mathcal{Y}} f_{X,Y}(x,y)$ and $y_{(2)} \equiv \arg \max_{y \in \mathcal{Y}, y \neq y_{(1)}} f_{X,Y}(x,y)$.\footnote{To simplify the notation, we omit the dependency of $y_{(1)}$  and $y_{(2)}$ on $x$.} Importantly, we have the following 
result: 
\begin{align} \label{eq_proof_th_OS_WIS_12}
	\lim_{\epsilon \rightarrow 0} \mu_x(A_{\mu_{X,Y}}^\epsilon) = \lim_{n \rightarrow \infty} \mu_x(A_{\mu_{X,Y}}^{1/n})= \mu_X(\bigcup_{n\geq 1} A_{\mu_{X,Y}}^{1/n}), 
\end{align}
where the last equality is from the continuity of $\mu_X$ under a sequence of monotonic events \cite{varadhan_2001}.
The following important result will be instrumental for our analysis: 
\begin{theorem}\label{lm_iif_unique_map_rule}
	The model $\mu_{X,Y}$ has a unique MPE decision rule (Def. \ref{def_unique_map_rule})  if, and only if, 
	\begin{align} \label{eq_lm_iif_unique_map_rule}
		\mu_X(\bigcup_{n\geq 1} A_{\mu_{X,Y}}^{1/n})= \mu_X(\left\{  x \in  \mathcal{X}, f_{X,Y}(x,y_{(1)}) > f_{X,Y}(x,y_{(2)})  \right\}) =1.
	\end{align}
\end{theorem}
This result offers a concrete characterization of models with a unique MPE decision rule. The proof is presented in Section \ref{proof_lm_iif_unique_map_rule}. 

Returning to the main argument, 
we have from (\ref{eq_proof_th_OS_WIS_7}) that for any $j\in \mathcal{Y}$
\begin{align} \label{eq_proof_th_OS_WIS_13}
&\mu_{X,Y}(A^*_j \setminus A^i_j  \times  \left\{ j \right\}) - \sum^M_{\tilde{j}=1, \tilde{j}\neq j}  \mu_{X,Y}(A^i_{\tilde{j}} \setminus A^*_{\tilde{j}} \cap A^*_j \setminus A^i_j  \times  \left\{ \tilde{j} \right\})\nonumber\\ 
	&\geq  \int_{A^*_j \setminus A^i_j }  \left[ f_{X,Y}(x,y_{(1)}) -  f_{X,Y}(x, y_{(2)})  \right] dx\\
	&\geq  \int_{A^*_j \setminus A^i_j  \cap A_{\mu_{X,Y}}^\epsilon}  \left[ f_{X,Y}(x,y_{(1)}) -  f_{X,Y}(x, y_{(2)})  \right] dx \geq \epsilon \int_{A^*_j \setminus A^i_j  \cap A_{\mu_{X,Y}}^\epsilon}  f_{X}(x) dx\\
	&=\epsilon \cdot \mu_X(A^*_j \setminus A^i_j  \cap A_{\mu_{X,Y}}^\epsilon), 
\end{align} 
where the second inequality comes from the fact that by definition $f_{X,Y}(x,y_{(1)}) -  f_{X,Y}(x, y_{(2)}) \geq 0$, and the 
third inequality from the definition of $A_{\mu_{X,Y}}^\epsilon$ in (\ref{eq_proof_th_OS_WIS_11}). Applying 
this last inequality in (\ref{eq_proof_th_OS_WIS_6}), it follows that for any $\epsilon>0$
\begin{align} \label{eq_proof_th_OS_WIS_14}
	\ell(\mu_{U_i,Y}) - \ell(\mu_{X,Y}) 	&\geq \epsilon \cdot \sum_{j=1}^M \mu_X(A^*_j \setminus A^i_j  \cap A_{\mu_{X,Y}}^\epsilon)\nonumber\\
								&= \epsilon \cdot  \mu_X( (\cup_{j=1}^M A^*_j \setminus A^i_j ) \cap A_{\mu_{X,Y}}^\epsilon).
\end{align} 
Consequently, using the assumption that  $\lim_{i \rightarrow \infty} \ell(\mu_{U_i,Y}) =\ell(\mu_{X,Y})$, it follows that for any $\epsilon>0$
\begin{align} \label{eq_proof_th_OS_WIS_15}
		\lim_{i \rightarrow \infty }  \mu_X( (\cup_{j=1}^M A^*_j \setminus A^i_j ) \cap A_{\mu_{X,Y}}^\epsilon)=0.
\end{align} 
Finally, for any $n\geq 1$, we have from (\ref{eq_proof_th_OS_WIS_15}) that 
\begin{align} \label{eq_proof_th_OS_WIS_16}
		\lim_{i \rightarrow \infty }  \mu_X( \cup_{j=1}^M A^*_j \setminus A^i_j ) &\leq  \lim_{i \rightarrow \infty }  \mu_X( (\cup_{j=1}^M A^*_j \setminus A^i_j ) \cap A_{\mu_{X,Y}}^{1/n}) + \mu_X((A_{\mu_{X,Y}}^{1/n})^c) \nonumber\\
		&=1- \mu_X(A_{\mu_{X,Y}}^{1/n}).
\end{align} 
This last bound implies that 
\begin{align} \label{eq_proof_th_OS_WIS_17}
\lim_{i \rightarrow \infty }  \mu_X( \cup_{j=1}^M A^*_j \setminus A^i_j ) 
				&\leq 1- \lim_{n \rightarrow \infty}  \mu_X(A_{\mu_{X,Y}}^{1/n})\nonumber\\
				&= 1- \mu_X(\bigcup_{n\geq 1} A_{\mu_{X,Y}}^{1/n}).
\end{align} 
At this point, we use the assumption that $\mu_{X,Y}$ has a unique MPE and Theorem \ref{lm_iif_unique_map_rule}
to obtain from (\ref{eq_proof_th_OS_WIS_17}) and (\ref{eq_proof_th_OS_WIS_4}) that 
\begin{align} \label{eq_proof_th_OS_WIS_18}
\lim_{i \rightarrow \infty }  \mu_X( \cup_{j=1}^M A^*_j \setminus A^i_j ) =0  \Rightarrow \lim_{i \rightarrow \infty } \ell(\mu_{U_i, \tilde{U}}) =0.
\end{align} 
\end{proof}

\subsubsection{Stage 2: $\lim_{i \rightarrow \infty} \ell(\mu_{U_i,\tilde{U}})=0 \Rightarrow$ WIS}
\begin{proof} 
For this part, we use the following result: 
\begin{lemma}\label{lm_fano}\cite[Th.1]{feder_1994}\footnote{The result in \cite[Th.1]{feder_1994} also offers a tight lower bound of the form $H(Y|X) \geq \psi(\ell(\mu_{X,Y}))$, where $\psi(\cdot)$ is presented in closed-form in \cite[Eq.(14)]{feder_1994}. Importantly, the result proves that both bounds are tight, i.e., they are achievable for some model $\mu_{X,Y}$ in the class $\mathcal{P}(\mathcal{X}\times \mathcal{Y})$.}
	For any model $\mu_{X,Y}\in \mathcal{P}(\mathcal{X}\times \mathcal{Y})$ in the mixed continuous-discrete setting introduced in Section \ref{sec_II},
	it follows that 
	$$\phi(\ell(\mu_{X,Y})) \geq H(Y|X)$$
	where $\phi(r)=h(r) + r \cdot \log (  \left| \mathcal{Y} \right| -1)$ and $h(r)=-r\log(r) -(1-r) \log (1-r)$ denotes the {\em binary entropy} \cite{cover_2006}.
\end{lemma}
Applying Lemma \ref{lm_fano} in our context, i.e. over the family of models $ \left\{\mu_{U_i,\tilde{U}} \right\}_{i\geq 1}$, we have that for any $i\geq 1$
\begin{align} \label{eq_proof_th_OS_WIS_19}
	h(\ell(\mu_{U_i,\tilde{U}})) + \ell(\mu_{U_i,\tilde{U}})\cdot \log (M-1) \geq  H(\tilde{U}|U_i).
\end{align} 
Using the result in (\ref{eq_proof_th_OS_WIS_18}) and the fact that $\lim_{r \rightarrow 0}h(r)=h(0)=0$ (the continuity of the binary entropy \cite{cover_2006}), we have from 
(\ref{eq_proof_th_OS_WIS_19}) that $\lim_{i \rightarrow \infty} H(\tilde{U}|U_i)=0$. Finally, by definition of the conditional MI \cite{cover_2006}, 
we have that $I(\tilde{U};Y|U_i) \leq H(\tilde{U}|U_i)$ which proves that $\lim_{i \rightarrow \infty} I(\tilde{U};Y|U_i)=0$ (WIS). 
\end{proof}

\subsection{Proof of Theorem \ref{lm_iif_unique_map_rule}}
\label{proof_lm_iif_unique_map_rule}
\begin{proof}
	First, it is simple to verify, from the definition of $A_{\mu_{X,Y}}^{\epsilon}$ in (\ref{eq_proof_th_OS_WIS_11}), 
	that  $\bigcup_{n\geq 1} A_{\mu_{X,Y}}^{1/n} = \left\{  x \in  \mathcal{X}, f_{X,Y}(x,y_{(1)}) > f_{X,Y}(x,y_{(2)})  \right\}$. 
	
	Let begin proving that if $\mu_{X,Y}$ has a unique MPE  rule  then $\mu_X(\bigcup_{n\geq 1} A_{\mu_{X,Y}}^{1/n} )=1$.
	We prove this implication by contradiction by assuming that $\mu_X(\bigcup_{n\geq 1} A_{\mu_{X,Y}}^{1/n} )<1$.  Let us denote 
	by 
	$$B \equiv \left\{  x \in  \mathcal{X}, f_{X,Y}(x,y_{(1)}) = f_{X,Y}(x,y_{(2)})  \right\} = (\bigcup_{n\geq 1} A_{\mu_{X,Y}}^{1/n})^c,$$ 
	which is non-empty by our assumption. From this set, we can 
	construct two different optimal (MPE) decision rules: 
	\begin{align}\label{eq_lm_iif_1}
		r_1(x)&=y_{(1)}=\tilde{r}_{\mu_{X,Y}}(x), \forall x\in \mathcal{X}
	\end{align}
	and 
	\begin{align}\label{eq_lm_iif_2}
	r_2(x) & = r_1(x), \forall x\in \mathcal{X} \setminus B\nonumber\\
	r_2(x) & = y_{(2)}, \forall x\in B.
	\end{align}
	By the definition of $y_{(1)}$, $y_{(2)}$ and $B$, we have that $\mathbb{P}(r_1(X)\neq Y)= \mathbb{P}(r_2(X)\neq Y)=\ell(\mu_{X,Y})$
	and $\mathbb{P}(r_1(X)\neq r_2(X))=\mu_X(B)>0$. Then,  $\mu_{X,Y}$  does not have a unique MPE from Definition \ref{def_unique_map_rule}. 
	
	For the other implication, let us assume that $\mu_X(B)=0$ and let us consider the optimal MAP rule $\tilde{r}_{\mu_{X,Y}}(x)=y_{(1)}$. 
	In addition, let us asume $r:\mathcal{X}  \rightarrow \mathcal{Y}$ s.t. $\mathbb{P}(r(X)\neq Y)=\ell({\mu_{X,Y}})$.  Both $\tilde{r}_{\mu_{X,Y}}(\cdot)$
	and $r(\cdot)$ induce an $M$-cell partition of $\mathcal{X}$ given by $\pi^*= \left\{A^*_j,j=1,..,M \right\}$ and $\pi = \left\{A_j= r^{-1}(\left\{ j \right\}) ,j=1,..,M \right\}$, respectively. 
	Using the same arguments used in the proof of Theorem \ref{th_OS_imply_WIS}, we have that:\footnote{In particular, from Eq.(\ref{eq_proof_th_OS_WIS_6}).} 
	\begin{align}\label{eq_lm_iif_3}
		&\mathbb{P}(r(X)\neq Y) - \ell(\mu_{X,Y}) =\nonumber\\  
		&\sum_{j=1}^M   \left[  \mu_{X,Y}(A^*_j \setminus A_j  \times  \left\{ j \right\}) - \sum^M_{\tilde{j}=1, \tilde{j}\neq j}  \mu_{X,Y}((A_{\tilde{j}} \setminus A^*_{\tilde{j}}) \cap (A^*_j \setminus A_j)  \times  \left\{ \tilde{j} \right\}) \right] =0,  
	\end{align}
	We observe that every term of this last summation over $j$ is non-negative from construction of the MAP rule. Consequently, to meet the zero condition in (\ref{eq_lm_iif_3}), it follows that for any $j=1,..,M$
	\begin{align}\label{eq_lm_iif_4}
		\mu_{X,Y}(A^*_j \setminus A_j  \times  \left\{ j \right\}) - \sum^M_{\tilde{j}=1, \tilde{j}\neq j}  \mu_{X,Y}((A_{\tilde{j}} \setminus A^*_{\tilde{j}}) \cap (A^*_j \setminus A_j)  \times  \left\{ \tilde{j} \right\})=0
	\end{align}
	Therefore, from the inequality presented in  Eq.(\ref{eq_proof_th_OS_WIS_7}) and (\ref{eq_lm_iif_4}), for any $j$
	\begin{align}\label{eq_lm_iif_5}
		\int_{A^*_j \setminus A_j }  \left[{ f_{X,Y}(x,y_{(1)}) - f_{X,Y}(x, y_{(2)})} \right]  dx = 0.
	\end{align}
	At this point, we consider the set $A_{\mu_{X,Y}}^\epsilon$ in (\ref{eq_proof_th_OS_WIS_11}),  where we have that 
	\begin{align}\label{eq_lm_iif_6}
		\int_{A^*_j \setminus A_j }  \left[{ f_{X,Y}(x,y_{(1)}) - f_{X,Y}(x, y_{(2)})} \right]  dx 
		&\geq  \int_{A^*_j \setminus A_j \cap  A_{\mu_{X,Y}}^{1/n}}  \left[{ f_{X,Y}(x,y_{(1)}) - f_{X,Y}(x, y_{(2)})} \right]  dx \nonumber\\
		&\geq \frac{1}{n} \int_{A^*_j \setminus A_j \cap  A_{\mu_{X,Y}}^{1/n}} f_{X}(x)  dx \nonumber\\
		&= \frac{1}{n}  \cdot \mu_X(A^*_j \setminus A_j \cap  A_{\mu_{X,Y}}^{1/n})=0, 
	\end{align}
	for any $j=1,..,M$ and any $n\geq 1$. Consequently, for additivity $\mu_X((\cup_{j=1}^M A^*_j \setminus A_j) \cap A_{\mu_{X,Y}}^{1/n} )=0$. 
	Then, for any $n\geq 1$
	\begin{align}\label{eq_lm_iif_7}
		\mu_X(\cup_{j=1}^M A^*_j \setminus A_j ) &\leq \mu_X((\cup_{j=1}^M A^*_j \setminus A_j) \cap A_{\mu_{X,Y}}^{1/n}) + (1-\mu_X( A_{\mu_{X,Y}}^{1/n}))\nonumber\\
		&= 1-\mu_X( A_{\mu_{X,Y}}^{1/n}).
	\end{align}
	Finally, taking the limit in $n$ in (\ref{eq_lm_iif_7}), we have that 
	\begin{align}\label{eq_lm_iif_8}
		\mu_X(\cup_{j=1}^M A^*_j \setminus A_j ) \leq 1 -\lim_{n \rightarrow \infty}\mu_X( A_{\mu_{X,Y}}^{1/n}) = 1- \mu_X(\cup_{n\geq 1} A_{\mu_{X,Y}}^{1/n})=0, 
	\end{align}
	the last equality from the assumption that $\mu_X(\cup_{n\geq 1} A_{\mu_{X,Y}}^{1/n})=1$.  To conclude, it is simple 
	to verify that \footnote{This inequality  
	follows from the same step presented to derive (\ref{eq_proof_th_OS_WIS_4}).}
	$$\mathbb{P}(r(X) \neq \tilde{r}_{\mu_{X,Y}}(X)) \leq  \mu_X(\cup_{j=1}^M A^*_j \setminus A_j )=0,$$ which 
	means that $r(\cdot)$ is equal to $\tilde{r}_{\mu_{X,Y}}(\cdot)$ almost surely. Therefore, the MPE rule associated to 
	$\mu_{X,Y}$ is unique (Def. \ref{def_unique_map_rule}) under the assumption that $\mu_X(\cup_{n\geq 1} A_{\mu_{X,Y}}^{1/n})=1$. 
\end{proof}

\subsection{Proof of Theorem \ref{th_wis_over_a_family}}
\label{proof_th_wis_over_a_family}
\begin{proof}
	Let us assume that $(X,Y)\sim \mu^\theta_{X,Y}$ for some arbitrary $\theta \in \Theta$.
	By the hypothesis in (\ref{eq_wis_learning_7}), we know that 
	\begin{equation}\label{eq_proof_th_wis_over_a_family_1}
		I(\eta^*(X);Y|\eta_i(X)) =  I((\eta^*(X),\eta_i(X));Y)  -  I(\eta_i(X);Y)   \longrightarrow 0
	\end{equation}
	as $i$ tends to infinity (the expressiveness condition of $\left\{\eta_i \right\}_{i\geq 1}$). 
	For the rest, we will use Theorem \ref{th_1}, for which we focus on $I((r_\theta(X),\eta_i(X));Y)  -  I((\eta_i(X));Y)$. 
	
	Using the fact that $H(r_\theta(X)|\eta^*(X))=0$,  it is simple to verify that for any $i\geq 1$
	\begin{equation}\label{eq_proof_th_wis_over_a_family_2}
	I((\eta^*(X), \eta_i(X));Y) \geq I((r_\theta(X), \eta_i(X));Y).  
	\end{equation}
	Indeed $\forall i\geq 1$
	\begin{align}\label{eq_proof_th_wis_over_a_family_3}
	&I((\eta^*(X), \eta_i(X));Y) - I((r_\theta(X), \eta_i(X));Y) \nonumber\\
	&= I((\eta^*(X), \eta_i(X), r_\theta(X));Y) - I((r_\theta(X), \eta_i(X));Y) \nonumber\\ 
	&= I(\eta^*(X);Y| r_\theta(X), \eta_i(X))  \geq 0,  
	\end{align}
	the first equality from the assumption that $H(r_\theta(X)|\eta^*(X))=0$ ($\eta^*(\cdot)$ is operationally sufficient for $\Lambda$) 
	and the second from the fact that conditional MI is non-negative \cite{cover_2006}.
	
	Using (\ref{eq_proof_th_wis_over_a_family_2}), we have that for any $i\geq 1$
	\begin{equation}\label{eq_proof_th_wis_over_a_family_4}
	 I((\eta^*(X),\eta_i(X));Y)  -  I(\eta_i(X);Y)  \geq   I((r_\theta(X),\eta_i(X));Y)  -  I(\eta_i(X);Y). 
	 \end{equation}
	Consequently, the asymptotic condition in (\ref{eq_proof_th_wis_over_a_family_1}) implies that 
	$(U_i=\eta_i(X))_{i\geq 1}$ is WIS (see Def. \ref{def_wis}) for $\mu^\theta_{X,Y}$ and the application of Theorem \ref{th_1} implies that (see Def. \ref{def_os}):
	\begin{equation}\label{eq_proof_th_wis_over_a_family_5}
		\lim_{i \rightarrow \infty} \ell(\mu^\theta_{\eta_i(X),Y}) = \ell(\mu^\theta_{X,Y}). 
	\end{equation}
	Finally, the presented argument is valid for any $\mu^\theta_{X,Y} \in \Lambda$, which concludes the proof.
\end{proof}


\appendices
\section{Proof of Proposition \ref{pro_opt_loss}}
\label{appendix_b}
\begin{proof}
From Bayes decision, it is known that $\tilde{U}=\tilde{r}_{\mu_{X,Y}}(X)$ is a sufficient statistic of $X$ in the operational sense, i.e.,  $\ell({\mu_{\tilde{U},Y}}) = \ell({\mu_{X,Y}})$. For this analysis, it is useful to consider the augmented observation vector $(\tilde{U},U_i)$, where its error $\ell({\mu_{(\tilde{U},U_i),Y}})$ is at most the error achieved by $\tilde{U}$.  Consequently,  we have that 
$\ell({\mu_{(\tilde{U},U_i),Y}}) = \ell ({\mu_{X,Y}}).$
This identity helps us to express the loss in (\ref{eq_sec_II_5}) conveniently:  
\begin{align}\label{eq_pro_opt_loss_1}
	\ell(\mu_{U_i,Y}) -  \ell(\mu_{X,Y}) &=  \ell(\mu_{U_i,Y})  -  \ell({\mu_{(\tilde{U},U_i),Y}}) =  \sum_{B_{i,j}\in \pi_i} \mu_X(B_{i,j}) \left[   1- \max_{y\in \mathcal{Y}} \mu_{Y|X}(y | B_{i,j}) \right] \nonumber\\
	&- \sum_{A^*_u\in \pi^*} \sum_{B_{i,j}\in \pi_i}  \mu_X(B_{i,j} \cap A^*_u)  \left[   1- \max_{y\in \mathcal{Y}} \mu_{Y|X}(y | A^*_u \cap B_{i,j}) \right].
\end{align}
Finally (\ref{eq_sec_II_8b}) follows directly from (\ref{eq_pro_opt_loss_1}).
\end{proof}

\section{Proof of Proposition \ref{pro_inf_loss}}
\label{appendix_c}
\begin{proof}
From the definition of MI and the discrete nature of the joint vector $(\tilde{U},U_i)$ \cite{cover_2006}, we have that 
\begin{equation}
	\mathcal{I}( \mu_{(\tilde{U}, U_i),Y})= H(Y) - \sum_{A^*_u\in \pi^*} \sum_{B_{i,j}\in \pi_i}   \mu_X(B_{i,j} \cap A^*_u)  \cdot \mathcal{H} (\mu_{Y|X}( \cdot |A^*_u \cap B_{i,j})).
\end{equation}
On the other hand, 
	$\mathcal{I}(\mu_{U_i,Y})=  H(Y) - \sum_{B_{i,j}\in \pi_i}   \mu_X(B_{i,j})  \cdot  \mathcal{H} (\mu_{Y|X}(\cdot |B_{i,j}))$.
The result in (\ref{eq_sec_II_8c}) derives directly from these expressions.
\end{proof}
	
\section{Proof of Lemma \ref{lm_maximum_entropy}}
\label{appendix_lm_maximum_entropy}
\begin{proof}
	 Let us consider an arbitrary $\mu\in \mathcal{P}^\epsilon([M])$,   
	 where we have that $\mu(1) \geq \mu(2) \geq \ldots \mu(K) >\theta$ and that 
	 $\sum_{j=2}^K(\mu(j)  -\theta) = \epsilon$. In this analysis, the dependency of $K$ and $\theta$ on $\mu$ and $\epsilon$
	 will be considered implicit.  
	 We consider the conditional probability $\tilde{\mu} \equiv \mu(\cdot |\beta) \in \mathcal{P}([M])$
	 for the set $\beta=\left\{2,\dots, K \right\}$, i.e., 
	 \begin{align}\label{eq_appen_lm_max_entr_1}
	 &\tilde{\mu}(2) = \frac{\mu(2)}{\theta(K-1) + \epsilon} \geq \tilde{\theta} \equiv \frac{\theta}{\theta(K-1) + \epsilon}>0,\nonumber\\
	 & \ldots\nonumber\\
	 &\tilde{\mu}(K) = \frac{\mu(K)}{\theta(K-1) + \epsilon} \geq  \tilde{\theta}. 
	 \end{align}
	 In this context, it is instrumental to introduce the following family of admissible distributions $\left\{\bar{e}_2, \ldots, \bar{e}_K \right\}\subset \mathcal{P}([M])$ with support in $\beta$, where $\bar{e}_j$ is given by
	 \begin{align}\label{eq_appen_lm_max_entr_2}
	 \bar{e}_j(2) &=\tilde{\theta}, 
	 \ldots,\nonumber\\ 
	 \bar{e}_j(j-1) &=\tilde{\theta}, \nonumber\\ 
	 \bar{e}_j(j) &=\tilde{\theta} + \frac{\epsilon}{\theta(K-1) + \epsilon}, \nonumber\\ 
	 \bar{e}_j(j+1) &=\tilde{\theta}, 
	 \ldots, \nonumber\\
	 \bar{e}_j(K) &=\tilde{\theta}, \nonumber\\ 
	 \end{align}
	 
	 Importantly, it is simple to verify that $\tilde{\mu}$ (in (\ref{eq_appen_lm_max_entr_1})) can be written as a convex combination of our admissible family $\left\{\bar{e}_2, \ldots, \bar{e}_K \right\}$, i.e., $\exists (w_2,..,w_K)\in [0,1]^{K-1}$ such that $\sum_{j=2}^{K}w_j=1$ and 
	 \begin{align}\label{eq_appen_lm_max_entr_3}
	 \tilde{\mu} = \sum_{j=2}^{K} w_j \cdot \bar{e}_j, 
	 \end{align}	
	 where $w_j= \frac{\tilde{\mu}(j) -  \tilde{\theta}}{ \tilde{\epsilon}}$ with $\tilde{\epsilon} \equiv \frac{\epsilon}{\theta(K-1) + \epsilon}>0$.
	 
	Let us define two random variables $Z$ and $O$ such that $Z$ takes values in $[M]$ and $O$ takes values in $\left\{2,..,K \right\}$
	and
	 \begin{align}\label{eq_appen_lm_max_entr_4}
	 P_{Z|O}(\cdot |k) = \bar{e}_k \in \mathcal{P}([M]),\text{ and }P_O(k)=w_k,
	 \end{align}	
	 $\forall k \in \left\{2,..,K \right\}$.  By construction, $P_Z=\sum_{j=2}^K w_j \cdot \bar{e}_j= \tilde{\mu}$. Therefore, we can use that 
	 $H(Z|O) \leq H(Z)$ \cite{cover_2006}, which implies that $\sum_{j=2}^K w_j \cdot \mathcal{H}(\bar{e}_j) \leq \mathcal{H}(\tilde{\mu})$.
	 Finally, by the invariant of the entropy to one-to-one permutations,  $\mathcal{H}(\bar{e}_2)=\ldots=\mathcal{H}(\bar{e}_K)$, then 
	 we have that  $\mathcal{H}(\bar{e}_2) \leq \mathcal{H}(\tilde{\mu})$, which implies that 
	 \begin{align}\label{eq_appen_lm_max_entr_5}
	 (\tilde{\theta} + \tilde{\epsilon})\log \frac{1}{\tilde{\theta} + \tilde{\epsilon}} + (K-2) \tilde{\theta} \log \frac{1}{ \tilde{\theta}} \leq \mathcal{H}(\tilde{\mu}) 
	 \end{align}
	 Returning to our original problem, we have that 	
	 \begin{align}\label{eq_appen_lm_max_entr_6}
	 &\sum_{j=2}^K {\mu}(j) \log \frac{1}{{\mu}(j)}= \mu(\beta) \mathcal{H}(\tilde{\mu})  + \mu(\beta) \log \frac{1}{\mu(\beta)}\geq\nonumber\\
	 &(\theta (K-1) + \epsilon) \cdot \left[   (\tilde{\theta} + \tilde{\epsilon})\log \frac{1}{\tilde{\theta} + \tilde{\epsilon}} + (K-2) \tilde{\theta} \log \frac{1}{ \tilde{\theta}} \right] +  (\theta (K-1) + \epsilon) \log \frac{1}{(\theta (K-1) + \epsilon)} \nonumber\\
	 &=(\theta + \epsilon) \log \frac{(K-1)\theta +\epsilon }{\theta + \epsilon}  + (K-2) \theta \log \frac{(K-1)\theta + \epsilon}{\theta}  + (\theta (K-1) + \epsilon) \log \frac{1}{(\theta (K-1) + \epsilon)}\nonumber\\
	 &= (\theta + \epsilon) \log \frac{1}{\theta + \epsilon}  +  (K-2) \theta \log \frac{1}{\theta}, 
	 \end{align}
	 where for the first inequality we use the lower bound in (\ref{eq_appen_lm_max_entr_5}) and the fact that $\mu(\beta)=\theta (K-1) + \epsilon$, 
	 and for the first equality we use that $\tilde{\theta}=\theta/((K-1)\theta +\epsilon)$ and 
	 $\tilde{\epsilon}=\epsilon/((K-1)\theta +\epsilon)$. Finally, (\ref{eq_appen_lm_max_entr_6}) proves the result. 	 
\end{proof}

\section{Proof of Proposition \ref{prop_representation_for_invariant_models}}
\label{proof_prop_representation_for_invariant_models}
\begin{proof}
	Let $\eta^*(\cdot)$ be maximal invariant for $\mathcal{G}$.  This means that for any pair $(x,y)\in \mathcal{X}^2$
	where $\mathcal{G}(x) \neq \mathcal{G}(y)$ then $\eta^*(x) \neq \eta^*(y)$. We have that for any $\mu^\theta_{X,Y} \in \Lambda$, $r_\theta(\cdot)$  solution of (\ref{eq_wis_learning_1}) is $\mathcal{G}$-invariant (see Def.\ref{def_inv_rules}), this invariant condition implies that $r_\theta(\cdot)$ is fully determined if we know the values of $r_\theta(\cdot)$ in every cell of $\pi_\mathcal{G}= \left\{\mathcal{G}(x), x \in \mathcal{X} \right\}$. Therefore, if we know $r_\theta(\eta^*(x))$ for any $x\in \mathcal{X}$, we fully determine $r_\theta(\cdot)$ using the fact that $\eta^*(\cdot)$ is maximal invariant. Indeed, we have that $r_\theta(x)=r_\theta(\eta^*(x))$, which implies that $H(r_\theta(X)|\eta^*(X))=0$ where $X\sim \mu^\theta_X$. Then, we use Proposition \ref{pro_OS_from_invariance} to conclude the proof.
\end{proof}

\section{Proposition \ref{pro_strong_invariace_weak_invariance}}
\label{app_pro_strong_invariace_weak_invariance}
\begin{proposition}\label{pro_strong_invariace_weak_invariance}
	If $\mu_{X,Y}\in \mathcal{P}(\mathcal{X}\times \mathcal{Y})$ is model-based $\mathcal{G}$-invariant,  in the sense that $(X,Y)=(g(X),Y)$ in distribution for any $g\in \mathcal{G}$, then $\mu_{X,Y}$ is operational $\mathcal{G}$-invariant (see Def.\ref{def_inv_models}).
\end{proposition}
\begin{proof}
	If $\mu_{X,Y}$ is model-based $\mathcal{G}$-invariant, this implies that $\mu_{Y|X}(y|x)=\mu_{Y|X}(y|g(x))$ for any $y\in \mathcal{Y}$ and $x\in \mathcal{X}$. From this property of the posterior, it is direct to show that there exists $r(\cdot)$, which is solution of (\ref{eq_sec_II_6}) (for the joint model $\mu_{X,Y}$) that is functional $\mathcal{G}$-invariant (Def. \ref{def_inv_rules}).
\end{proof}

\section{The Construction of $\mathcal{R}(\mu ,\epsilon)$}
\label{appendix_rate_distortion_function}
The model $\mathcal{R}(\mu ,\epsilon)$ that solves the optimization problem in (\ref{eq_main_1}) is presented here for completeness.  As this optimization problem is function of $\mu\in \mathcal{P}(\left[ M\right])$ and $\epsilon>0$, the model $\mathcal{R}(\mu ,\epsilon) \in \mathcal{P}(\left[ M\right])$ is an exploit function of these two elements.  
To find the simples description of this object, we assume, without loss of generality,  
that $\mu(1)\geq \mu(2) \geq \ldots \mu(M)$. In this context, Ho and Verd\'{u} \cite{ho_2010} showed that the model $\mu^{\bar{\epsilon}} \equiv \mathcal{R}(\mu , prior(\mu)- \bar{\epsilon})$ for $\bar{\epsilon} \in [0,prior(\mu)]$ has the follows form:  
	\begin{align}
	\label{eq_app_rate_distortion_function_1}
	&\mu^{\bar{\epsilon}}(1)= \mu(1) + \bar{\epsilon} \leq 1 \\
	\label{eq_app_rate_distortion_function_1b}
	&\mu^{\bar{\epsilon}}(2)=  \theta\nonumber\\ 
	&\ldots	\nonumber\\ 
	&\mu^{\bar{\epsilon}}(K)=  \theta\\ 
	\label{eq_app_rate_distortion_function_1c}
	&\mu^{\bar{\epsilon}}(K+1) = \mu(K+1) \nonumber\\ 
	&\ldots	\nonumber\\ 
	&\mu^{\bar{\epsilon}}(M) = \mu(M),  
	\end{align}
where $K\in \left\{2,..,M \right\}$ and $\theta \in (0,\mu(1))$ are functions of $\mu$ and $\bar{\epsilon}$ meeting the 
following condition
	\begin{align}\label{eq_app_rate_distortion_function_2}
	\sum_{j=2}^K  \left(  \mu(i)-\theta \right) = \bar{\epsilon}>0. 
	\end{align}
Importantly, Ho and Verd\'{u} \cite{ho_2010} showed that for any $\epsilon \leq prior(\mu)$, $\exists \theta \in (0,\mu(1))$
and $K\in \left\{2,..,M \right\}$ that meet the condition in  (\ref{eq_app_rate_distortion_function_2}) for ${\bar{\epsilon}}=prior(\mu) - \epsilon$. From the 
estructure of $\mu^{\bar{\epsilon}}$ in (\ref{eq_app_rate_distortion_function_1}), it follows that if  $\epsilon=prior(\mu)$ then $\mu^{\bar{\epsilon}} = \mu$. 
On the other hand, if we assume that $\epsilon < prior(\mu)$  then $\mu^{\bar{\epsilon}} \neq \mu$ and $H(\mu) > H(\mu^{\bar{\epsilon}})$, which implies that  $f(\mu, \epsilon)>0$ in (\ref{eq_main_1}) in this regime. 

Finally, from the closed-from expressions in (\ref{eq_app_rate_distortion_function_1})-(\ref{eq_app_rate_distortion_function_1c}) we have that: 
	\begin{align}\label{eq_app_rate_distortion_function_3}
	f(\mu, \epsilon = prior(\mu)- \bar{\epsilon}) &= H(\mu) - H(\mu^{\bar{\epsilon}})\nonumber\\
				&= \mu(1) \log \frac{1}{ \mu(1)} - (\mu(1) + \bar{\epsilon}) \log \frac{1}{ \mu(1) + \bar{\epsilon}}  \nonumber\\
				&+ \sum^{K(\mu,\epsilon)}_{j=2} \mu(j) \log \frac{1}{ \mu(j)} -  (K(\mu,\epsilon)-1) \theta(\mu,\epsilon) \log \frac{1}{\theta(\mu,\epsilon)} \geq 0, 
	\end{align}
	where in this last expression we make explicit the fact that the parameters $K$ and $\theta$ are explicit 
	function of $\mu$ and $\epsilon$. 
\bibliographystyle{plain}				
\bibliography{main_jorge_silva}	

\end{document}